\documentclass[10pt,twocolumn,letterpaper]{article}

\usepackage[pagenumbers]{cvpr} 

\usepackage{times}
\usepackage{epsfig}
\usepackage{graphicx}
\usepackage{amsmath}
\usepackage{amssymb}
\usepackage{multirow}
\usepackage{booktabs}

\usepackage{amsfonts}
\usepackage{bm}
\usepackage{amsmath,amsthm,amssymb,rotating, mathrsfs}

\usepackage{mathtools}
\usepackage{cancel} 

\usepackage[linesnumbered,lined,boxed,commentsnumbered, ruled,vlined]{algorithm2e}
\usepackage{xfrac} 
\usepackage{xparse} 

\makeatletter
\@namedef{ver@everyshi.sty}{}
\makeatother
\usepackage{tikz}

\usepackage{enumitem}
\newtheorem{theorem}{Theorem}

\newtheorem{prop}{Proposition}

\newtheorem{lemma}{Lemma}

\theoremstyle{definition}
\newtheorem{definition}{Definition}

\newtheorem{assumption}{Assumption}

\newtheorem{problem}{Problem}

\theoremstyle{remark}
\newtheorem{remark}{Remark}

\newcommand{\cC}{\mathcal{C}}

\newcommand{\cG}{\mathcal{G}}

\newcommand{\cI}{\mathcal{I}}
\newcommand{\cJ}{\mathcal{J}}

\newcommand{\cN}{\mathcal{N}}

\newcommand{\cP}{\mathcal{P}}
\newcommand{\cQ}{\mathcal{Q}}

\newcommand{\cS}{\mathcal{S}}

\newcommand{\bbE}{\mathbb{E}}

\newcommand{\bbP}{\mathbb{P}}

\newcommand{\bbR}{\mathbb{R}}
\newcommand{\bbS}{\mathbb{S}}

\newcommand{\bA}{\bm{A}}

\newcommand{\bC}{\bm{C}}
\newcommand{\bD}{\bm{D}}

\newcommand{\bH}{\bm{H}}
\newcommand{\bI}{\bm{I}}

\newcommand{\bK}{\bm{K}}

\newcommand{\bR}{\bm{R}}

\newcommand{\bU}{\bm{U}}

\newcommand{\bX}{\bm{X}}

\newcommand{\bZ}{\bm{Z}}

\newcommand{\ba}{\bm{a}}

\newcommand{\bo}{\bm{o}}
\newcommand{\bp}{\bm{p}}
\newcommand{\bq}{\bm{q}}

\newcommand{\bt}{\bm{t}}

\newcommand{\bv}{\bm{v}}
\newcommand{\bw}{\bm{w}}
\newcommand{\bx}{\bm{x}}
\newcommand{\by}{\bm{y}}
\newcommand{\bz}{\bm{z}}

\newcommand{\beps}{\bm{\epsilon}}

\NewDocumentCommand{\norm}{mG{2}}{\big\|#1\big\|_{#2}}

\newcommand{\trans}{\top}
\newcommand{\trsp}[1]{#1^\trans}

\DeclareMathOperator{\rank}{rank}

\DeclareMathOperator{\ICP}{\texttt{ICP}}
\DeclareMathOperator{\GOICP}{\texttt{GO-ICP}}

\DeclareMathOperator{\TIM}{\texttt{TIM}}

\DeclareMathOperator{\SO}{SO}

\DeclareMathOperator{\FPFH}{\texttt{FPFH}}
\DeclareMathOperator{\FGR}{\texttt{FGR}}
\DeclareMathOperator{\RANSAC}{\texttt{RANSAC}}
\DeclareMathOperator{\GORE}{\texttt{GORE}}

\DeclareMathOperator{\TEASER}{\texttt{TEASER++}}

\DeclareMathOperator{\GNC}{\texttt{GNC}}
\DeclareMathOperator{\GNCTLS}{\texttt{GNC-TLS}}

\DeclareMathOperator{\QUASAR}{\texttt{QUASAR}}

\DeclareMathOperator{\ARCS}{\texttt{ARCS}}
\DeclareMathOperator{\ARCSplus}{\texttt{ARCS+}}
\DeclareMathOperator{\ARCSplusplus}{\texttt{ARCS++}}

\DeclareMathOperator{\dist}{dist}

\NewDocumentCommand{\seq}{mG{n}}{{#1}_1, \dots, {#1}_{#2}}
\NewDocumentCommand{\seqp}{mG{n}}{{#1}_1-\cdots+ {#1}_{#2}}
\NewDocumentCommand{\seqm}{mG{n}}{{#1}_1-\cdots- {#1}_{#2}}

\DeclarePairedDelimiter\ceil{\lceil}{\rceil}
\DeclarePairedDelimiter\floor{\lfloor}{\rfloor}
\newcommand{\myparagraph}[1]{\smallskip\noindent\textbf{#1.}}

\DeclareMathOperator{\rsgrad}{\tilde{\nabla}_{\textnormal{s}}}
\DeclareMathOperator{\sgrad}{\nabla_{\textnormal{s}}}

\DeclareMathOperator{\Proj}{Proj}

\usepackage{epigraph}

\usepackage[pagebackref,breaklinks,colorlinks]{hyperref}

\usepackage[capitalize]{cleveref}
\crefname{section}{Sec.}{Secs.}
\Crefname{section}{Section}{Sections}
\Crefname{table}{Table}{Tables}
\crefname{table}{Tab.}{Tabs.}

\def\cvprPaperID{*****} 
\def\confName{CVPR}
\def\confYear{2022}

\begin{document}

\title{ARCS: Accurate Rotation and Correspondence Search}

\author{Liangzu Peng \\
	Johns Hopkins University\\
	{\tt\small lpeng25@jhu.edu}\\
	\and 
	\and
	Manolis C. Tsakiris\\
	ShanghaiTech University\\
	{\tt\small mtsakiris@shanghaitech.edu.cn}
	\and
	Ren\'e Vidal \\
	Johns Hopkins University\\
	{\tt\small rvidal@jhu.edu}\\
}

\maketitle

\begin{abstract}
	This paper is about the old Wahba problem in its more general form, which we call ``simultaneous rotation and correspondence search''. In this generalization we need to find a rotation that best aligns two partially overlapping $3$D point sets, of sizes $m$ and $n$ respectively with $m\geq n$. We first propose a solver, $\texttt{ARCS}$, that i) assumes noiseless point sets in general position, ii) requires only $2$ inliers, iii) uses $O(m\log m)$ time and $O(m)$ space, and iv) can successfully solve the problem even with, \eg, $m,n\approx 10^6$ in about $0.1$ seconds. We next robustify $\texttt{ARCS}$ to noise, for which we approximately solve consensus maximization problems using ideas from robust subspace learning and interval stabbing. Thirdly, we refine the approximately found consensus set by a Riemannian subgradient descent approach over the space of unit quaternions, which we show converges globally to an $\varepsilon$-stationary point in $O(\varepsilon^{-4})$ iterations, or locally to the ground-truth at a linear rate in the absence of noise.  We combine these algorithms into $\texttt{ARCS+}$, to simultaneously search for rotations and correspondences. Experiments show that $\texttt{ARCS+}$  achieves state-of-the-art performance on large-scale datasets with more than $10^6$ points with a $10^4$ time-speedup over alternative methods. \url{https://github.com/liangzu/ARCS}
\end{abstract}

\section{Introduction}\label{section:intro}
\epigraph{The villain Procrustes forced his victims to sleep on an iron bed; if they did not fit the bed he cut off or stretched their limbs to make them fit \cite{Everson-1998}. }{Richard Everson}

Modern sensors have brought the classic \textit{Wahba} problem \cite{Wahba-SIAM-Review-1965}, or slightly differently the \textit{Procrustes analysis} problem \cite{Gower-2004}, into greater generality that has increasing importance to computer vision \cite{Li-ICCV07,Hartley-IJCV2009}, computer graphics \cite{Maron-ToG2016}, and robotics \cite{Bernreiter-RA-L2021}. We formalize this generalization as follows.
\begin{problem}[\textit{simultaneous rotation and correspondence search}]\label{problem:SRSC}
	Consider point sets $\cQ=\{\bq_1,\dots,\bq_m\}\subset\bbR^3$ and $\cP=\{\bp_1,\dots,\bp_n\}\subset\bbR^3$ with $m\geq n$. 
	Let $\cC^*$ be a subset of $[m]\times [n]:=\{1,\dots,m\}\times \{1,\dots, n\}$ of size $k^*$, called the \textit{inlier correspondence} set, such that all pairs $(i_1,j_1)$ and $(i_2,j_2)$ of $\cC^*$ satisfy $i_1\neq i_2$ and $j_1\neq j_2$. Assume that
	\begin{align}\label{eq:inlier-outlier-def}
		\bq_{i} &= \bR^*\bp_j + \beps_{i,j}, \ \ \  \textnormal{if}\  (i,j)\in \cC^*
	\end{align} 
	where $\beps_{i,j}\sim \cN(0,\sigma^2\bI_3)$ is noise, $\bR^*$ is an unknown $3$D rotation, and $(\bq_i,\bp_j)$ is called an \textit{inlier}. If $(i,j)\notin \cC^*$ then $(\bq_i,\bp_j)$ is arbitrary and is called an \textit{outlier}. The goal of the simultaneous rotation and correspondence search problem is to simultaneously estimate the $3$D rotation $\bR^*$ and the \textit{inlier correspondence set} $\cC^*$ from point sets $\cQ$ and $\cP$.
	
	
\end{problem}
We focus on Problem \ref{problem:SRSC} for two reasons. First, it already encompasses several vision applications such as image stitching \cite{Bustos-ICCV2015}. Second, the more general and more important \textit{simultaneous pose and correspondence} problem, which involves an extra unknown translation in \eqref{eq:inlier-outlier-def}, reduces to Problem \ref{problem:SRSC} by eliminating the translation parameters (at the cost of squaring the number of measurements) \cite{Yang-T-R2021}. As surveyed in \cite{Huang-arXiv2021}, whether accurate and fast algorithms exist for solving the pose and correspondence search is largely an open question. Therefore, solving the simpler Problem \ref{problem:SRSC} efficiently is an important step for moving forward.


For Problem \ref{problem:SRSC} or its variants, there is a vast literature of algorithms that are based on i) local optimization via \textit{iterative closest points} ($\ICP$) \cite{Besl-PAMI1992,Chetverikov-ICPR2002,Rusinkiewicz-2001} or \textit{graduated non-convexity} ($\GNC$) \cite{Zhou-ECCV2016,Yang-RA-L2020,Antonante-arXiv2020} or others \cite{Myronenko-PAMI10,Jian-PAMI11,Chui-CVIU03}, ii) global optimization by branch $\&$ bound \cite{Li-ICCV07,Chin-CVPR2016,Yang-PAMI16,Bustos-PAMI16,Campbell-CVPR2016,Straub-CVPR2017,Lian-PAMI17,Liu-ECCV18}, iii) outlier removal techniques \cite{Bustos-ICCV2015,Bustos-TPAMI2018,Parra-arXiv2020,Yang-T-R2021,Shi-arXiv2020v2}, iv) semidefinite programming \cite{Maron-ToG2016,Yang-ICCV2019,Iglesias-CVPR2020,Yang-arXiv2021,Sun-arXiv2021b}, v) RANSAC \cite{Fischler-C-ACM1981,Li-ISPRS-J-PRS2020,Sun-arXiv2021,Li-TGRS2021}, vi) deep learning \cite{Choy-CVPR2020,Huang-CVPR2021,Bauer-CVPR2021,Bai-CVPR2021}, and vii) spherical Fourier transform \cite{Bernreiter-RA-L2021}. But all these methods, if able to accurately solve Problem \ref{problem:SRSC} with the number $k^*$ of inliers extremely small, take $\Omega(mn)$ time. Yet we have: 
\begin{theorem}[$\ARCS$]\label{theorem:I2}
	Suppose there are at least two inliers, $k^*\geq2$, and that the point sets $\cQ$ and $\cP$ of Problem \ref{problem:SRSC} are noiseless ``in general position''. Then there is an algorithm that solves Problem \ref{problem:SRSC} in $O(m\log m)$ time and $O(m)$ space.
\end{theorem}
\begin{remark}[\textit{general position assumption}]\label{remark:generalposition}
	In Theorem \ref{theorem:I2}, by ``\textit{in general position}'' we mean that i) for any outlier $(\bq_i,\bp_j)$, we have $\norm{\bq_i}\neq \norm{\bp_j}$, ii) there exists some inlier pairs $(\bq_{i_1},\bp_{j_1})$ and $(\bq_{i_2},\bp_{j_2})$ such that  $\bq_{i_1}$ and $\bq_{i_2}$ are not parallel. If point sets $\cQ$ and $\cP$ are randomly sampled from $\bbR^3$, these two conditions hold true with probability $1$.
\end{remark}

A numerical illustration of Theorem \ref{theorem:I2} is that our $\ARCS$ solver, to be described  in \S \ref{section:I2}, can handle the case where $m=10^6,n=8\times 10^5$ and $k^*=2$, in about $0.1$ seconds (cf. Table \ref{table:I2}).\footnote{We run experiments on an Intel(R) i7-1165G7, $16$GB laptop. In the paper we consider random instead of adversarial outliers.} However, like other correspondence-based minimal solvers for geometric vision \cite{Gao-TPAMI2003,Nister-TPAMI2004,Kukelova-ECCV2008,Larsson-CVPR2017,Larsson-CVPR2018}, $\ARCS$ might be fragile to noise. That being said, it can be extended to the noisy case, leading to a three-step algorithm called $\ARCSplus$, which we summarize next.

The first step $\ARCSplus_\texttt{N}$ of $\ARCSplus$ extends $\ARCS$ by \textit{establishing correspondences under noise}. $\ARCSplus_\texttt{N}$ outputs in $O(\ell+ m\log m)$ time a candidate correspondence set $\overline{\cC}$  of size $\ell$  that contains $\cC^*$. Problem \ref{problem:SRSC} then reduces to estimating $\bR^*$ and $\cC^*$ from $\cP,\cQ,$ and hypothetical correspondences $\overline{\cC}$, a simpler task of \textit{robust rotation search} \cite{Zhou-ECCV2016,Bustos-ICCV2015,Bustos-TPAMI2018,Yang-ICCV2019}. 

The second step $\ARCSplus_\texttt{O}$ of $\ARCSplus$ is to \textit{remove outliers} from the previous step $1$. To do so we approximately maximize an appropriate consensus over $\SO(3)$ (\S \ref{section:STABBER3D}). Instead of mining inliers in $\SO(3)$ \cite{Li-ICCV07,Hartley-IJCV2009,Bazin-ACCV2012,Bustos-PAMI16,Joo-PAMI2021}, we show that the parameter space of consensus maximization can be reduced from $\SO(3)$ to $\bbS^2$ and further to $[0,\pi]$ (see \cite{Bustos-ICCV2015} for a different reduction). With this reduction, $\ARCSplus_\texttt{O}$ removes outliers via repeatedly solving in $O(\ell\log \ell)$ time a computational geometry problem, \textit{interval stabbing} \cite{Berg-1997} (\S \ref{subsection:interval stabbing}). Note that $\ARCSplus_\texttt{O}$ only repeats for $s\approx 90$ times to reach satisfactory accuracy. Therefore, conceptually, for $\ell\geq 10^6$, it is $10^4$ times faster than the most related outlier removal method $\GORE$ \cite{Bustos-ICCV2015}, which uses $O(\ell^2\log \ell)$ time (Table \ref{table:RRS-scalibility}). 


The third and final step $\ARCSplus_\texttt{R}$ of our $\ARCSplus$ pipeline is to accurately estimate the rotation, using the consensus set from the second step (\S \ref{section:R}). In short, $\ARCSplus_\texttt{R}$ is a \textit{Riemannian subgradient descent} method. Our novelty here is to descend in the space $\bbS^3$ of unit quaternions, not $\SO(3)$ \cite{Bohorquez-arXiv2020v3}. This allows us to derive, based on \cite{Li-SIAM-J-O2021}, that $\ARCSplus_\texttt{R}$ converges linearly though locally to the ground-truth unit quaternion, thus obtaining the first to our knowledge convergence rate guarantee for robust rotation search.

Numerical highlights are in order (\S \ref{section:experiments}). $\ARCSplus_\texttt{O}$ is an outlier pruning procedure for robust rotation search that can handle extremely small inlier ratios $k^*/\ell=3000/10^7=0.03\%$ in $5$ minutes; $\ARCSplus_{\texttt{O}}+\ARCSplus_{\texttt{R}}$, or $\ARCSplus_{\texttt{OR}}$ for short, accurately solves the robust rotation search problem with $k^*/\ell= 10^3/10^6$ in $23$ seconds (see Table \ref{table:RRS-scalibility}). $\ARCSplus_{\texttt{N}}+\ARCSplus_{\texttt{OR}}$, that is $\ARCSplus$, solves Problem \ref{problem:SRSC} with $m=10^4,n=8000,k^*=2000$ in $90$ seconds (see Figure \ref{fig:experiments_SRSC}). 
To the best of our knowledge, all these challenging cases have not been considered in prior works. In fact, as we will review soon (\S \ref{subsection:prior-art}), applying state-of-the-art methods to those cases either gives wrong estimates of rotations, or takes too much time ($\geq 8$ hours), or exhausts the memory (Table \ref{table:RRS-scalibility}). 

\section{Prior Art: Accuracy Versus Scalability}\label{subsection:prior-art}
Early efforts on Problem \ref{problem:SRSC} have encountered an \textit{accuracy versus scalability} dilemma. The now classic $\ICP$ algorithm \cite{Besl-PAMI1992} estimates the rotation and correspondences in an alternating fashion, running in real time but requiring a high-quality and typically unavailable initialization to avoid local and usually poor minima; the same is true for its successors \cite{Chetverikov-ICPR2002,Rusinkiewicz-2001,Chui-CVIU03,Myronenko-PAMI10,Jian-PAMI11}. The $\GOICP$ method \cite{Yang-ICCV2013,Yang-PAMI16} of the branch $\&$ bound type enumerates initializations fed to $\ICP$ to reach a global minimum---in exponential time; the same running time bound is true for its successors \cite{Campbell-CVPR2016,Bustos-PAMI16,Liu-ECCV18}.

The above $\ICP$ versus $\GOICP$ dilemma was somewhat alleviated by a two-step procedure: i) compute a candidate correspondence set $\hat{\cC}$,  via hand-crafted \cite{Rusu-ICRA2009} or learned \cite{Gojcic-CVPR2019} feature descriptors, and ii) estimate the rotation from point sets indexed by $\hat{\cC}$. But, as observed in \cite{Yang-T-R2021}, due to the quality of the feature descriptors, there could be fewer than $2$ inliers remaining in $\hat{\cC}$, from which the ground-truth rotation can never be determined. An alternative and more conservative idea is to use \textit{all-to-all} correspondences $\hat{\cC}:=[m]\times [n]$, although now the inlier ratio becomes extremely small.

This justifies why researchers have recently focused on designing robust rotation search algorithms for extreme outlier rates, \eg, $\geq 90$ outliers out of $100$. One such design is $\GORE$ \cite{Bustos-ICCV2015}, a guaranteed outlier removal algorithm of $O(\ell^2\log \ell)$ time complexity that heavily exploits the geometry of $\SO(3)$. The other one is the semidefinite relaxation $\QUASAR$ of \cite{Yang-ICCV2019}, which involves sophisticated manipulation on unit quaternions; $\ell\approx 1000$ constitutes the current limit on the number of points this relaxation can handle. Yet another one is $\TEASER$ \cite{Yang-T-R2021}; its robustness to outliers comes mainly from finding via parallel branch $\&$ bound \cite{Rossi-SIAM-J-SC2015} a maximum clique of the graph whose vertices represent point pairs and whose edges indicate whether two point pairs can simultaneously be inliers. This maximum clique formulation was also explored by \cite{Parra-arXiv2020} where it was solved via a different branch $\&$ bound algorithm. Since finding a maximum clique is in general NP-hard, their algorithms take exponential time in the worst case; in addition, $\TEASER$ was implemented to trade $O(\ell^2)$ space for speed. One should also note though that if noise is small then the graph is sparse so that the otherwise intractable branch $\&$ bound algorithm can be efficient. Since constructing such a graph entails checking $\binom{\ell}{2}$ point pairs, recent follow-up works \cite{Sun-arXiv2021,Sun-arXiv2021b,Li-ISPRS-J-PRS2020,Shi-arXiv2020v2,Lusk-arXiv2020v2} that use such a graph entail $O(\ell^2)$ time complexity. While all these methods are more accurate than scalable, the following two are on the other side. 
$\FGR$ \cite{Zhou-ECCV2016} combines \textit{graduated non-convexity} ($\GNC$) and alternating minimization, while $\GNCTLS$ \cite{Yang-RA-L2020} combines \textit{truncated least squares}, iteratively reweighted least-squares, and $\GNC$. Both of them scale gracefully with $\ell$, while being robust against up to  $80/100=80\%$ outliers. 

Is such accuracy versus scalability dilemma of an inherent nature of the problems here, or can we escape from it?

\section{ARCS: Accuracy \& Scalability}\label{section:I2}
\myparagraph{Basic Idea} Although perhaps not explicitly mentioned in the literature, it should be known that there is a simple algorithm that solves Problem \ref{problem:SRSC} under the assumptions of Theorem \ref{theorem:I2}. This algorithm first computes the $\ell_2$ norm of each point in $\cQ$ and $\cP$ and the difference $d_{i,j}:=\norm{\bq_i}-\norm{\bp_j}$. Since $\cQ$ and $\cP$ are in general position (Remark \ref{remark:generalposition}), we have that $(\bq_i,\bp_j)$ is an inlier pair if and only if $d_{i,j}=0$. Based on the $d_{i,j}$'s, extract all such inlier pairs. Since $k^*\geq 2$, and by the general position assumption (Remark \ref{remark:generalposition}), there exist two inlier pairs say $(\bq_1,\bp_1),(\bq_2,\bp_2)$ such that $\bq_1$ and $\bq_2$ are not parallel. As a result and as it has been well-known since the $1980$'s \cite{Horn-JOSAA1987,Horn-JOSAA1988,Markley-JOSA1988,Arun-TPAMI1987}, if not even earlier \cite{Wahba-SIAM-Review-1965, Schonemann-1966},  $\bR^*$ can be determined from the two inlier pairs by SVD.


\myparagraph{ARCS: Efficient Implementation} Not all the $d_{i,j}$'s should be computed in order to find the correspondence set $\cC^*$, meaning that the otherwise $O(mn)$ time complexity can be reduced. Our $\ARCS$ Algorithm \ref{algo:I2} seeks all point pairs $(\bq_i,\bp_j)$'s whose norms are close, \ie, they satisfy $|d_{i,j}|\leq c$, for some sufficiently small $c\geq0$. Here $c$ is provided as an input of $\ARCS$ and  set as $0$ in the current context. It is clear that, under the general position assumption of Theorem \ref{theorem:I2}, the set $\overline{\cC}$ returned by $\ARCS$ is exactly the ground-truth correspondence set $\cC^*$. It is also clear that $\ARCS$ takes $O(m\log m)$ time and $O(m)$ space (recall $m\geq n\geq |\cC^*|$). 
\begin{algorithm}
	\SetAlgoLined
	Input: $\cQ=\{\bq_i\}_{i=1}^m, \cP=\{\bp_j\}_{j=1}^n,c\geq 0$;
	
	Sort $\cQ$ so that  (w.l.o.g.) $\norm{\bq_1}\leq\cdots\leq \norm{\bq_m}$; 
	
	Sort $\cP$ so that  (w.l.o.g.) $\norm{\bp_1}\leq\cdots\leq \norm{\bp_n}$;
	
	$i=1$; $j=1$; $\overline{\cC} = \varnothing$;
	
	\While{$i\leq m$ and $j\leq n$}{ 
		$d_{i,j}\gets \norm{\bq_i}-\norm{\bp_j}$;
		
		\If{$d_{i,j}> c$}{$j \gets j + 1$;}
		\If{$d_{i,j}< -c$}{$i \gets i + 1$;}
		
		\If{$-c \leq d_{i,j}\leq c$}{
			$\overline{\cC}\gets\overline{\cC}\cup (i,j)$; $(i,j)\gets (i+1,j+1)$;
		}
	}
	return $\overline{\cC}$;
	\caption{$\ARCS$ } \label{algo:I2}
\end{algorithm}
\begin{table}
	\centering
	\caption{Time (msec) of generating noiseless Gaussian point sets (G) and solving Problem \ref{problem:SRSC} by $\ARCS$ ($100$ trials, $k^*=2$). \label{table:I2}} 
	\begin{tabular}{cccc}  
		\toprule
		$m$ & $10^4$ & $10^5$ & $10^6$\\
		$n$ &$8\times 10^3$ & $8\times 10^4$& $8\times 10^5$\\
		\midrule
		G &  $5.9$ & $15.0$ & $212.8$ \\
		Brute Force & $73.8$ & $8304$ & $8380441.5$ \\ 
		$\ARCS$  & $1.51$ & $8.4$ & $121.1$ \\
		\bottomrule
	\end{tabular}
\end{table}

We proved Theorem \ref{theorem:I2}. It is operating in the noiseless case that allows us to show that Problem \ref{problem:SRSC} can be solved accurately and at large scale. Indeed, $\ARCS$ can handle more than $10^6$ points with $k^*=2$ in about $0.1$ seconds, even though generating those points has taken more than $0.2$ seconds, as shown in Table \ref{table:I2}.\footnote{For experiments in Tables \ref{table:I2} and \ref{table:l-mn} we generate data as per Section \ref{subsection:experiments-synthetic}.} Note that in the setting of Table \ref{table:I2} we have only $k^*=2$ overlapping points, a situation where all prior methods mentioned in \S \ref{section:intro} and \S \ref{subsection:prior-art}, if directly applicable, in principle break down. One reason is that they are not designed to handle the noiseless case. The other reason is that the overlapping ratio $k^*/m$ of Table \ref{table:I2} is the minimum possible. While the achievement in Table \ref{table:I2} is currently limited to the noiseless case, it forms a strong motivation that urges us to robustify $\ARCS$ to noise, while keeping as much of its accuracy and scalability as possible. Such robustification is the main theme of the next section.

\section{ARCS+: Robustifying ARCS to Noise}
Here we consider Problem \ref{problem:SRSC} with noise $\beps_{i,j}$. We will illustrate our algorithmic ideas by assuming  $\beps_{i,j}\sim \cN(0,\sigma^2\bI_3)$, although this is not necessary for actual implementation. As indicated in \S \ref{section:intro}, $\ARCSplus$ has three steps. We introduce them respectively in the next three subsections.

\subsection{Step 1: Finding Correspondences Under Noise}
A simple probability fact is $\norm{\bq_i - \bR^* \bp_j}\leq 5.54\sigma$ for any inlier $(\bq_i,\bp_j)$, so $|d_{i,j}|\leq 5.54\sigma$ with probability at least $1-10^{-6}$ (see, \eg, \cite{Yang-T-R2021}). To establish correspondences under noise, we need to modify\footnote{The details of this modification can be found at: \url{https://github.com/liangzu/ARCS/blob/main/ARCSplus_N.m}} the while loop of Algorithm \ref{algo:I2}, such that, in $O(\ell+m\log m)$ time, it returns the set $\overline{\cC}$ of all correspondences of size $\ell$ where each  $(i,j)\in \overline{\cC}$ satisfies $|d_{i,j}|\leq c$, with $c$ now set to $5.54\sigma$. Note that, to store the output correspondences, we need an extra $O(\ell)$ time, which can not be simply ignored as $\ell$ is in general larger than $m$ in the presence of noise (Table \ref{table:l-mn}). We call this modified version $\ARCSplus_\texttt{N}$. $\ARCSplus_\texttt{N}$ gives a set $\overline{\cC}$ that contains all inlier correspondences $\cC^*$ with probability at least $(1-10^{-6})^{k^*}$. This probability is larger than $99.9\%$ if $k^*\leq 10^3$, or larger than $99\%$ if $k^*\leq 10^4$. 



\begin{remark}[\textit{feature matching versus all-to-all correspondences versus $\ARCSplus_\texttt{N}$}]\label{remark:I2-featmatching}
	Feature matching methods provide fewer than $n$ hypothetical correspondences and thus speed up the subsequent computation, but they might give no inliers. Using all-to-all correspondences preserves all inliers, but a naive computation needs $O(mn)$ time and leads to a large-scale problem with extreme outlier rates. $\ARCSplus_\texttt{N}$ strikes a balance by delivering in $O(\ell + m\log m)$ time  a candidate correspondence set $\overline{\cC}$ of size $\ell$ containing all inliers with high probability and with $\ell\ll mn$.
\end{remark}
For illustration, Table \ref{table:l-mn} reports the number $\ell$ of correspondences that $\ARCSplus_\texttt{N}$ typically yields. As shown, even though $\ell/(mn)$ is usually smaller than $5\%$, yet $\ell$ itself could be very large, and the inlier ratio $k^*/\ell$ is extremely small (\eg, $\leq 0.05\%$). This is perhaps the best we could do for the current stage, because for now we only considered every point pair individually, while any pair $(\bq_i,\bp_i)$ is a potential inlier if it satisfies the necessary (but no longer sufficient) condition $|d_{i,j}|\leq c$. On the other hand, collectively analyzing the remaining point pairs allows to further remove outliers, and this is the major task of our next stage (\S \ref{section:STABBER3D}).

\begin{table}
	\centering
	\caption{The number $\ell$ of candidate correspondences produced by $\ARCSplus_\texttt{N}$ on synthetic noisy Gaussian point sets. A single trial.} \label{table:l-mn}
	\begin{tabular}{rccc}  
		\toprule
		$m$ & $1000$ & $5000$ & $10000$\\
		$n$ & $800$ & $4000$ & $8000$\\
		$k^*$ & $200$ & $1000$ & $2000$ \\
		\midrule
		$\ell$ & $36622$ & $931208$ & $3762888$\\
		$\ell/(mn)$  & $4.58\%$ & $4.66\%$ & $4.70\%$ \\
		\bottomrule
	\end{tabular}
\end{table}

\subsection{Step 2: Outlier Removal}\label{section:STABBER3D}
Let there be some correspondences given, by, \eg, either $\ARCSplus_\texttt{N}$ or feature matching (cf. Remark \ref{remark:I2-featmatching}). Then we arrive at an important special case of Problem \ref{problem:SRSC}, called \textit{robust rotation search}. For convenience we formalize it below:
\begin{problem}{(\textit{robust rotation search})}\label{problem:RRS}
	Consider $\ell$ pairs of $3$D points $\{(\by_i,\bx_i)\}_{i=1}^\ell$, with each pair satisfying
	\begin{align}
		\by_i = \bR^* \bx_i + \bo_i + \beps_i.
	\end{align}
	Here  $\beps_{i}\sim \cN(0,\sigma^2\bI_3)$ is noise, $\bo_{i}=\bm{0}$ if $i\in\cI^*$ where $\cI^*\subset[\ell]$ is of size $k^*$, and if $i\notin \cI^*$ then $\bo_{i}$ is nonzero and arbitrary. The task is to find $\bR^*$ and $\cI^*$.
\end{problem}
The percentage of outliers in Problem \ref{problem:RRS} can be quite large (cf. Table \ref{table:l-mn}), so our second step $\ARCSplus_\texttt{O}$ here is to remove outliers. In \S \ref{subsection:interval stabbing}, we shortly review the interval stabbing problem, on which $\ARCSplus_\texttt{O}$ of \S \ref{subsection:STABBER3D} is based.

\subsubsection{Preliminaries: Interval Stabbing}\label{subsection:interval stabbing}
Consider a collection of subsets of $\bbR$, $\{\cJ_i\}_{i=1}^L$, where each $\cJ_i$ is an interval of the form $[a,b]$. In the interval stabbing problem, one needs to determine a point $\omega\in\bbR$ and a subset $\cI$ of $\{\cJ_i\}_{i=1}^L$, so that $\cI$ is a maximal subset whose intervals overlap at $\omega$. Formally, we need to solve 
\begin{align}\label{eq:interval-stabbing}
	&\max_{\cI\subset [L],\omega\in\bbR} \ \ \ \ \ \ \  |\cI| \\
	\textnormal{s.t.}\ \ \ \  &\ \   \omega\in \cJ_i,\ \ \forall i\in \cI \nonumber
\end{align}
For this purpose, the following result is known.
\begin{lemma}[\textit{interval stabbing}]\label{lemma:interval_stabbing}
	Problem \eqref{eq:interval-stabbing} can be solved in $O(L \log L)$ time and $O(L)$ space.
\end{lemma}
Actually, the interval stabbing problem can be solved using sophisticated data structures such as \textit{interval tree} \cite{Berg-1997} or \textit{interval skip list} \cite{Hanson-1991}. On the other hand, it is a basic exercise to find an algorithm that solves Problem \eqref{eq:interval-stabbing}, which, though also in $O(L\log L)$ time, involves only a sorting operation and a for loop (details are omitted, see, \eg, \cite{Cai-ISPRS-J-PRS2019}). Finally, note that the use of interval stabbing for robust rotation search is not novel, and can be found in $\GORE$ \cite{Bustos-ICCV2015,Bustos-TPAMI2018}. However, as the reader might realize after \S \ref{subsection:STABBER3D}, our use of interval stabbing is quite different from $\GORE$.

\subsubsection{The Outlier Removal Algorithm}\label{subsection:STABBER3D}
We now consider the following consensus maximization:
\begin{equation}
	\begin{split} \label{eq:R-3D}
		&\max_{\cI\subset [\ell],\bR \in\SO(3)} \ \ \ \ \ \ \  |\cI| \\
		\textnormal{s.t.}\ \ \ \  &\ \   \norm{\by_i-\bR\bx_i}\leq c,\ \ \forall i\in \cI .
	\end{split}
\end{equation}
It has been shown in \cite{Chin-IJCV2020} that for the very related \textit{robust fitting} problem, such consensus maximization is in general NP-hard\footnote{Interestingly, consensus maximization over $\SO(2)$, \ie, the $2$D version of \eqref{eq:R-3D}, can be solved in $O(\ell\log\ell)$ time; see  \cite{Cai-ISPRS-J-PRS2019}.}. Thus it seems only prudent to switch our computational goal from solving \eqref{eq:R-3D} exactly to approximately. 

\myparagraph{From $\SO(3)$ to $\bbS^2$}  Towards this goal, we first shift our attention to $\bbS^2$ where the rotation axis $\bm{b}^*$ of $\bR^*$ lives. An interesting observation is that the axis $\bm{b}^*$ has the following interplay with data, independent of the rotation angle of $\bR^*$.
\begin{prop}\label{prop:axis}
	Let $\bv_i:=\by_i-\bx_i$. Recall $\beps_{i}\sim \cN(0,\sigma^2\bI_3)$. If $(\by_i,\bx_i)$ is an inlier pair, then $\trsp{\bv_i}\bm{b}^*\sim \cN(0,\sigma^2)$, and so $|\trsp{\bv_i}\bm{b}^*|\leq 4.9\sigma$ with probability at least $1-10^{-6}$.
\end{prop}
Proposition \ref{prop:axis} (cf. Appendix \ref{appendix:prop-axis}) leads us to Problem \eqref{eq:max-consensus-b}:
\begin{equation}\label{eq:max-consensus-b}
	\begin{split}
		&\max_{\cI\subset [\ell],\bm{b} \in\bbS^2} \ \ \ \ \ \ \  |\cI| \\
		\textnormal{s.t.}\ \ \ \  &\ \   |\trsp{\bv_i}\bm{b}|\leq \bar{c},\ \ \forall i\in \cI \\
		&\ \ \ \ \ \ \ \ b_2 \geq 0. 
	\end{split}
\end{equation}
In \eqref{eq:max-consensus-b} the constraint on the second entry $b_2$ of $\bm{b}$ is to eliminate the symmetry, and Proposition \ref{prop:axis} suggests to set $\bar{c}:=4.9\sigma$. Problem \eqref{eq:max-consensus-b} is easier than \eqref{eq:R-3D} as it has fewer degrees of freedom; see also \cite{Bustos-ICCV2015} where a different reduction to a 2 DoF (sub-)problem was derived for $\GORE$. 

Solving \eqref{eq:max-consensus-b} is expected to yield an accurate estimate of $\bm{b^*}$, from which the rotation angle can later be estimated. Problem \eqref{eq:max-consensus-b} reads: find a plane (defined by the normal $\bm{b}$) that approximately contains as much points $\bv_i$'s as possible. This is an instance of the \textit{robust subspace learning} problem \cite{Tsakiris-JMLR2018,Zhu-NeurIPS2018,Zhu-NeurIPS2019,Lerman-IEEE2018,Ding-ICML2019,Ding-ICML2021,Yao-NeurIPS2021}, for which various scalable algorithms with strong theoretical guarantees have been developed in more tractable formulations (\eg, $\ell_1$ minimization) than consensus maximization. Most notably, the so-called \textit{dual principal component pursuit} formulation \cite{Tsakiris-JMLR2018} was proved in \cite{Zhu-NeurIPS2018} to be able to tolerate $O\big((k^*)^2\big)$ outliers. Still, all these methods can not handle as many outliers as we currently have (cf. Table \ref{table:l-mn}), even though they can often minimize their objective functions to global optimality.

\myparagraph{From $\bbS^2$ to $[0,\pi]$} We can further ``reduce'' the degrees of freedom in \eqref{eq:max-consensus-b} by $1$, through the following lens. Certainly $\bm{b}\in\bbS^2$ in \eqref{eq:max-consensus-b} is determined by two angles $\theta\in[0,\pi]$, $\phi\in[0,\pi]$. Now consider the following problem:
\begin{equation}\label{eq:max-consensus-theta}
	\begin{split}
		&\max_{\cI\subset [\ell],\theta\in[0,\pi]} \ \ \ \ \ \ \  |\cI| \\
		\textnormal{s.t.}\ \ \ \  &\ \   |\trsp{\bv_i}\bm{b}|\leq \bar{c},\ \ \forall i\in \cI \\
		&\  \bm{b}=\trsp{[\sin(\theta)\cos(\phi),\ \sin(\theta)\sin(\phi),\ \cos(\theta)]} .
	\end{split}
\end{equation}
Problem \eqref{eq:max-consensus-theta} is a simplified version of \eqref{eq:max-consensus-b} with $\phi$ given. Clearly, to solve \eqref{eq:max-consensus-b} it suffices to minimize the function $f:[0,\pi]\to\bbR$ which maps any $\phi_0\in [0,\pi]$ to the objective value of \eqref{eq:max-consensus-theta} with $\phi=\phi_0$. Moreover, we have:
\begin{prop}\label{prop:f}
	Problem \eqref{eq:max-consensus-theta} can be solved in $O(\ell\log \ell)$ time and $O(\ell)$ space via interval stabbing.  
\end{prop}
Proposition \ref{prop:f} gives an $O(\ell\log \ell)$ time oracle to access the values of $f$. Since computing the objective value of \eqref{eq:max-consensus-b} given $\theta,\phi$ already needs $O(\ell)$ time, the extra cost of the logarithmic factor in Proposition \ref{prop:f} is nearly negligible. Since $f$ has only one degree of freedom, its global minimizer can be found by \textit{one-dimensional branch $\&$ bound} \cite{Jiao-IROS2020}. But this entails exponential time complexity in the worst case, a situation we wish to sidestep. Alternatively, the search space $[0,\pi]$ is now so small that the following  algorithm $\ARCSplus_\texttt{O}$ turns out to be surprisingly efficient and robust: i) \textit{sampling} from $[0,\pi]$, ii) \textit{stabbing} in $\bbS^2$, and iii) \textit{stabbing} in $\SO(3)$.

\myparagraph{Sampling from $[0,\pi]$} Take $s$ equally spaced points $\phi_j=(2j-1)\pi/(2s)$, $\forall j\in [s]$, on $[0,\pi]$. The reader may find this choice of $\phi_j$'s similar to the \textit{uniform grid approach} \cite{Nesterov-2018}; in the latter Nesterov commented that ``the reason why it works here is related to the \textit{dimension} of the problem''.

\myparagraph{Stabbing in $\bbS^2$} For each $j\in[s]$, solve \eqref{eq:max-consensus-theta} with $\phi=\phi_j$ to get $s$ candidate consensus set $\cI_j$'s and $s$ angles $\theta_j$'s. 
From each $\phi_j$ and $\theta_j$ we obtain a candidate rotation axis $\bm{b}_j$.

\myparagraph{Stabbing in $\SO(3)$} Since now we have estimates of rotation axes, $\bm{b}_j$'s, there is one degree of freedom remaining, the rotation angle $\omega$. For this we consider:
\begin{equation} \label{eq:R-angle}
	\begin{split}
		&\max_{\cI\subset [\ell],\omega \in[0,2\pi]} \ \ \ \ \ \ \  |\cI| \\
		\textnormal{s.t.}\ \ \ \  &\ \   \norm{\by_i-\bR\bx_i}\leq c,\ \ \forall i\in \cI  \\
		\bR&=\bm{b}\trsp{\bm{b}}+[\ \bm{b}\ ]_{\times}\sin(\omega)+ (\bI_3-\bm{b}\trsp{\bm{b}})\cos(\omega) 
	\end{split}
\end{equation}
Here $[\ \bm{b}\ ]_{\times}\in\bbR^{3\times 3}$ denotes the matrix generating the cross product $\times$ by $\bm{b}$, that is $[\ \bm{b}\ ]_{\times} \ba = \bm{b}\times \ba$ for all $\ba\in\bbR^3$. Similarly to Proposition \ref{prop:f}, we have the following result:
\begin{prop}\label{lemma:g}
	Problem \eqref{eq:R-angle} can be solved in $O(\ell\log \ell)$ time and $O(\ell)$ space via interval stabbing.
\end{prop}

\begin{figure}
	\centering
	\subfloat[$\ARCSplus_\texttt{O}$]{\input{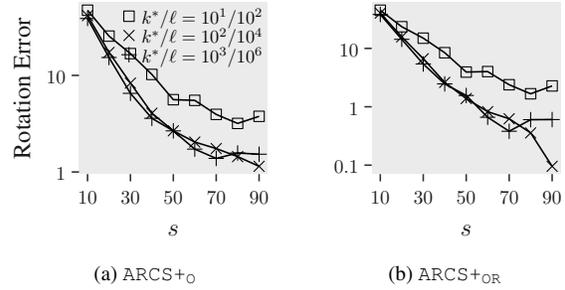} \label{fig:choice_s_10} }
	\subfloat[$\ARCSplus_{\texttt{OR}}$]{
\begin{tikzpicture}[x=1pt,y=1pt]
\definecolor{fillColor}{RGB}{255,255,255}
\path[use as bounding box,fill=fillColor,fill opacity=0.00] (0,0) rectangle (103.35,101.18);
\begin{scope}
\path[clip] (  0.00,  0.00) rectangle (103.35,101.18);
\definecolor{drawColor}{RGB}{255,255,255}
\definecolor{fillColor}{RGB}{255,255,255}

\path[draw=drawColor,line width= 0.6pt,line join=round,line cap=round,fill=fillColor] (  0.00,  0.00) rectangle (103.35,101.18);
\end{scope}
\begin{scope}
\path[clip] ( 26.35, 30.69) rectangle ( 97.85, 95.68);
\definecolor{fillColor}{gray}{0.92}

\path[fill=fillColor] ( 26.35, 30.69) rectangle ( 97.85, 95.68);
\definecolor{drawColor}{RGB}{0,0,0}

\path[draw=drawColor,line width= 0.6pt,line join=round] ( 29.60, 92.72) --
	( 37.72, 86.48) --
	( 45.85, 82.15) --
	( 53.97, 76.64) --
	( 62.10, 69.33) --
	( 70.22, 69.50) --
	( 78.35, 64.52) --
	( 86.47, 61.10) --
	( 94.60, 64.07);

\path[draw=drawColor,line width= 0.6pt,line join=round] ( 29.60, 91.64) --
	( 37.72, 82.65) --
	( 45.85, 74.38) --
	( 53.97, 65.42) --
	( 62.10, 59.30) --
	( 70.22, 54.33) --
	( 78.35, 51.65) --
	( 86.47, 46.30) --
	( 94.60, 33.64);

\path[draw=drawColor,line width= 0.6pt,line join=round] ( 29.60, 91.12) --
	( 37.72, 81.72) --
	( 45.85, 72.38) --
	( 53.97, 64.84) --
	( 62.10, 60.46) --
	( 70.22, 52.21) --
	( 78.35, 46.92) --
	( 86.47, 51.30) --
	( 94.60, 51.34);

\path[draw=drawColor,line width= 0.4pt,line join=round,line cap=round] ( 27.64, 90.76) rectangle ( 31.56, 94.69);

\path[draw=drawColor,line width= 0.4pt,line join=round,line cap=round] ( 35.76, 84.52) rectangle ( 39.68, 88.44);

\path[draw=drawColor,line width= 0.4pt,line join=round,line cap=round] ( 43.89, 80.18) rectangle ( 47.81, 84.11);

\path[draw=drawColor,line width= 0.4pt,line join=round,line cap=round] ( 52.01, 74.68) rectangle ( 55.93, 78.61);

\path[draw=drawColor,line width= 0.4pt,line join=round,line cap=round] ( 60.13, 67.36) rectangle ( 64.06, 71.29);

\path[draw=drawColor,line width= 0.4pt,line join=round,line cap=round] ( 68.26, 67.54) rectangle ( 72.18, 71.46);

\path[draw=drawColor,line width= 0.4pt,line join=round,line cap=round] ( 76.38, 62.56) rectangle ( 80.31, 66.48);

\path[draw=drawColor,line width= 0.4pt,line join=round,line cap=round] ( 84.51, 59.14) rectangle ( 88.43, 63.07);

\path[draw=drawColor,line width= 0.4pt,line join=round,line cap=round] ( 92.63, 62.11) rectangle ( 96.56, 66.03);

\path[draw=drawColor,line width= 0.4pt,line join=round,line cap=round] ( 27.64, 89.68) -- ( 31.56, 93.60);

\path[draw=drawColor,line width= 0.4pt,line join=round,line cap=round] ( 27.64, 93.60) -- ( 31.56, 89.68);

\path[draw=drawColor,line width= 0.4pt,line join=round,line cap=round] ( 35.76, 80.68) -- ( 39.68, 84.61);

\path[draw=drawColor,line width= 0.4pt,line join=round,line cap=round] ( 35.76, 84.61) -- ( 39.68, 80.68);

\path[draw=drawColor,line width= 0.4pt,line join=round,line cap=round] ( 43.89, 72.41) -- ( 47.81, 76.34);

\path[draw=drawColor,line width= 0.4pt,line join=round,line cap=round] ( 43.89, 76.34) -- ( 47.81, 72.41);

\path[draw=drawColor,line width= 0.4pt,line join=round,line cap=round] ( 52.01, 63.45) -- ( 55.93, 67.38);

\path[draw=drawColor,line width= 0.4pt,line join=round,line cap=round] ( 52.01, 67.38) -- ( 55.93, 63.45);

\path[draw=drawColor,line width= 0.4pt,line join=round,line cap=round] ( 60.13, 57.34) -- ( 64.06, 61.27);

\path[draw=drawColor,line width= 0.4pt,line join=round,line cap=round] ( 60.13, 61.27) -- ( 64.06, 57.34);

\path[draw=drawColor,line width= 0.4pt,line join=round,line cap=round] ( 68.26, 52.37) -- ( 72.18, 56.30);

\path[draw=drawColor,line width= 0.4pt,line join=round,line cap=round] ( 68.26, 56.30) -- ( 72.18, 52.37);

\path[draw=drawColor,line width= 0.4pt,line join=round,line cap=round] ( 76.38, 49.68) -- ( 80.31, 53.61);

\path[draw=drawColor,line width= 0.4pt,line join=round,line cap=round] ( 76.38, 53.61) -- ( 80.31, 49.68);

\path[draw=drawColor,line width= 0.4pt,line join=round,line cap=round] ( 84.51, 44.33) -- ( 88.43, 48.26);

\path[draw=drawColor,line width= 0.4pt,line join=round,line cap=round] ( 84.51, 48.26) -- ( 88.43, 44.33);

\path[draw=drawColor,line width= 0.4pt,line join=round,line cap=round] ( 92.63, 31.68) -- ( 96.56, 35.60);

\path[draw=drawColor,line width= 0.4pt,line join=round,line cap=round] ( 92.63, 35.60) -- ( 96.56, 31.68);

\path[draw=drawColor,line width= 0.4pt,line join=round,line cap=round] ( 26.82, 91.12) -- ( 32.37, 91.12);

\path[draw=drawColor,line width= 0.4pt,line join=round,line cap=round] ( 29.60, 88.35) -- ( 29.60, 93.90);

\path[draw=drawColor,line width= 0.4pt,line join=round,line cap=round] ( 34.95, 81.72) -- ( 40.50, 81.72);

\path[draw=drawColor,line width= 0.4pt,line join=round,line cap=round] ( 37.72, 78.95) -- ( 37.72, 84.50);

\path[draw=drawColor,line width= 0.4pt,line join=round,line cap=round] ( 43.07, 72.38) -- ( 48.62, 72.38);

\path[draw=drawColor,line width= 0.4pt,line join=round,line cap=round] ( 45.85, 69.61) -- ( 45.85, 75.16);

\path[draw=drawColor,line width= 0.4pt,line join=round,line cap=round] ( 51.20, 64.84) -- ( 56.75, 64.84);

\path[draw=drawColor,line width= 0.4pt,line join=round,line cap=round] ( 53.97, 62.07) -- ( 53.97, 67.62);

\path[draw=drawColor,line width= 0.4pt,line join=round,line cap=round] ( 59.32, 60.46) -- ( 64.87, 60.46);

\path[draw=drawColor,line width= 0.4pt,line join=round,line cap=round] ( 62.10, 57.69) -- ( 62.10, 63.24);

\path[draw=drawColor,line width= 0.4pt,line join=round,line cap=round] ( 67.45, 52.21) -- ( 73.00, 52.21);

\path[draw=drawColor,line width= 0.4pt,line join=round,line cap=round] ( 70.22, 49.44) -- ( 70.22, 54.99);

\path[draw=drawColor,line width= 0.4pt,line join=round,line cap=round] ( 75.57, 46.92) -- ( 81.12, 46.92);

\path[draw=drawColor,line width= 0.4pt,line join=round,line cap=round] ( 78.35, 44.14) -- ( 78.35, 49.69);

\path[draw=drawColor,line width= 0.4pt,line join=round,line cap=round] ( 83.70, 51.30) -- ( 89.25, 51.30);

\path[draw=drawColor,line width= 0.4pt,line join=round,line cap=round] ( 86.47, 48.52) -- ( 86.47, 54.07);

\path[draw=drawColor,line width= 0.4pt,line join=round,line cap=round] ( 91.82, 51.34) -- ( 97.37, 51.34);

\path[draw=drawColor,line width= 0.4pt,line join=round,line cap=round] ( 94.60, 48.56) -- ( 94.60, 54.11);
\end{scope}
\begin{scope}
\path[clip] (  0.00,  0.00) rectangle (103.35,101.18);
\definecolor{drawColor}{gray}{0.10}

\node[text=drawColor,anchor=base east,inner sep=0pt, outer sep=0pt, scale=  0.73] at ( 21.40, 31.05) {$0.1$};

\node[text=drawColor,anchor=base east,inner sep=0pt, outer sep=0pt, scale=  0.73] at ( 21.40, 53.15) {$1$};

\node[text=drawColor,anchor=base east,inner sep=0pt, outer sep=0pt, scale=  0.73] at ( 21.40, 75.25) {$10$};
\end{scope}
\begin{scope}
\path[clip] (  0.00,  0.00) rectangle (103.35,101.18);
\definecolor{drawColor}{gray}{0.20}

\path[draw=drawColor,line width= 0.6pt,line join=round] ( 23.60, 34.08) --
	( 26.35, 34.08);

\path[draw=drawColor,line width= 0.6pt,line join=round] ( 23.60, 56.18) --
	( 26.35, 56.18);

\path[draw=drawColor,line width= 0.6pt,line join=round] ( 23.60, 78.28) --
	( 26.35, 78.28);
\end{scope}
\begin{scope}
\path[clip] (  0.00,  0.00) rectangle (103.35,101.18);
\definecolor{drawColor}{gray}{0.20}

\path[draw=drawColor,line width= 0.6pt,line join=round] ( 29.60, 27.94) --
	( 29.60, 30.69);

\path[draw=drawColor,line width= 0.6pt,line join=round] ( 45.85, 27.94) --
	( 45.85, 30.69);

\path[draw=drawColor,line width= 0.6pt,line join=round] ( 62.10, 27.94) --
	( 62.10, 30.69);

\path[draw=drawColor,line width= 0.6pt,line join=round] ( 78.35, 27.94) --
	( 78.35, 30.69);

\path[draw=drawColor,line width= 0.6pt,line join=round] ( 94.60, 27.94) --
	( 94.60, 30.69);
\end{scope}
\begin{scope}
\path[clip] (  0.00,  0.00) rectangle (103.35,101.18);
\definecolor{drawColor}{gray}{0.10}

\node[text=drawColor,anchor=base,inner sep=0pt, outer sep=0pt, scale=  0.73] at ( 29.60, 19.68) {$10$};

\node[text=drawColor,anchor=base,inner sep=0pt, outer sep=0pt, scale=  0.73] at ( 45.85, 19.68) {$30$};

\node[text=drawColor,anchor=base,inner sep=0pt, outer sep=0pt, scale=  0.73] at ( 62.10, 19.68) {$50$};

\node[text=drawColor,anchor=base,inner sep=0pt, outer sep=0pt, scale=  0.73] at ( 78.35, 19.68) {$70$};

\node[text=drawColor,anchor=base,inner sep=0pt, outer sep=0pt, scale=  0.73] at ( 94.60, 19.68) {$90$};
\end{scope}
\begin{scope}
\path[clip] (  0.00,  0.00) rectangle (103.35,101.18);
\definecolor{drawColor}{gray}{0.10}

\node[text=drawColor,anchor=base,inner sep=0pt, outer sep=0pt, scale=  0.92] at ( 62.10,  7.64) {$s$};
\end{scope}
\end{tikzpicture} \label{fig:choice_s} }
	\caption{Rotation errors (in degrees) of steps 2 and 3 for robust rotation search methods with $s$ varying ($500$ trials, $\sigma=0.01$). \label{fig:choice_of_s}}
\end{figure}

After solving \eqref{eq:R-angle} with $\bm{b}=\bm{b}_j$ for each $j\in[s]$, we obtain $s$ candidate consensus sets $\tilde{\cI}_1,\dots,\tilde{\cI}_s$, and we choose the one with maximal cardinality as an approximate solution to \eqref{eq:R-3D}.
Finally, notice that the time complexity $O(s\ell \log \ell)$ of $\ARCSplus_\texttt{O}$ depends on the hyper-parameter $s$. We set $s=90$ as an invariant choice, as suggested by Figure \ref{fig:choice_of_s}. 

This output consensus set $\tilde{\cI}$ typically has very few outliers; see Table \ref{table:I-tilde}. Thus it will be used next in $\ARCSplus_\texttt{R}$, our final step for accurately estimating the rotation (\S \ref{section:R}).
\begin{table}[h]
	\centering
	\caption{The output of $\ARCSplus_\texttt{O}$ with inputs from Table \ref{table:l-mn}.} \label{table:I-tilde}
	\begin{tabular}{rccc}  
		\toprule
		Input Inlier Ratio & $\frac{200}{36622}$ & $\frac{1000}{931208}$ & $\frac{2000}{3762888}$ \\
		\midrule
		Output Inlier Ratio & $\frac{199}{213}$ & $\frac{993}{1314}$ &  $\frac{1951}{3184}$ \\
		\bottomrule
	\end{tabular}
\end{table}

\subsection{Step 3: Rotation Estimation}\label{section:R}
The final step $\ARCSplus_\texttt{R}$ of $\ARCSplus$ is a refinement procedure that performs robust rotation search on the output correspondences $\tilde{\cI}$ of $\ARCSplus_\texttt{O}$. Since $\tilde{\cI}$ contains much fewer outlier correspondences than we previously had (cf. Table \ref{table:l-mn} and \ref{table:I-tilde}), in what follows we simplify the notations by focusing on the point set $\{(\by_i,\bx_i)\}_{i\in[\ell]}$, which we assume has few outliers (say $\leq 50\%$). Then, a natural formulation is
\begin{align}\label{eq:L1-SO3}
	\min_{\bR\in\SO(3)} \sum_{i=1}^{\ell}\norm{\by_i-\bR\bx_i}.
\end{align}
Problem \eqref{eq:L1-SO3} appears easier to solve than consensus maximization \eqref{eq:R-3D}, as it has a convex objective function at least. Next we present the $\ARCSplus_\texttt{R}$ algorithm and its theory.

\noindent\textbf{Algorithm.} We start with the following equivalence. 

\begin{prop}\label{prop:SO3toDPCP}
	We have $\trsp{\bw}\bD_i\bw=\norm{\by_i - \bR \bx_i}^2$, where $\bw\in\bbS^3$ is a quaternion representation of $\bR$ of $\eqref{eq:L1-SO3}$, and $\bD_i\in\bbR^{4\times 4}$ is a positive semi-definite matrix whose entries depend on $\bx_i$, $\by_i$. So	Problem \eqref{eq:L1-SO3} is equivalent to
	\begin{align}\label{eq:L1-S3}
		\min_{\bw\in\bbS^3} h(\bw),\ \ \ \ h(\bw) = \sum_{i=1}^\ell \sqrt{\trsp{\bw}\bD_i\bw}.
	\end{align}
\end{prop}
The exact relation between unit quaternions and rotations is reviewed in Appendix \ref{appendix:proof-prop-SO3toS3}, where Proposition \ref{prop:SO3toDPCP} is proved and the expression of $\bD_i$ is given. For what follows, it suffices to know that a unit quaternion is simply a unit vector of $\bbR^4$, and that the space of unit quaternions is $\bbS^3$. 

Note that the objective $h$ of \eqref{eq:L1-S3} is convex, while both problems \eqref{eq:L1-SO3} and \eqref{eq:L1-S3} are nonconvex (due to the constraint) and nonsmooth (due to the objective). Though   \eqref{eq:L1-SO3} and \eqref{eq:L1-S3} are equivalent, the advantage of \eqref{eq:L1-S3} will manifest itself soon. Before that, we first introduce the $\ARCSplus_\texttt{R}$ algorithm for solving \eqref{eq:L1-S3}. $\ARCSplus_\texttt{R}$ falls into the general \textit{Riemannian subgradient descent} framework (see, \eg, \cite{Li-SIAM-J-O2021}). It is initialized at some unit quaternion $\bw^{(0)}\in\bbS^3$ and proceeds by
\begin{align}\label{eq:R-update}
	\bw^{(t+1)}\gets \Proj_{\bbS^3}\big(\bw^{(t)}-\gamma^{(t)}\rsgrad h(\bw^{(t)})\big),
\end{align}
where $\Proj_{\bbS^3}(\cdot)$ projects a vector onto $\bbS^3$, $\gamma^{(t)}$ is some stepsize, $\rsgrad h(\bw^{(t)})$ is a \textit{Riemannian subgradient}\footnote{We follow \cite{Li-SIAM-J-O2021} where a Riemannian subgradient $\rsgrad h(\bw)$ at $\bw\in\bbS^3$ is defined as the projection of some subgradient $\sgrad h(\bw)$ of $h$ at $\bw$ onto the tangent space of $\bbS^3$ at $\bw$, \ie, $\rsgrad h(\bw):=(\bI_4-\bw\trsp{\bw})\sgrad h (\bw)$. See \cite{Beck-OptBook2017} for how to compute a subgradient of some given function.} of $h$ at $\bw^{(t)}$.

\myparagraph{Theory} Now we are able to compare \eqref{eq:L1-SO3} and \eqref{eq:L1-S3} from a theoretical perspective. As proved in \cite{Bohorquez-arXiv2020v3}, for any fixed outlier ratio and $k^*> 0$, \textit{Riemannian subgradient descent} when applied to \eqref{eq:L1-SO3} with proper initialization converges to $\bR^*$ in finite time, as long as i) $\ell$ is sufficiently large, ii) all points $\by_i$'s and $\bx_i$'s are uniformly distributed on $\bbS^2$, iii) there is no noise. But in \cite{Bohorquez-arXiv2020v3} no convergence rate is given. One main challenge of establishing convergence rates there is that projecting on $\SO(3)$ does not enjoy a certain kind of \textit{nonexpansiveness} property, which is important for convergence analysis (cf. Lemma 1 of \cite{Li-SIAM-J-O2021}). On the other hand, projection onto $\bbS^3$ of \eqref{eq:L1-S3} does satisfy such property. As a result, we are able to provide convergence rate guarantees for $\ARCSplus_\texttt{R}$. For example, it follows directly from Theorem 2 of \cite{Li-SIAM-J-O2021} that $\ARCSplus_\texttt{R}$ \eqref{eq:R-update} converges to an $\varepsilon$-\textit{stationary} point in $O(\varepsilon^{-4})$ iterations, even if initialized arbitrarily.

We next give conditions for $\ARCSplus_\texttt{R}$ to converge linearly to the ground-truth unit quaternion $\pm \bw^*$ that represents $\bR^*$. Let the distance between a unit quaternion $\bw$ and $\pm\bw^{*}$ be
\begin{align*}
	\dist(\bw,\pm \bw^*):=\min \big\{\norm{\bw - \bw^*}, \norm{\bw + \bw^*} \big \}.
\end{align*}
If $\dist(\bw,\pm\bw^*)<\rho$ with $\rho>0$ then $\bw$ is called $\rho$-\textit{close} to $\pm\bw^*$. We need the following notion of \textit{sharpness}.
\begin{definition}[\textit{sharpness} \cite{Burke-SIAM-J-CO1993,Li-SIAM-J-O2011,Karkhaneei-JOTA2019,Li-SIAM-J-O2021}]
	We say that $\pm \bw^*$ is an $\alpha$-\textit{sharp minimum} of \eqref{eq:L1-S3} if $\alpha>0$ and if there exists a number $\rho_{\alpha}>0$ such that any unit quaternion $\bw\in\bbS^3$ that is $\rho_{\alpha}$-\textit{close} to $\pm\bw^*$ satisfies the inequality
	\begin{align}
		h(\bw) - h(\bw^*) \geq \alpha \dist(\bw,\pm\bw^*).
	\end{align}
\end{definition}
We provide a condition below for $\pm\bw^*$ to be $\alpha^*$-sharp: 
\begin{prop}\label{prop:sharpness}
	If $\alpha^*:=k^* \eta_{\min}/\sqrt{2}-(\ell-k^*)\eta_{\max}>0$ and if $\beps_{i}=\bm{0}$ in Problem \ref{problem:RRS}, then Problem \eqref{eq:L1-S3}  admits $\pm w^*$ as an $\alpha^*$-sharp minimum. Here $\eta_{\min}$, $\eta_{\max}$ are respectively
	\begin{align}
		\eta_{\min}&:=\frac{1}{k^*}\min_{\bw\in\cS^*\cap \bbS^3}\sum_{i\in\cI^*}\sqrt{\trsp{\bw}\bD_i\bw}, \text{\ \  and } \label{eq:etamin}\\
		\eta_{\max}&:= \frac{1}{\ell-k^*} \max_{\bw\in\bbS^3} \sum_{i\in [\ell]\backslash \cI^*}\sqrt{\trsp{\bw}\bD_i\bw}, \label{eq:etamax}
	\end{align}
	where $\cS^*$ is the hyperplane of $\bbR^4$ perpendicular to $\pm\bw^*$.
\end{prop}
Proposition \ref{prop:sharpness} is proved in Appendix \ref{appendix:proof-sharpness}.
The condition $\alpha^* > 0$ defines a relation between the number of inliers $(k^*)$ and outliers $(\ell-k^*)$, and involves two quantities $\eta_{\min}$ and $\eta_{\max}$ whose values depend on how $\bD_i$'s are distributed on the positive semi-definite cone. We offer probabilistic interpretations for $\eta_{\min}$ and $\eta_{\max}$ in Appendix \ref{appendix:probability-prop-sharpness}.

\begin{table*}[t]
	\centering
	\caption{Average errors in degrees $|$ standard deviation $|$ running times in seconds of various algorithms on synthetic data ($20$ trials). } \label{table:RRS-scalibility}
	\begin{tabular}{cccccc}  
		\toprule
		Inlier Ratio $\frac{k^*}{\ell}$ & $\frac{10^3}{10^5}=1\%$ & $\frac{10^3}{10^6}=0.1\%$  & $\frac{3\times 10^3}{5\times 10^6}=0.06\%$ & $\frac{3\times 10^3}{10^7}=0.03\%$ &  $\frac{10^3}{10^7}=0.01\%$  \\
		\midrule
		$\TEASER$ \cite{Yang-T-R2021} & out-of-memory &   &  &  & \\
		$\RANSAC$ & $0.39$ $|$ $0.20$ $|$ $29.1$ &  $\geq 8.4$ hours & & & \\
		$\GORE$ \cite{Bustos-ICCV2015,Bustos-TPAMI2018}  &$3.43$ $|$ $2.10$ $|$ $1698$  & $\geq 12$ hours &&&\\
		$\FGR$ \cite{Zhou-ECCV2016} & $52.2$ $|$ $68.5$  $|$ $3.64$  & $95.0$ $|$ $60.9$  $|$ $37.7$ &  $84.9$ $|$ $59.4$ $|$  $145$  & $86.5$ $|$ $56.9$ $|$ $311$ &   $97.3$ $|$ $61.3$ $|$ $314$ \\
		$\GNCTLS$ \cite{Yang-RA-L2020}& $3.86$ $|$ $9.51$ $|$ $0.13$ & $63.4$ $|$ $50.5$ $|$ $2.26$ & $49.9$ $|$ $31.1$ $|$ $15.9$  & $90.2$ $|$ $45.6$ $|$ $40.1$ &  $120$ $|$ $34.3$ $|$ $36.3$ \\
		\midrule
		$\ARCSplus_\texttt{R}$ &$9.92$ $|$ $13.1$ $|$ $0.12$ & $65.2$ $|$ $48.9$ $|$ $0.96$  & $55.6$ $|$ $38.3$  $|$ $5.58$ &  $88.4$ $|$ $36.2$ $|$ $12.6$  & $98.2$ $|$ $36.0$ $|$ $12.2$\\
		$\ARCSplus_\texttt{O}$&  $0.86$ $|$ $0.29$ $|$ $1.71$ & $0.99$ $|$ $0.37$ $|$ $23.2$  & $0.91$ $|$ $0.30$ $|$ $125$ & $0.98$ $|$ $0.42$  $|$ $287$  & $55.6$ $|$ $60.9$ $|$ $281$\\
		$\ARCSplus_{\texttt{OR}}$ & $0.03$ $|$ $0.03$ $|$ $1.72$ & $0.09$ $|$ $0.07$ $|$ $23.2$  & $0.11$ $|$ $0.07$ $|$ $125$ & $0.22$ $|$ $0.15$ $|$ $287$ & $55.4$ $|$ $60.1$ $|$ $281$ \\
		\bottomrule
	\end{tabular}
\end{table*}

With Theorem 4 of \cite{Li-SIAM-J-O2021} and Proposition \ref{prop:sharpness} we have that $\ARCSplus_\texttt{R}$ \eqref{eq:R-update}, if initialized properly and with suitable stepsizes, converges linearly to the ground-truth unit quaternion $\pm\bw^*$, as long as $\pm\bw^*$ is $\alpha^*$-sharp. A formal statement is:
\begin{theorem}\label{theorem:R-convergence}
	Suppose $\alpha^*:=k^* \eta_{\min}/\sqrt{2}-(\ell-k^*)\eta_{\max}>0$. Let $L_h$ be a Lipschitz constant of $h$. Run Riemannian subgradient descent $\ARCSplus_\texttt{R}$ \eqref{eq:R-update} with initialization  $\bw^{(0)}$ satisfying $\dist(\bw^{(0)},\pm \bw^*)\leq \min\{\alpha^*/L_h,\rho_{\alpha^*}\}$ and with geometrically diminishing stepsizes $\gamma^{(t)}=\beta^t\gamma^{(0)}$, where
	\begin{align*}
		\gamma^{(0)}&< \min\Bigg\{\frac{2e_0(\alpha^*-L_he_0)}{L_h^2},\frac{e_0}{2(\alpha^*-L_he_0)}\Bigg\},\\
		\beta^2&\in\Bigg[1+2\Big(L_h-\frac{\alpha^*}{e_0}\Big)\gamma^{(0)}+\frac{L_h^2(\gamma^{(0)})^2}{e_0^2},\ 1\Bigg),\\
		e_0&=\min\Bigg\{ \max\Big\{\dist(\bw^{(0)},\pm \bw^*), \frac{\alpha^*}{2L_h} \Big\},\rho_{\alpha^*} \Bigg\}.
	\end{align*}
	In the noiseless case ($\beps_{i}=0$) we have each $\bw^{(t)}$ satisfying
	\begin{align}
		\dist(\bw^{(t)},\pm \bw^*)\leq \beta^t e_0.
	\end{align}
\end{theorem}	
\begin{remark}[\textit{a posteriori optimality guarantees}]\label{remark:posteriori}
	Theorem \ref{theorem:R-convergence} endows $\ARCSplus_\texttt{R}$ \eqref{eq:R-update} with convergence guarantee. On the other hand, \textit{a posteriori} optimality guarantees can be obtained via semidefinite certification \cite{Bandeira-2016,Carlone-ToR2016,Yang-NeurIPS2020,Yang-T-R2021}.
\end{remark}

\section{Experiments}\label{section:experiments}
In this section we evaluate $\ARCSplus$ via synthetic and real experiments for Problem \ref{problem:SRSC}, simultaneous rotation and correspondence search. We also evaluate its components, namely $\ARCSplus_{\texttt{O}}$ (\S \ref{section:STABBER3D}) and $\ARCSplus_{\texttt{R}}$ (\S \ref{section:R}) for Problem \ref{problem:RRS}, robust rotation search, as it is a task of independent interest. For both of the two problems we compare the following state-of-the-art methods (reviewed in \S \ref{subsection:prior-art}): $\FGR$ \cite{Zhou-ECCV2016}, $\GORE$ \cite{Bustos-ICCV2015}, $\RANSAC$, $\GNCTLS$ \cite{Yang-RA-L2020}, and $\TEASER$ \cite{Yang-T-R2021}.

\subsection{Experiments on Synthetic Point Clouds}\label{subsection:experiments-synthetic}
\myparagraph{Setup} We set $\sigma=0.01$, $\bar{c}=c=5.54\sigma$, $n=\floor{0.8m}$, and $s=90$ unless otherwise specified. For all other methods we used default or otherwise appropriate parameters.  We implemented $\ARCSplus$ in MATLAB. No parallelization was explicitly used and no special care was taken for speed.

\myparagraph{Robust Rotation Search}  From $\cN(0,\bI_3)$ we randomly sampled point pairs $\{(\by_i,\bx_i)\}_{i=1}^\ell$ with $k^*$ inliers and noise $\beps_{i}\sim \cN(0,\sigma^2\bI_3)$. Specifically, we generated the ground-truth rotation $\bR^*$ from an axis randomly sampled from $\bbS^2$ and an angle from $[0,2\pi]$, rotated $k^*$ points randomly sampled from $\cN(0,\bI_3)$ by $\bR^*$, and added noise to obtain $k^*$ inlier pairs. Every outlier point $\by_j$ or $\bx_j$ was randomly sampled from $\cN(0,\bI_3)$ with the constraint $-c\leq \norm{\by_j} - \norm{\bx_j}\leq c$; otherwise $(\by_j,\bx_j)$ might simply be detected and removed by computing $\norm{\by_j} - \norm{\bx_j}$.

We compared $\ARCSplus_{\texttt{O}}$ and $\ARCSplus_{\texttt{R}}$ and their combination $\ARCSplus_{\texttt{OR}}$ with prior works. The results are in Table \ref{table:RRS-scalibility}. We first  numerically illustrate the \textit{accuracy versus scalability} dilemma in prior works (\S \ref{subsection:prior-art}). On the one hand, we observed an extreme where \textit{accuracy overcomes scalability}: $\RANSAC$ performed well with error $0.39$ when $k^*/\ell = 10^3/10^5$, but its running time increased greatly with decreasing inlier ratio, from $29$ seconds to more than $8.4$ hours. The other extreme is where \textit{scalability overcomes accuracy}: Both $\GNCTLS$ and $\FGR$ failed in presence of such many outliers---as expected---even though their running time scales linearly with $\ell$. 

Table \ref{table:RRS-scalibility} also depicted the performance of our proposals $\ARCSplus_{\texttt{O}}$ and $\ARCSplus_{\texttt{R}}$. Our approximate consensus strategy $\ARCSplus_\texttt{O}$ reached a balance between accuracy and scalability. In terms of accuracy, it made errors smaller than $1$ degree, as long as there are more than $3\times10^3/10^7=0.03\%$ inliers; this was further refined by Riemannian subgradient descent $\ARCSplus_{\texttt{R}}$, so that their combination $\ARCSplus_{\texttt{OR}}$ had even lower errors. In terms of scalability, we observed that $\ARCSplus_{\texttt{OR}}$ is uniformly faster than $\FGR$, and is at least $1800$ times faster than $\GORE$ for $k^*/\ell=10^3/10^6=0.1\%$. But it had been harder to measure exactly how faster $\ARCSplus_{\texttt{OR}}$ is than $\GORE$ and $\RANSAC$ for even larger point sets. Finally, $\ARCSplus_{\texttt{OR}}$ failed at $k^*/\ell = 10^3/10^7=0.01\%$. 

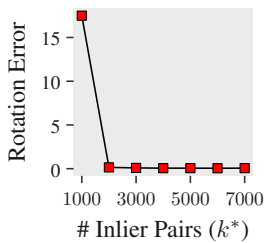
\begin{figure}
	\centering
\begin{tikzpicture}[x=1pt,y=1pt]
\definecolor{fillColor}{RGB}{255,255,255}
\path[use as bounding box,fill=fillColor,fill opacity=0.00] (0,0) rectangle (104.79, 99.73);
\begin{scope}
\path[clip] (  0.00,  0.00) rectangle (104.79, 99.73);
\definecolor{drawColor}{RGB}{255,255,255}
\definecolor{fillColor}{RGB}{255,255,255}

\path[draw=drawColor,line width= 0.6pt,line join=round,line cap=round,fill=fillColor] ( -0.00,  0.00) rectangle (104.79, 99.73);
\end{scope}
\begin{scope}
\path[clip] ( 31.53, 30.69) rectangle ( 99.29, 94.23);
\definecolor{fillColor}{gray}{0.92}

\path[fill=fillColor] ( 31.53, 30.69) rectangle ( 99.29, 94.23);
\definecolor{drawColor}{RGB}{0,0,0}

\path[draw=drawColor,line width= 0.6pt,line join=round] ( 34.61, 91.34) --
	( 44.88, 33.91) --
	( 55.14, 33.70) --
	( 65.41, 33.61) --
	( 75.68, 33.62) --
	( 85.94, 33.57) --
	( 96.21, 33.63);
\definecolor{fillColor}{RGB}{255,0,0}

\path[draw=drawColor,line width= 0.4pt,line join=round,line cap=round,fill=fillColor] ( 94.47, 31.89) rectangle ( 97.95, 35.37);

\path[draw=drawColor,line width= 0.4pt,line join=round,line cap=round,fill=fillColor] ( 84.21, 31.84) rectangle ( 87.68, 35.31);

\path[draw=drawColor,line width= 0.4pt,line join=round,line cap=round,fill=fillColor] ( 73.94, 31.88) rectangle ( 77.42, 35.36);

\path[draw=drawColor,line width= 0.4pt,line join=round,line cap=round,fill=fillColor] ( 63.67, 31.87) rectangle ( 67.15, 35.34);

\path[draw=drawColor,line width= 0.4pt,line join=round,line cap=round,fill=fillColor] ( 53.40, 31.96) rectangle ( 56.88, 35.44);

\path[draw=drawColor,line width= 0.4pt,line join=round,line cap=round,fill=fillColor] ( 43.14, 32.17) rectangle ( 46.61, 35.65);

\path[draw=drawColor,line width= 0.4pt,line join=round,line cap=round,fill=fillColor] ( 32.87, 89.61) rectangle ( 36.35, 93.08);
\end{scope}
\begin{scope}
\path[clip] (  0.00,  0.00) rectangle (104.79, 99.73);
\definecolor{drawColor}{gray}{0.10}

\node[text=drawColor,anchor=base east,inner sep=0pt, outer sep=0pt, scale=  0.73] at ( 26.58, 30.37) {0};

\node[text=drawColor,anchor=base east,inner sep=0pt, outer sep=0pt, scale=  0.73] at ( 26.58, 46.93) {5};

\node[text=drawColor,anchor=base east,inner sep=0pt, outer sep=0pt, scale=  0.73] at ( 26.58, 63.49) {10};

\node[text=drawColor,anchor=base east,inner sep=0pt, outer sep=0pt, scale=  0.73] at ( 26.58, 80.05) {15};
\end{scope}
\begin{scope}
\path[clip] (  0.00,  0.00) rectangle (104.79, 99.73);
\definecolor{drawColor}{gray}{0.20}

\path[draw=drawColor,line width= 0.6pt,line join=round] ( 28.78, 33.41) --
	( 31.53, 33.41);

\path[draw=drawColor,line width= 0.6pt,line join=round] ( 28.78, 49.96) --
	( 31.53, 49.96);

\path[draw=drawColor,line width= 0.6pt,line join=round] ( 28.78, 66.52) --
	( 31.53, 66.52);

\path[draw=drawColor,line width= 0.6pt,line join=round] ( 28.78, 83.08) --
	( 31.53, 83.08);
\end{scope}
\begin{scope}
\path[clip] (  0.00,  0.00) rectangle (104.79, 99.73);
\definecolor{drawColor}{gray}{0.20}

\path[draw=drawColor,line width= 0.6pt,line join=round] ( 96.21, 27.94) --
	( 96.21, 30.69);

\path[draw=drawColor,line width= 0.6pt,line join=round] ( 75.68, 27.94) --
	( 75.68, 30.69);

\path[draw=drawColor,line width= 0.6pt,line join=round] ( 55.14, 27.94) --
	( 55.14, 30.69);

\path[draw=drawColor,line width= 0.6pt,line join=round] ( 34.61, 27.94) --
	( 34.61, 30.69);
\end{scope}
\begin{scope}
\path[clip] (  0.00,  0.00) rectangle (104.79, 99.73);
\definecolor{drawColor}{gray}{0.10}

\node[text=drawColor,anchor=base,inner sep=0pt, outer sep=0pt, scale=  0.73] at ( 96.21, 19.68) {$7000$};

\node[text=drawColor,anchor=base,inner sep=0pt, outer sep=0pt, scale=  0.73] at ( 75.68, 19.68) {$5000$};

\node[text=drawColor,anchor=base,inner sep=0pt, outer sep=0pt, scale=  0.73] at ( 55.14, 19.68) {$3000$};

\node[text=drawColor,anchor=base,inner sep=0pt, outer sep=0pt, scale=  0.73] at ( 34.61, 19.68) {$1000$};
\end{scope}
\begin{scope}
\path[clip] (  0.00,  0.00) rectangle (104.79, 99.73);
\definecolor{drawColor}{gray}{0.10}

\node[text=drawColor,anchor=base,inner sep=0pt, outer sep=0pt, scale=  0.92] at ( 65.41,  7.64) {\# Inlier Pairs ($k^*$)};
\end{scope}
\begin{scope}
\path[clip] (  0.00,  0.00) rectangle (104.79, 99.73);
\definecolor{drawColor}{gray}{0.10}

\node[text=drawColor,rotate= 90.00,anchor=base,inner sep=0pt, outer sep=0pt, scale=  0.92] at ( 13.08, 62.46) {Rotation Error};
\end{scope}
\end{tikzpicture} \label{fig:experiments_SRSC1} 
	\caption{Rotation errors of $\ARCSplus$ on synthetic Gaussian point clouds. $20$ trials, $m=10^4$, $n=\floor{0.8m}$, $\sigma=0.01$. \label{fig:experiments_SRSC}}
\end{figure}

\myparagraph{Simultaneous Rotation and Correspondence Search} 
We randomly sampled point sets $\cQ$ and $\cP$ from  $\cN(0,\bI_3)$ with $k^*$ inlier pairs and noise $\beps_{i,j}\sim \cN(0,\sigma^2\bI_3)$ (cf. Problem \ref{problem:SRSC}). Each outlier point was randomly and independently drawn also from $\cN(0,\bI_3)$. Figure \ref{fig:experiments_SRSC} shows that $\ARCSplus$ accurately estimated the rotations for $k^*\geq 2000$ (in $90$ seconds), and broke down at $k^*=1000$, a situation where there were $k^*/m= 10\%$ overlapping points. We did not compare methods like $\TEASER$, $\GORE$, $\RANSAC$ here, because giving them correspondences from $\ARCSplus_\texttt{N}$ would result unsatisfactory running time or accuracy (recall Tables \ref{table:l-mn} and \ref{table:RRS-scalibility}), while feature matching methods like $\FPFH$ do not perform well on random synthetic data.

\begin{table*}[t]
	\centering
	\caption{Success rates of methods run on the scene pairs of the 3DMatch dataset \cite{Zeng-CVPR2017} for which the ground-truth rotation and translation are provided (rotation error smaller than $10$ degree means a success \cite{Yang-T-R2021}; see also the first paragraph of Appendix \ref{appendix:experiments}). \label{Table:3DMatch-MainPaper}} 
	\begin{tabular}{cccccccccc}  
		\toprule
		Scene Type & Kitchen &Home 1 & Home 2& Hotel 1& Hotel 2& Hotel 3& Study Room & MIT Lab & Overall  \\
		\# Scene Pairs & $506$ & $156$ & $208$ & $226$ & $104$ & $54$ & $292$ & $77$ & $1623$ \\
		\midrule
		$\TEASER$ & $\textbf{99.0\%}$ &$\textbf{98.1\%}$ & $94.7\%$ & $98.7\%$ & $\textbf{99.0\%}$ & $98.1\%$& $97.0\%$ & $94.8\%$  & $97.72\%$ \\
		$\ARCSplusplus_{\texttt{OR}}$  & $98.4\%$ & $97.4\%$&$\textbf{95.7\%}$&$98.7\%$ & $98.1\%$ & $\textbf{100\%}$ & $\textbf{97.3\%}$ & $\textbf{96.1\%}$  & $97.72\%$ \\
		\bottomrule
	\end{tabular}
\end{table*}

\subsection{Experiments on 3DMatch}
The 3DMatch\footnote{License info: \url{https://3dmatch.cs.princeton.edu/}} dataset \cite{Zeng-CVPR2017} contains more than $1000$ point clouds for testing, representing $8$ different scenes (such as kitchen, hotel, etc.), while the number of  point clouds for each scene ranges from $77$ to $506$. Each point cloud has more than $10^5$ points, yet in \cite{Zeng-CVPR2017} there are $5000$ keypoints for each cloud. We used the pretrained model\footnote{\url{https://github.com/zgojcic/3DSmoothNet}} of the \texttt{3DSmoothNet} \cite{Gojcic-CVPR2019} to extract descriptors from these key points, and matched them using the Matlab function \texttt{pcmatchfeatures}, with its parameter \texttt{MatchThreshold} set to the maximum $1$. We assume that the ground-truth translation $\bt^*$ is given, and run $\TEASER$ and $\ARCSplus_{\texttt{OR}}$ on $(\by_i-\bt^*,\bx_i)$'s; the performance is comparable. We did not compare other methods here, as $\TEASER$ currently has the best performance  on 3DMatch (to the best of our knowledge); see \cite{Yang-T-R2021} for comparison with optimization-based methods, and see \cite{Choy-CVPR2020} for the success rates (recall) of deep learning methods.

See the supplementary materials for more experiments.
\section{Discussion and Future Work}
Despite of the progress that we made for robust rotation search and simultaneous rotation and correspondence  search on large-scale point clouds, our $\ARCSplus$ pipeline has a few limitations, and we discuss them next.

For small datasets (\eg, $\ell \leq 500$), as in homography fitting \cite{Bustos-ICCV2015}, other methods, \eg,  \texttt{MAGSAC++} \cite{Barath-CVPR2019,Barath-CVPR2020,Barath-TPAMI2021}, \texttt{VSAC} \cite{Ivashechkin-ICCV2021}, $\TEASER$ \cite{Yang-T-R2021}, and $\GORE$ \cite{Bustos-ICCV2015} might be considered with higher priority; they come with efficient C++ implementations. For more points, \eg, $\ell \geq 10^4$, but with higher inlier rates than in Table \ref{table:RRS-scalibility} (\eg, $\geq 15\%$), $\GNCTLS$ \cite{Yang-RA-L2020} and $\RANSAC$ are our recommendations for what to use.

Modern point clouds have more than $10^5$ points, and are naturally correspondences-less (cf. \cite{Cai-ISPRS-J-PRS2019}). $\ARCS$ operates at that scale in the absence of noise (Table \ref{table:I2}), while $\ARCSplus$ can handle $m,n\approx 10000$ (Figure \ref{fig:experiments_SRSC}) and $\ARCSplus_{\texttt{OR}}$ can handle $\ell \approx10^7$ correspondences (Table \ref{table:RRS-scalibility}); all these are limited to the rotation-only case. To find rotation (and translation) from such point sets ``in the wild'', it seems inevitable to downsample them. An interesting future work is to theoretically quantify the tradeoff between downsampling factors and the final registration performance. Another tradeoff to quantify, as implied by Remark \ref{remark:I2-featmatching}, is this: Can we design a correspondence matching algorithm that better balances the number of remaining points and the number of remaining inliers? In particular, such matching should take specific pose into consideration (cf. $\ARCS$); many methods did not.

Like $\TEASER$, $\GORE$, $\GNCTLS$, $\RANSAC$, our algorithm relies on an inlier threshold $c$. While how to set this hyper-parameter suitably is known for Gaussian noise with given variance, \textit{in practice the distance threshold is usually chosen empirically}, as Hartley \& Zisserman wrote \cite{Hartley-2004}. While mis-specification of $c$ could fail the registration, certain heuristics have been developed to alleviate the sensitivity to such mis-specification; see \cite{Barath-CVPR2019,Barath-CVPR2020,Barath-TPAMI2021,Antonante-TRO2021}. Finally, our experience is to set $c$ based on the scale of the point clouds. 

Our outlier removal component $\ARCSplus_{\texttt{O}}$ presented good performance (Table \ref{table:I-tilde}), yet with no optimality guarantees. Note that, with $s=90$ we have $|\phi_j - \phi^*|\leq 1$ for some $\phi_j$, while Figure \ref{fig:choice_s_10} shows that $\ARCSplus_{\texttt{O}}$ gave roughly $1$ degree error at $s=90$.  Theoretically justifying this is left as future work. Without guarantees, registration could fail, which might lead to undesired consequences in safety-critical applications. On the other hand, we believe that $\ARCSplus$ is a good demonstration of trading optimality guarantees for accuracy and scalability; enforcing all of the three properties amounts to requiring solving NP hard problems efficiently at large scale! In fact, since any solutions might get certified for optimality (Remark \ref{remark:posteriori}), bold algorithmic design ideas can be taken towards improving accuracy and scalability, while relying on other tools for optimality certification.

\myparagraph{Acknowledgments} The first author was supported by the MINDS PhD fellowship at Johns Hopkins University. This work was supported by NSF Grants 1704458 and 1934979, and by the Northrop Grumman Mission Systems Research in Applications for Learning Machines (REALM) initiative.

{\small

}

\newpage
\appendix

\section{Proof of Proposition \ref{prop:SO3toDPCP}}\label{appendix:proof-prop-SO3toS3}
We consider a stronger version of Proposition \ref{prop:SO3toDPCP}:
\begin{prop}\label{prop:SO3toDPCP1}
	We have $\trsp{\bw}\bD_i\bw=\norm{\by_i - \bR \bx_i}^2$, where $\bw$ is a quaternion representation of $\bR$ of $\eqref{eq:L1-SO3}$, and $\bD_i\in\bbR^{4\times 4}$ is a positive semi-definite matrix whose entries depend on $\bx_i$, $\by_i$. So	Problem \eqref{eq:L1-SO3} is equivalent to 
	\begin{align}\label{eq:L1-S31}
		\min_{\bw\in\bbS^3} h(\bw),\ \ \ \ h(\bw) = \sum_{i=1}^\ell \sqrt{\trsp{\bw}\bD_i\bw}.
	\end{align}
	Moreover, $\bD_i$ has eigenvalues $4,4,0,0$ if $\bx_i$ and $\by_i$ are normalized (that is $\norm{\bx_i}=\norm{\by_i}=1$).
\end{prop}

We first recall some basics about unit quaternions, an algebraic construction invented by Hamilton in the $1840$'s, when the notion of \textit{vector} does not exist; see the beautiful account of \cite{Altmann-1989}. In our current notation, each element $\bw$ of $\bbS^3$ is called a unit quaternion. The most crucial fact is that $\SO(3)$ is isomorphic to the $3$-sphere $\bbS^3$ \textit{up to sign}, that is $\SO(3)\equiv \bbS^3/\{\pm 1\}$. This implies a two-to-one correspondence between unit quaternions and $3$D rotations. Algebraically, any $\bR\in\SO(3)$ can be written as a $3\times 3$ matrix
\begin{align*}
	\footnotesize{\begin{matrix}
			w_1^2+w_2^2-w_3^2 -w_4^2 & 2(w_2w_3-w_1w_4)& 2(w_2w_4+w_1w_3) \\
			2(w_2w_3+w_1w_4)& w_1^2+w_3^2-w_2^2-w_4^2& 2(w_3w_4-w_1w_2)\\
			2(w_2w_4-w_1w_3)& 2(w_3w_4+w_1w_2)& w_1^2+w_4^2-w_2^2-w_3^2\\
	\end{matrix}}
\end{align*}
where $\bw=\trsp{[w_1,w_2,w_3,w_4]}\in\bbS^3$. We can now write the three entries of $\bR \bx_i$ as quadratic forms $\trsp{\bw} \bX_{i,1} \bw$, $\trsp{\bw} \bX_{i,2} \bw$, and $\trsp{\bw} \bX_{i,3} \bw $, respectively. Here $\bX_{i,1}$, $\bX_{i,2}$, and $\bX_{i,3}$ are $4\times 4$ symmetric matrices, defined as
\begin{align}
	\bX_{i,1}&=\begin{bmatrix}
		[\bx_i]_1&0&[\bx_i]_3&-[\bx_i]_2\\
		0&[\bx_i]_1&[\bx_i]_2&[\bx_i]_3\\
		[\bx_i]_3&[\bx_i]_2&-[\bx_i]_1&0\\
		-[\bx_i]_2&[\bx_i]_3&0& -[\bx_i]_1
	\end{bmatrix}\\
	\bX_{i,2}&= \begin{bmatrix}
		[\bx_i]_2&-[\bx_i]_3&0&[\bx_i]_1\\
		-[\bx_i]_3&-[\bx_i]_2&[\bx_i]_1&0\\
		0&[\bx_i]_1&[\bx_i]_2&[\bx_i]_3\\
		[\bx_i]_1&0&[\bx_i]_3&-[\bx_i]_2
	\end{bmatrix}\\
	\bX_{i,3}&= \begin{bmatrix}
		[\bx_i]_3&[\bx_i]_2&-[\bx_i]_1&0\\
		[\bx_i]_2&-[\bx_i]_3&0&[\bx_i]_1\\
		-[\bx_i]_1&0&-[\bx_i]_3&[\bx_i]_2\\
		0&[\bx_i]_1&[\bx_i]_2&[\bx_i]_3
	\end{bmatrix}
\end{align}
Defining $\bC_i:=[\by_i]_1\bX_{i,1} +[\by_i]_2\bX_{i,2}+[\by_i]_3\bX_{i,3}$, we get that $\trsp{\by_i}\bR\bx_i=\trsp{\bw}\bC_i\bw$. And defining 
\begin{align}
	\bD_i = (\norm{\by_i}^2 + \norm{\bx_i}^2)\bI_4-2\bC_{i}
\end{align}
with $\bI_4$ the $4\times 4$ identity matrix, we obtain the equality
\begin{align}
	\norm{\by_i - \bR \bx_i}^2&= \norm{\by_i}^2 + \norm{\bx_i}^2 - 2\trsp{\by_i}\bR\bx_i\\
	&=\trsp{\bw} \bD_i \bw.
\end{align}
Since $\bD_i$ is symmetric and $\trsp{\bw} \bD_i \bw\geq 0$ for any $\bw\in\bbS^3$, we know that $\bD_i\in\bbR^{4\times 4}$ is positive semi-definite. 

Suppose $\norm{\by_i}=\norm{\bx_i}=1$. Then there is at least two different $3$D rotations $\bR_1$ and $\bR_2$ satisfying $\by_i=\bR_1\bx_i=\bR_2\bx_i$. Thus, with the factorization $\bD_i = \bZ_i\trsp{\bZ_i}$, there are at least two quaternions $\bw_1$ and $\bw_2$ with $\bw_1\neq \pm \bw_2$ satisfying that $\trsp{\bZ_i}\bw_1=\trsp{\bZ_i}\bw_2=0$. So $\rank(\bD_i)=\rank(\bZ_i)\leq 2$. Recalling $\bD_i = 2\bI_4-2\bC_{i}$, we see that $1$ is an eigenvalue of $\bC_i$ that has multiplicity at least $2$. Similarly, we can derive that $\norm{\by_i+\bR\bx_i}^2=\trsp{\bw}\bD_i'\bw$ where $\bD_i'=(\norm{\by_i}^2 + \norm{\bx_i}^2)\bI_4+2\bC_{i}=2\bI_4+2\bC_i$ is positive semi-definite of rank at most $2$. That is, $-1$ is an eigenvalue of $\bC_i$ of multiplicity at least $2$. Concluding, $\bC_i$ has eigenvalues $1,1,-1,-1$ and $\bD_i$ has eigenvalues $4,4,0,0$.

\section{Proposition \ref{prop:sharpness}: Proof and Interpretation}
Here we provide a proof (Appendix \ref{appendix:proof-sharpness}) and probabilistic interpretation (Appendix \ref{appendix:probability-prop-sharpness}) for Proposition \ref{prop:sharpness}. In this section, we use the notation $\bD_i=\bZ_i\trsp{\bZ_i}$ from Appendix \ref{appendix:proof-prop-SO3toS3} where we decomposed every positive semidefinite matrix $\bD_i$ into the product of its root $\bZ_i$. Since we could always normalize the point sets $\by_i$ and $\bx_i$, and then normalize $\bD_i$, we assume without loss of generality that $\bD_i$ has eigenvalues $1,1,0,0$ (cf. Proposition \ref{prop:SO3toDPCP1}). In this situation, we can now specify that $\bZ_i$ is a matrix of size  $4\times 2$ and it has orthonormal columns, i.e., $\trsp{\bZ_i}\bZ_i=\bI_2$. Also, we see that the objective function \eqref{eq:L1-S3} of interest can be rewritten as 
\begin{align}\label{eq:L1-S3-gDPCP}
	\min_{\bw\in\bbS^3} h(\bw),\ \ \ \ h(\bw) = \sum_{i=1}^\ell \norm{\trsp{\bZ_i}\bw}.
\end{align}
Note that, if $\bZ_i$ had a single column, then \eqref{eq:L1-S3-gDPCP} is exactly the problem of \textit{dual principal component pursuit} (DPCP) \cite{Tsakiris-JMLR2018}. On the other hand, one could think of \eqref{eq:L1-S3-gDPCP} as a \textit{group} version of DPCP, as $\norm{\trsp{\bZ_i}\bw}$ here promotes group sparsity. A similar group version of DPCP was considered by \cite{Ding-CVPR2020} in the context of homography estimation. In \cite{Ding-CVPR2020}, the authors provided conditions under which any global minimizer of \eqref{eq:L1-S3-gDPCP} coincides with the ground-truth normal vector, or, in our context, the ground-truth unit quaternion $\pm \bw^*$. Thus, our contribution here, if viewed from the angle of group-DPCP, is to show that, there is actually an efficient algorithm that exactly reaches the guaranteed ground-truth normal. We present our contribution next.

\subsection{Proof of Proposition \ref{prop:sharpness}}\label{appendix:proof-sharpness}
The proof follows from Proposition 4 of \cite{Li-SIAM-J-O2021} with some simplification for specializing arbitrary Stiefel manifolds to $\bbS^3$, and with some modification to tighten a constant factor (from $2$ to $\sqrt{2}$). We also note that $\eta_{\min}$ and $\eta_{\max}$ are motivated from their corresponding definitions.

Write $\bw:=c_0\bw_0 + c^*\bw^*$ with $c_0^2+(c^*)^2=1$ and $\bw_0\in\cS^*$. Without loss of generality assume $c^*\geq 0$. Then
\begin{align}\label{eq:c0_dist}
	\dist(\bw,\pm\bw^*)&= \min \big\{\sqrt{2+2c^*}, \sqrt{2-2c^*} \big \} \nonumber \\
	&=\sqrt{2-2c^*} \\
	&\leq \sqrt{2-2(c^*)^2}=\sqrt{2}c_0. \nonumber
\end{align}
If $i\in\cI^*$ then by Proposition \ref{prop:SO3toDPCP} we have
\begin{align}
	\norm{\trsp{\bZ_i} \bw^*}=\sqrt{\trsp{\bw}\bD_i\bw}=\norm{\by_i-\bR^*\bx_i}=0.
\end{align} 
Hence the difference $h(\bw) - h(\bw^*)$ is equal to 
\begin{align*}
	c_0\sum_{i\in\cI^*} \norm{\trsp{\bZ_i}\bw_0} + \sum_{i\in [\ell]\backslash \cI^*} \Big(\norm{\trsp{\bZ_i}\bw} -\norm{\trsp{\bZ_i}\bw^*}\Big).
\end{align*}
By \eqref{eq:c0_dist} and the definition of $\eta_{\min}$ \eqref{eq:etamin}, we know that 
\begin{align}
	c_0\sum_{i\in\cI^*} \norm{\trsp{\bZ_i}\bw_0}\geq \frac{k^* \eta_{\min}\dist(\bw,\pm\bw^*)}{\sqrt{2}}.
\end{align}
By triangle inequality the second summation in the above the difference $h(\bw) - h(\bw^*)$ is smaller than or equal to $\sum_{i\in [\ell]\backslash \cI^*}\norm{\trsp{\bZ_i}(\bw-\bw^*)}$, but this bound satisfies
\begin{align*}
	\sum_{i\in [\ell]\backslash \cI^*}\norm{\trsp{\bZ_i}(\bw-\bw^*)}\leq  (\ell-k^*)\eta_{\max} \dist(\bw,\pm\bw^*),
\end{align*}
where we used $\dist(\bw,\pm\bw^*)=\sqrt{2-2c^*}=\norm{\bw-\bw^*}$ \eqref{eq:c0_dist} and the definition of \eqref{eq:etamax}. We finished the proof.

\subsection{Probabilistic Interpretation of Proposition \ref{prop:sharpness}}\label{appendix:probability-prop-sharpness}

\subsubsection{Technical Assumptions}\label{appendix:assumptions}
We assume there is no noise for two reasons. First, analysis for noisy data is more challenging and requires a full different chapter to penetrate. Second, analysis in the noiseless case typically serves as a starting point for and sheds enough light on analysis for noise. For example, see the trajectory of the development from the noiseless case \cite{Tsakiris-JMLR2018} to the noisy case \cite{Ding-ICML2019} in the context of DPCP.

Next, we discuss probabilistic assumptions on inliers. For an inlier index $i\in\cI^*$, each column of $\bZ_i$ lies in the ground-truth hyperplane $\cS^*\subset\bbR^4$ that is perpendicular to the ground-truth unit quaternion $\pm \bw^*$, and the two columns of $\bZ_i$ span a subspace $\cS_i$ of dimension $2$ that is contained in $\cS^*$. Note that any $\bZ_i'\in\bbR^{4\times 2}$ whose columns are in $\cS_i\cap \bbS^3$ are equivalent to $\bZ_i$ in the sense that $\norm{\trsp{\bZ_i}\bw^*}=\norm{\trsp{(\bZ_i')}\bw^*}=0$. To impose randomness assumptions on $\bZ_i$, one could simply replace $\bZ_i$ by a $4\times 2$ random matrix whose columns are independently sampled uniformly at random from the intersection $\cS_i\cap \bbS^3$. In fact, we need a slightly stronger assumption:
\begin{assumption}[\textit{randomness on inliers}]\label{assumption:inlier}
	For each $i\in\cI^*$, every column of $\bZ_i$ is independently sampled uniformly at random from the intersection $\cS^*\cap \bbS^3$.
\end{assumption}
This assumption destroys some good property of $\bZ_i$: it might not be orthonormal in general. However, it is orthonormal in expectation, i.e., it satisfies $\bbE[\trsp{\bZ_i}\bZ_i]=\bI_2$. This will suffice for our later analysis.

On the other hand, Assumption \ref{assumption:inlier} simplifies matters by a lot. This can be appreciated in comparison with a ``common'' approach, where one makes assumptions on the ``source data'', which are point pairs $(\by_i,\bx_i)$'s in our case. Let us first recall the ``data flow'' from $(\by_i,\bx_i)$ to $\bZ_i$:
\begin{align}
	(\by_i,\bx_i) \xmapsto{\text{Proposition \ref{prop:SO3toDPCP1}}} \bD_i \xmapsto{\text{factorizing $\bD_i$}} \bZ_i
\end{align}
In view of the above flow (or graphical model), one intuitively (not very rigorously) feels that, if  $(\by_i,\bx_i)$'s are independent, then $\bZ_i$'s are independent; the latter is implied by Assumption \ref{assumption:inlier}. On the other hand, it seems hard to know the distribution of $\bZ_i$'s, even if the distribution of $(\by_i,\bx_i)$'s is given or assumed. It is via Assumption \ref{assumption:inlier} that this challenge is circumvented and that our theorems are developed.

Finally, we need randomness on outliers:
\begin{assumption}[\textit{randomness on outliers}]\label{assumption:outlier}
	Each column of any outlier $\bZ_j$, where $j\in [\ell]\backslash \cI^*$, is independently sampled uniformly at random from $\bbS^3$.
\end{assumption}
Since an outlier $\bZ_j$ could be distributed arbitrarily, this assumption is the most natural, if not the most challenging, as the outliers try their best to mimic the distribution of inliers. Assumptions \ref{assumption:inlier} and \ref{assumption:outlier} (together with the noiseless assumption) are all we need for the next section.

\subsubsection{Probabilistic Interpretation}\label{subsubsection:PI}
Recall that the quantities $\eta_{\min}$, $\eta_{\max}$ of interest are equal to
\begin{align}
	\eta_{\min}&=\frac{1}{k^*}\min_{\bw\in\cS^*\cap \bbS^3}\sum_{i\in\cI^*}\norm{\trsp{\bZ_i}\bw}, \text{\ \  and } \\
	\eta_{\max}&= \frac{1}{\ell-k^*} \max_{\bw\in\bbS^3} \sum_{j\in [\ell]\backslash \cI^*}\norm{\trsp{\bZ_j}\bw}.
\end{align}
The following proposition gives probabilistic upper and lower bounds for $\eta_{\max}$ and $\eta_{\min}$ respectively. 
\begin{prop}\label{prop:bound-eta}
	Under the assumptions of \S \ref{appendix:assumptions}, we have
	\begin{enumerate}[label=(\roman*)]
		\item\label{enum:eta1} With probability at least $1-\exp(-\zeta^2/2)$ it holds that 
		\begin{align}
			\eta_{\max} \leq \frac{1}{\sqrt{2}} + \frac{(4+\zeta)}{\sqrt{\ell - k^*}}.
		\end{align}
		\item\label{enum:eta2} With probability at least $1-\exp(-\zeta^2/2)$ it holds that
		\begin{align}
			\eta_{\min} \geq \sqrt{\frac{2}{3}} - \frac{(4+\zeta)}{\sqrt{k^*}}
		\end{align}
	\end{enumerate}
\end{prop}
To prove Proposition \ref{prop:bound-eta} (cf. Appendix \ref{appendix:proof-bound}), we combine the proof strategies of \cite{Li-SIAM-J-O2021} and \cite{Zhu-NeurIPS2018}, where both sets of the authors found inspirations from \cite{Lerman-FoCM2015}.
We can now see that the condition of Proposition \ref{prop:sharpness}, $\alpha^*:=k^* \eta_{\min}/\sqrt{2}-(\ell-k^*)\eta_{\max}>0$, holds with high probability as long as 
\begin{align*}
	\sqrt{\frac{2}{3}}k^* -  (4+\zeta)\sqrt{k^*} \geq \frac{1}{\sqrt{2}}(\ell-k^*) + (4+\zeta)\sqrt{\ell - k^*}.
\end{align*}
Ignoring lower-order terms we get the condition
\begin{align}\label{eq:condition-inlier-ratio}
	k^*\gtrsim \frac{\sqrt{3}}{2}(\ell-k^*) \Leftrightarrow \frac{k^*}{\ell}\gtrsim \frac{\sqrt{3}}{\sqrt{3}+2},
\end{align}
which holds true whenever there are sufficiently many inliers. This condition ensures the $\alpha^*$-sharpness, from which local linear convergence to $\pm \bw^*$ from a good enough initialization with proper stepsize ensues.
\subsubsection{Details: Proof of Proposition \ref{prop:bound-eta}}\label{appendix:proof-bound}
We need the following simple result, with its proof omitted.
\begin{lemma}
	If $\bz=\trsp{[z_1,z_2,z_3,z_4]}$ sampled uniformly at random from $\bbS^3$, we have for any $\bw\in\bbS^3$ that 
	\begin{align}\label{eq:E-zw}
		\bbE\big[(\trsp{\bz}\bw)^2\big] = \frac{1}{4}.
	\end{align}
	On the other hand, if $\hat{\bz}$ is sampled uniformly at random from $\bbS^3\cap \cS$ where $\cS$ is a linear subspace of $\bbR^4$ of dimension $3$, then we have for every $\hat{\bw}\in\bbS^3\cap \cS$ that
	\begin{align}\label{eq:E-zwS}
			\bbE\big[(\trsp{\hat{\bz}}\hat{\bw})^2\big] = \frac{1}{3}.
	\end{align}
\end{lemma}

\myparagraph{Upper Bounding $\eta_{\max}$ \ref{enum:eta1}} We first prove \ref{enum:eta1} of Proposition \ref{prop:bound-eta}. Consider  matrix  $\bZ\in\bbR^{4\times 2}$ whose columns are sampled independently and uniformly at random from the $3$-sphere $\bbS^3$. We will give upper bounds respectively for  
\begin{align}
	&(\ell-k^*) \max_{\bw\in\bbS^3}\bbE\Big[\norm{\trsp{\bZ}\bw}\Big] \text{ and} \label{eq:37-first} \\
	&\max_{\bw\in\bbS^3} \sum_{j\in [\ell]\backslash \cI^*} \bigg( \norm{\trsp{\bZ_j}\bw} - \bbE\Big[\norm{\trsp{\bZ}\bw}\Big] \bigg), \label{eq:37-second}
\end{align}
while summing the two bounds gives an upper bound for $(\ell-k^*)\eta_{\max}$. For \eqref{eq:37-first},  Jensen's inequality gives
\begin{align}
	\max_{\bw\in\bbS^3}\bbE\Big[\norm{\trsp{\bZ}\bw}\Big] &\leq \max_{\bw\in\bbS^3} \sqrt{\bbE\Big[\norm{\trsp{\bZ}\bw}^2\Big]} \\
	&=\max_{\bw\in\bbS^3} \sqrt{2\cdot \frac{1}{4}}  = \frac{1}{\sqrt{2}}. \label{eq:39}
\end{align}
To obtain \eqref{eq:39} we used \eqref{eq:E-zw} and the linearity of the expectation. The second term \eqref{eq:37-second} is harder to handle, and we first consider its expectation $\bbE[\eqref{eq:37-second}]$. We know from a standard symmetrization argument (cf. \cite{Kakade-2011}, Lemma 11.4 of \cite{Boucheron-book2013}) that, since $\bZ_j$'s are independent (Assumption \ref{assumption:outlier}), the expectation $\bbE[\eqref{eq:37-second}]$ has the following bound:
\begin{align}\label{eq:symmetrization}
	\bbE[\eqref{eq:37-second}]\leq 2\ \bbE\bigg[ \max_{\bw\in\bbS^3} \sum_{j\in [\ell]\backslash \cI^*} r_j\norm{\trsp{\bZ_j}\bw} \bigg],
\end{align}
where $r_j$'s are independent Radeamacher random variables which take values $1$, $-1$ with probabilities $1/2$ each and independent of $\bZ_j$'s. We also know from the \textit{vector contraction inequality} (cf. Corollary 1 of \cite{Maurer-ICALT2016}) that the right-hand side of \eqref{eq:symmetrization}, and thus $\bbE[\eqref{eq:37-second}]$, is has the following bound:
\begin{align}
	\bbE[\eqref{eq:37-second}]\leq &\ 2\sqrt{2}\ \bbE \bigg[ \max_{\bw\in\bbS^3} \sum_{j\in [\ell]\backslash \cI^*} \Big(r_{j1}\trsp{\bZ_{j1}}\bw+r_{j2}\trsp{\bZ_{j2}}\bw \Big) \bigg] \nonumber\\
	=&\ 2\sqrt{2}\ \bbE \bigg[   \Big\| \sum_{j\in [\ell]\backslash \cI^*}\big(r_{j1}\bZ_{j1} + r_{j2}\bZ_{j2} \big) \Big\|_2 \bigg]  \label{eq:vector-contraction}
\end{align}
where $\bZ_{j1}$'s and $\bZ_{j2}$'s are the first and second columns of $\bZ_j$ respectively, while $r_{j1}$'s and $r_{j2}$'s are independent Radeamacher random variables that are also independent of entries of $\bZ_{j}$'s. Applying Jensen's inequality to \eqref{eq:vector-contraction} we get 
\begin{align*}
	\bbE[\eqref{eq:37-second}] \leq &\ 2\sqrt{2}\ \sqrt{\bbE \bigg[   \Big\| \sum_{j\in [\ell]\backslash \cI^*}\big(r_{j1}\bZ_{j1} + r_{j2}\bZ_{j2} \big) \Big\|_2^2 \bigg]} \\
	= &\ 2\sqrt{2}\ \sqrt{\bbE \bigg[    \sum_{j\in [\ell]\backslash \cI^*}\big(r_{j1}^2\trsp{\bZ_{j1}}\bZ_{j1} + r_{j2}^2\trsp{\bZ_{j2}}\bZ_{j2} \big)  \bigg]} \\
	=&\ 4 \sqrt{\ell - k^*}
\end{align*}
To summarize, we have $\bbE[\eqref{eq:37-second}]\leq 4 \sqrt{\ell - k^*}$. Treat now \eqref{eq:37-second} as a function of $\bZ_j$'s. It is straightforward to verify that this function has \textit{bounded difference} $2$ (cf. \cite{Mcdiarmid-1989}). Since $\bZ_j$'s are independent (Assumption \ref{assumption:outlier}), Mcdiarmid's Lemma \cite{Mcdiarmid-1989} or the \textit{bounded difference inequality} is applicable, from which we obtain the following probability bound:
\begin{align}
	\bbP\Big(\eqref{eq:37-second} \geq \bbE[\eqref{eq:37-second}]+\zeta_0\Big) \leq \exp\Big(-\frac{\zeta_0^2}{2(\ell-k^*)}\Big).
\end{align}
With $\bbE[\eqref{eq:37-second}]\leq 4 \sqrt{\ell - k^*}$ and $\zeta:=\zeta_0/\sqrt{(\ell-k^*)}$, we get
\begin{align}
	\bbP\Big(\eqref{eq:37-second} \leq (4+\zeta) \sqrt{\ell-k^*} \Big) \geq 1- \exp\Big(-\frac{\zeta^2}{2}\Big).
\end{align}
Combining this with \eqref{eq:39} finishes proving \ref{enum:eta1}.

\myparagraph{Lower Bounding $\eta_{\min}$ \ref{enum:eta2}} Let $\bU\in\bbR^{4\times 3}$ have orthonormal columns and have $\cS^*$ as its column space, then there is a unique $\bv\in\bbS^2$ so that $\bU \bv = \bw$ for any $\bw\in\bbS^3$. Also, since for any $i\in\cI^*$ every column of $\bZ_i$ is in $\cS^*$, there is a unique $\bA_i\in\bbR^{3\times 2}$ with orthonormal columns satisfying $\bZ_i = \bU \bA_i$. Moreover, by rotation invariance we know that each column of $\bA_i$ is uniformly distributed on $\bbS^2$. As a result, we get $\trsp{\bZ_i}\bw=\trsp{\bA_i}\bv,\forall i\in\cI^*$, and $\eta_{\min}$ is equal to
\begin{align}
		\eta_{\min}&=\frac{1}{k^*}\min_{\bv\in \bbS^2}\sum_{i\in\cI^*}\norm{\trsp{\bA_i}\bv}.
\end{align}
Now, lower bounding $\eta_{\min}$ can be done in a similar way to upper bounding $\eta_{\max}$; thus we only give a proof sketch next. Similarly to \eqref{eq:37-first} and \eqref{eq:37-second}, to bound $\eta_{\min}$ we will find lower bounds respectively for the two terms 
\begin{align}
	& k^*\min_{\bv\in \bbS^2} \bbE\Big[\norm{\trsp{\bA}\bv}\Big]  \text{ and} \\
	&\min_{\bv\in \bbS^2} \sum_{i\in \cI^*} \bigg( \norm{\trsp{\bA_i}\bv} - \bbE\Big[\norm{\trsp{\bA}\bv}\Big] \bigg), \label{eq:43-second}
\end{align}
where $\bA$ is an i.i.d. copy of $\bA_i$. Similarly to \eqref{eq:39}, the first term here is bounded using \eqref{eq:E-zwS} and Jensen's inequality:
\begin{align}\label{eq:bound-2sqrt3}
	\min_{\bv\in \bbS^2} \bbE\Big[\norm{\trsp{\bA}\bv}\Big] \leq \sqrt{\frac{2}{3}}
\end{align}
Using the symmetric argument, the vector contraction inequality, and Jensen's inequality, the expectation $\bbE[\eqref{eq:43-second}]$ of the second term \eqref{eq:43-second} is bounded below by $-4\sqrt{k^*}$. Similarly, invoking Mcdiarmid's Lemma gives that 
\begin{align}
	&\ \bbP\Big(\eqref{eq:43-second} \leq \bbE[\eqref{eq:43-second}] - \zeta_0 \Big) \leq \exp\Big(-\frac{\zeta_0^2}{2k^*}\Big) \\
	\Rightarrow&\ \bbP\Big(\eqref{eq:43-second} \geq - (4+\zeta) \sqrt{k^*}\Big) \geq 1- \exp\Big(-\frac{\zeta^2}{2}\Big)
\end{align}
where $\zeta_0$ is any positive constant and we set $\zeta:=\zeta_0/\sqrt{k^*}$. Combining \eqref{eq:bound-2sqrt3} with the above bound finishes the proof.

\section{Proof of Proposition \ref{prop:axis}}\label{appendix:prop-axis}
Since $\bm{b}^*$ is the rotation axis of $\bR^*$, we have $\trsp{(\bR^*)}\bm{b}^*=\bm{b}^*$. Recall $\bv_i = \by_i-\bx_i$ for every $i\in\cI$. If $i\in\cI^*$ then 
\begin{align}
	\trsp{\bv_i}\bm{b}^*=\trsp{(\by_i-\bx_i)}\bm{b}^* = \trsp{(\by_i-\bR^*\bx_i)}\bm{b}^*,
\end{align}
and further more if $(\by_i,\bx_i)$ is an inlier pair we get  that 
\begin{align}
	|\trsp{\bv_i}\bm{b}^*| = \trsp{\beps_{i}}\bm{b}.
\end{align}
Clearly $\trsp{\beps_{i}}\bm{b}$ is a Gaussian random variable with zero mean and variance $\sigma^2$. The rest of the proof follows from a standard probability calculation.

\section{Interval Stabbing}
Here we provide proofs for Propositions \ref{prop:f} and \ref{lemma:g}. Along the way we will need multiple temporary variables to illustrate the idea; we use $a_{i,j}$'s to denote those variables. Here, $i$ denotes the $i$-th point pair, and $j$ denotes the order in which $a_{i,j}$ appears for the first time. In \S \ref{subsection:interval stabbing} we reviewed interval stabbing for closed intervals $\cJ_i$ of the form $[a,b]$. One should note and verify that this can be easily extended to the case where $\cJ_i$ is a finite (disjoint) union of closed intervals.
\subsection{Proof of Proposition \ref{prop:f}}\label{appendix:prop-f}
Recall $\bm{b}=\trsp{[\sin(\theta)\cos(\phi),\ \sin(\theta)\sin(\phi),\ \cos(\theta)]}$ with  $\theta\in[0,\pi]$, $\phi\in[0,\pi]$. Denote by $a_{i,1}:=[\bv_i]_1\cos(\phi)+[\bv_i]_2\sin(\phi)$, then $|\trsp{\bv_i}\bm{b}|\leq c$ is equivalent to
\begin{align}\label{eq:a1}
	|a_{i,1}\sin(\theta)+[\bv_i]_3\cos(\theta)|\leq c.
\end{align}
Without loss of generality we can assume that $a_{i,1}\geq 0$. So there is a unique $a_{i,2}\in[0,\pi]$ which satisfies
\begin{align*}
	\cos(a_{i,2})=\frac{[\bv_i]_3}{\sqrt{[\bv_i]_3^2+a_{i,1}^2}},\ \ \  \sin(a_{i,2})=\frac{a_{i,1}}{\sqrt{[\bv_i]_3^2+a_{i,1}^2}}.
\end{align*}
Hence \eqref{eq:a1} is equivalent to
\begin{align*}
	|\cos(\theta-a_{i,2})|\leq c_i,\ \ \  c_i:= \min\Bigg\{1, \frac{c}{\sqrt{[\bv_i]_3^2+a_{i,1}^2}}\Bigg\}
\end{align*}
Since the trigonometric function $\arccos:[0,\pi]\to[-1,1]$ is decreasing and $|\theta-a_{i,2}|\leq \pi$, the above is equivalent to
\begin{align*}
	a_{i,3}:=\arccos(-c_i)\geq |\theta-a_{i,2}|\geq \arccos(c_i) =: a_{i,4}.
\end{align*}
Define $a_{i,5}=a_{i,2}-a_{i,3},\ a_{i,6}=a_{i,2}-a_{i,4},\  a_{i,7}=a_{i,2}+a_{i,4},$ and $a_{i,8}=a_{i,2}+a_{i,3}$. Then $|\trsp{\bv_i}\bm{b}|\leq c$ is the same as
\begin{align}\label{eq:theta_in_interval}
	\theta \in \big([a_{i,5}, a_{i,6}]\cup [a_{i,7}, a_{i,8}]\big)\cap [0,\pi] 
\end{align}
To summarize, given $\phi\in[0,\pi]$, the $i$-th constraint of \eqref{eq:max-consensus-b} requires $\theta$ to lie in the union of some disjoint intervals defined in \eqref{eq:theta_in_interval}. So maximizing \eqref{eq:max-consensus-theta} amounts to finding a maximal set of intervals of the form \eqref{eq:theta_in_interval} that overlap a point $\theta$, and can be solved by interval stabbing in $O(\ell\log \ell)$ time. 

\subsection{Proof of Proposition \ref{lemma:g}}\label{appendix:lemma-g}
Assume that the rotation axis $\bm{b}$ of the $3$D rotation
\begin{align}\label{eq:R-b-omega2}
	\bR=\bm{b}\trsp{\bm{b}}+[\bm{b}]_{\times}\sin(\omega)+ (\bI_3-\bm{b}\trsp{\bm{b}})\cos(\omega)
\end{align} 
is given, and we now solve \eqref{eq:R-angle}. Let $a_{i,9}=\trsp{\by_i}\bm{b}\trsp{\bm{b}}\bx_i$, $a_{i,10}=\trsp{\by_i}[\bm{b}]_{\times}\bx_i$ and $a_{i,11}=\trsp{\by_i}(\bI_3-\bm{b}\trsp{\bm{b}})\bx_i$. Then
\begin{align}
	\trsp{\by_i}\bR\bx_i=a_{i,9}+a_{i,10}\sin(\omega) + a_{i,11}\cos(\omega).
\end{align}
Hence the constraint of \eqref{eq:R-3D} can be written as
\begin{align}\label{eq:R-3D-constraint}
	&\norm{\by_i}^2+\norm{\bx_i}^2-c^2\leq 2\trsp{\by_i}\bR\bx_i \\
	\Leftrightarrow& \ \ a_{i,10}\sin(\omega) + a_{i,11} \cos(\omega) \geq a_{i,12}
\end{align}
where we defined $a_{i,12}=(\norm{\by_i}^2+\norm{\bx_i}^2-c^2)/2-a_{i,9}$. There is a unique angle $a_{i,13}\in[0,2\pi)$ satisfying
\begin{align*}
	\cos(a_{i,13}) =\frac{a_{i,11}}{\sqrt{a_{i,10}^2+a_{i,11}^2}}, \ \sin(a_{i,13})=\frac{a_{i,10}}{\sqrt{a_{i,10}^2+a_{i,11}^2}}
\end{align*}
Thus, the constraint $\norm{\by_i - \bR \bx_i}\leq c$ of \eqref{eq:R-3D} is the same as
\begin{align*}
	\cos(\omega - a_{i,13})\geq \max\Bigg\{ \frac{a_{i,12}}{\sqrt{a_{i,10}^2+a_{i,11}^2}},-1\Bigg\}=:a_{i,14}.
\end{align*}
Without loss of generality assume $a_{i,14}\leq 1$, for otherwise we could simply ignore this constraint. Define $a_{i,15}=\arccos(a_{i,14})\in[0,\pi]$. Since $|\omega - a_{i,13}|\in[0,2\pi]$, we consider two cases, namely $|\omega - a_{i,13}|\leq \pi$ and $|\omega - a_{i,13}|> \pi$. In the former case, since $\arccos$ is a decreasing function, the above constraint is equivalent to $|\omega - a_{i,13}|\leq a_{i,15}$. In the later case the above constraint is equivalent to
\begin{align*}
	\cos(2\pi-|\omega - a_{i,13}|)\geq a_{i,14}\Leftrightarrow 2\pi-|\omega - a_{i,13}|\leq a_{i,15}.
\end{align*}
Thus, the constraint $\norm{\by_i - \bR \bx_i}\leq c$ of \eqref{eq:R-3D} requires $\omega\in[0,2\pi]$ to lie in the union of the following intervals.
\begin{align}
	&[a_{i,13}-a_{i,15},\ a_{i,13} + a_{i,15}]\cap [0,2\pi]\\
	&[a_{i,13}- a_{i,15}+2\pi,\ 2\pi]\\
	&[0,\ a_{i,13}+a_{i,15}-2\pi].
\end{align}
In the above, the invalid interval where the right endpoint is smaller than its left endpoint, if any, should  be discarded. To conclude, \eqref{eq:R-angle} can be solved via interval stabbing.

\section{More Experiments}\label{appendix:experiments}
In this section we present more experiments. Besides rotation errors, we will also use another metric for evaluation, that is \textit{success rate}. Given two point clouds as input, an algorithm \textit{succeeds} if it outputs a rotation that has error smaller than a certain threshold; by default the threshold is set to $10$ degree (as in \cite{Yang-T-R2021}) but we will also vary it when appropriate. The success rate is the number of success divided by the total number of experiments that were run. This metric was referred to as \textit{recall} in other related papers (cf. \cite{Choy-CVPR2020}).

Note that, like $\GORE$ \cite{Bustos-ICCV2015} and $\QUASAR$ \cite{Yang-ICCV2019}, $\ARCSplus_{\texttt{OR}}$ can be applied to image stitching, because sometimes the translation is negligible and thus the scene can be justified by a homography $\bH\in\bbR^{3\times 3}$ that involves a pure $3$D rotation $\bR$, i.e., $\bH = \bK \bR\bK^{-1}$ (cf. \cite{Szeliski-2010}); here $\bK\in\bbR^{3\times 3}$ is a matrix of intrinsic camera parameters given by the dataset. However, we noticed that the recent approaches \texttt{MAGSAC++} \cite{Barath-CVPR2019,Barath-CVPR2020,Barath-TPAMI2021} and \texttt{VSAC} \cite{Ivashechkin-ICCV2021} achieved surprising performance and run in fewer than $10$ milliseconds for image stitching, hence we would recommend them for this task.

\subsection{Robustness on Gaussian Point Sets}
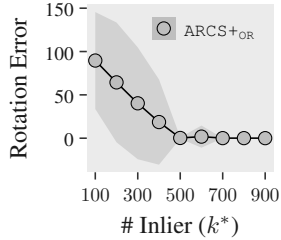
\begin{figure}
	\centering 
\begin{tikzpicture}[x=1pt,y=1pt]
\definecolor{fillColor}{RGB}{255,255,255}
\path[use as bounding box,fill=fillColor,fill opacity=0.00] (0,0) rectangle (112.02, 99.73);
\begin{scope}
\path[clip] (  0.00,  0.00) rectangle (112.02, 99.73);
\definecolor{drawColor}{RGB}{255,255,255}
\definecolor{fillColor}{RGB}{255,255,255}

\path[draw=drawColor,line width= 0.6pt,line join=round,line cap=round,fill=fillColor] (  0.00,  0.00) rectangle (112.02, 99.73);
\end{scope}
\begin{scope}
\path[clip] ( 35.84, 30.69) rectangle (106.52, 94.23);
\definecolor{fillColor}{gray}{0.92}

\path[fill=fillColor] ( 35.84, 30.69) rectangle (106.52, 94.23);
\definecolor{fillColor}{RGB}{190,190,190}

\path[fill=fillColor,fill opacity=0.55] ( 39.05, 91.34) --
	( 47.08, 87.39) --
	( 55.11, 77.98) --
	( 63.15, 65.82) --
	( 71.18, 43.93) --
	( 79.21, 48.38) --
	( 87.24, 43.64) --
	( 95.27, 43.64) --
	(103.31, 43.63) --
	(103.31, 43.60) --
	( 95.27, 43.60) --
	( 87.24, 43.60) --
	( 79.21, 40.00) --
	( 71.18, 43.42) --
	( 63.15, 33.57) --
	( 55.11, 35.63) --
	( 47.08, 42.05) --
	( 39.05, 54.64) --
	cycle;

\path[] ( 39.05, 91.34) --
	( 47.08, 87.39) --
	( 55.11, 77.98) --
	( 63.15, 65.82) --
	( 71.18, 43.93) --
	( 79.21, 48.38) --
	( 87.24, 43.64) --
	( 95.27, 43.64) --
	(103.31, 43.63);

\path[] (103.31, 43.60) --
	( 95.27, 43.60) --
	( 87.24, 43.60) --
	( 79.21, 40.00) --
	( 71.18, 43.42) --
	( 63.15, 33.57) --
	( 55.11, 35.63) --
	( 47.08, 42.05) --
	( 39.05, 54.64);
\definecolor{drawColor}{RGB}{0,0,0}

\path[draw=drawColor,line width= 0.6pt,line join=round] ( 39.05, 72.99) --
	( 47.08, 64.72) --
	( 55.11, 56.80) --
	( 63.15, 49.70) --
	( 71.18, 43.68) --
	( 79.21, 44.19) --
	( 87.24, 43.62) --
	( 95.27, 43.62) --
	(103.31, 43.61);
\definecolor{fillColor}{RGB}{190,190,190}

\path[draw=drawColor,line width= 0.4pt,line join=round,line cap=round,fill=fillColor] (103.31, 43.61) circle (  2.50);

\path[draw=drawColor,line width= 0.4pt,line join=round,line cap=round,fill=fillColor] ( 95.27, 43.62) circle (  2.50);

\path[draw=drawColor,line width= 0.4pt,line join=round,line cap=round,fill=fillColor] ( 87.24, 43.62) circle (  2.50);

\path[draw=drawColor,line width= 0.4pt,line join=round,line cap=round,fill=fillColor] ( 79.21, 44.19) circle (  2.50);

\path[draw=drawColor,line width= 0.4pt,line join=round,line cap=round,fill=fillColor] ( 71.18, 43.68) circle (  2.50);

\path[draw=drawColor,line width= 0.4pt,line join=round,line cap=round,fill=fillColor] ( 63.15, 49.70) circle (  2.50);

\path[draw=drawColor,line width= 0.4pt,line join=round,line cap=round,fill=fillColor] ( 55.11, 56.80) circle (  2.50);

\path[draw=drawColor,line width= 0.4pt,line join=round,line cap=round,fill=fillColor] ( 47.08, 64.72) circle (  2.50);

\path[draw=drawColor,line width= 0.4pt,line join=round,line cap=round,fill=fillColor] ( 39.05, 72.99) circle (  2.50);
\end{scope}
\begin{scope}
\path[clip] (  0.00,  0.00) rectangle (112.02, 99.73);
\definecolor{drawColor}{gray}{0.10}

\node[text=drawColor,anchor=base east,inner sep=0pt, outer sep=0pt, scale=  0.73] at ( 30.89, 40.56) {0};

\node[text=drawColor,anchor=base east,inner sep=0pt, outer sep=0pt, scale=  0.73] at ( 30.89, 56.97) {50};

\node[text=drawColor,anchor=base east,inner sep=0pt, outer sep=0pt, scale=  0.73] at ( 30.89, 73.37) {100};

\node[text=drawColor,anchor=base east,inner sep=0pt, outer sep=0pt, scale=  0.73] at ( 30.89, 89.78) {150};
\end{scope}
\begin{scope}
\path[clip] (  0.00,  0.00) rectangle (112.02, 99.73);
\definecolor{drawColor}{gray}{0.20}

\path[draw=drawColor,line width= 0.6pt,line join=round] ( 33.09, 43.59) --
	( 35.84, 43.59);

\path[draw=drawColor,line width= 0.6pt,line join=round] ( 33.09, 60.00) --
	( 35.84, 60.00);

\path[draw=drawColor,line width= 0.6pt,line join=round] ( 33.09, 76.40) --
	( 35.84, 76.40);

\path[draw=drawColor,line width= 0.6pt,line join=round] ( 33.09, 92.81) --
	( 35.84, 92.81);
\end{scope}
\begin{scope}
\path[clip] (  0.00,  0.00) rectangle (112.02, 99.73);
\definecolor{drawColor}{gray}{0.20}

\path[draw=drawColor,line width= 0.6pt,line join=round] (103.31, 27.94) --
	(103.31, 30.69);

\path[draw=drawColor,line width= 0.6pt,line join=round] ( 87.24, 27.94) --
	( 87.24, 30.69);

\path[draw=drawColor,line width= 0.6pt,line join=round] ( 71.18, 27.94) --
	( 71.18, 30.69);

\path[draw=drawColor,line width= 0.6pt,line join=round] ( 55.11, 27.94) --
	( 55.11, 30.69);

\path[draw=drawColor,line width= 0.6pt,line join=round] ( 39.05, 27.94) --
	( 39.05, 30.69);
\end{scope}
\begin{scope}
\path[clip] (  0.00,  0.00) rectangle (112.02, 99.73);
\definecolor{drawColor}{gray}{0.10}

\node[text=drawColor,anchor=base,inner sep=0pt, outer sep=0pt, scale=  0.73] at (103.31, 19.68) {$900$};

\node[text=drawColor,anchor=base,inner sep=0pt, outer sep=0pt, scale=  0.73] at ( 87.24, 19.68) {$700$};

\node[text=drawColor,anchor=base,inner sep=0pt, outer sep=0pt, scale=  0.73] at ( 71.18, 19.68) {$500$};

\node[text=drawColor,anchor=base,inner sep=0pt, outer sep=0pt, scale=  0.73] at ( 55.11, 19.68) {$300$};

\node[text=drawColor,anchor=base,inner sep=0pt, outer sep=0pt, scale=  0.73] at ( 39.05, 19.68) {$100$};
\end{scope}
\begin{scope}
\path[clip] (  0.00,  0.00) rectangle (112.02, 99.73);
\definecolor{drawColor}{gray}{0.10}

\node[text=drawColor,anchor=base,inner sep=0pt, outer sep=0pt, scale=  0.92] at ( 71.18,  7.64) {\# Inlier ($k^*$)};
\end{scope}
\begin{scope}
\path[clip] (  0.00,  0.00) rectangle (112.02, 99.73);
\definecolor{drawColor}{gray}{0.10}

\node[text=drawColor,rotate= 90.00,anchor=base,inner sep=0pt, outer sep=0pt, scale=  0.92] at ( 13.08, 62.46) {Rotation Error};
\end{scope}
\begin{scope}
\path[clip] (  0.00,  0.00) rectangle (112.02, 99.73);
\definecolor{fillColor}{RGB}{190,190,190}

\path[fill=fillColor] ( 60.60, 80.78) rectangle ( 69.30, 89.48);
\end{scope}
\begin{scope}
\path[clip] (  0.00,  0.00) rectangle (112.02, 99.73);
\definecolor{drawColor}{RGB}{0,0,0}
\definecolor{fillColor}{RGB}{190,190,190}

\path[draw=drawColor,line width= 0.4pt,line join=round,line cap=round,fill=fillColor] ( 64.95, 85.13) circle (  2.50);
\end{scope}
\begin{scope}
\path[clip] (  0.00,  0.00) rectangle (112.02, 99.73);
\definecolor{drawColor}{RGB}{0,0,0}

\node[text=drawColor,anchor=base west,inner sep=0pt, outer sep=0pt, scale=  0.69] at ( 72.61, 82.29) {$\texttt{ARCS+}_{\texttt{OR}}$};
\end{scope}
\end{tikzpicture}
	\caption{Performance of $\ARCSplus_{\texttt{OR}}$ on Gaussian point sets with $\ell=10^7$ point pairs, $\sigma=0.01$, $100$ trials. $\ARCSplus_{\texttt{OR}}$ works well until there are fewer than $500/10^7=0.005\%$ inliers. \label{fig:experiments_RRS_scalability_Gaussian}}
\end{figure}

In previous synthetic experiments on robust rotation search (Table \ref{table:RRS-scalibility}), we generated data by ensuring that each point pair $\by_i$ and $\bx_i$ has nearly the same norm. This is for fair comparison of the methods, and it might not be true in practice. Here we show that, without this norm constraint, $\ARCSplus_{\texttt{OR}}$ can tolerate even more outliers. In the experiment here we generated point sets $\{\by_i,\bx_i\}_{i=1}^\ell$ as in Table \ref{table:RRS-scalibility} except without the norm constraint. Then, we first perform a simple step, that removes all point pairs $(\by_i,\bx_i)$ which satisfy $\big| \norm{\by_i}-\norm{\bx_i} \big| > c$, and then feed the remaining points to $\ARCSplus_{\texttt{OR}}$. We reported the results in Figure \ref{fig:experiments_RRS_scalability_Gaussian}, where we observed that $\ARCSplus_{\texttt{OR}}$ worked well until there are fewer than $500/10^7=0.005\%$ inliers.

\subsection{Sensitivity to The Ground-Truth Rotation}
\begin{figure}
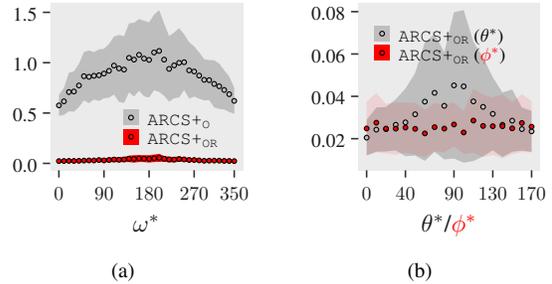

	\centering
	\subfloat[]{\input{./Figures/experiments_VaryingOmega.tex} \label{fig:experiments_VaryingOmega} }
	\subfloat[]{\input{./Figures/experiments_VaryingThetaPhi.tex} \label{fig:experiments_VaryingThetaPhi} }
	\caption{Average rotation errors (in degrees) and standard deviations with respect to the ground-truth rotation angle $\omega^*$ and axis $\bm{b}^*=\trsp{[\sin(\theta^*)\cos(\phi^*),\ \sin(\theta^*)\sin(\phi^*),\ \cos(\theta^*)]}$. Experiments run with $100$ trials, $\ell=10^5$, $k^*=1000$, $\sigma=0.01$. \label{fig:experiments_Sensitivity}}
\end{figure}

In Figure \ref{fig:experiments_Sensitivity} we presented the sensitivity of  $\ARCSplus_{\texttt{OR}}$ to the ground-truth rotation $\bR^*$. Figure \ref{fig:experiments_VaryingOmega} depicted that, with the ground-truth rotation angle $\omega^*$ changing, the mean estimation error of $\ARCSplus_{\texttt{O}}$  varied from $0.5$ to $1$, while the standard derivation ranged from $0$ to $0.5$. One the other hand, $\ARCSplus_\texttt{R}$ refined the estimate from $\ARCSplus_{\texttt{O}}$, so that their combination $\ARCSplus_{\texttt{OR}}$ had much smaller mean error and standard derivation, nearly imperceivable from Figure \ref{fig:experiments_VaryingOmega}. In Figure \ref{fig:experiments_VaryingThetaPhi} we kept $\omega^*$ fixed and presented how the errors of $\ARCSplus_{\texttt{OR}}$ vary with $\theta^*$ and $\phi^*$, the two angles for the ground-truth rotation axis $\bm{b}^*$; we fixed one of them when varying the other. We observed that $\ARCSplus_{\texttt{OR}}$ is immune to the change of $\phi^*$, as it consistently gave about $0.02$ errors  and  $0.01$ standard derivation. This is expected as $\ARCSplus_\texttt{O}$ selects from multiple $\phi_j$'s a best one based on consensus maximization. On the other hand, varying $\phi^*$ does make an impact on the performance of $\ARCSplus_{\texttt{OR}}$; the standard deviation reached its peak, around $0.04$, when $\theta^*=\pi/4$. Theoretically justifying the phenomenon presented here can be an interesting future work.

\begin{figure}
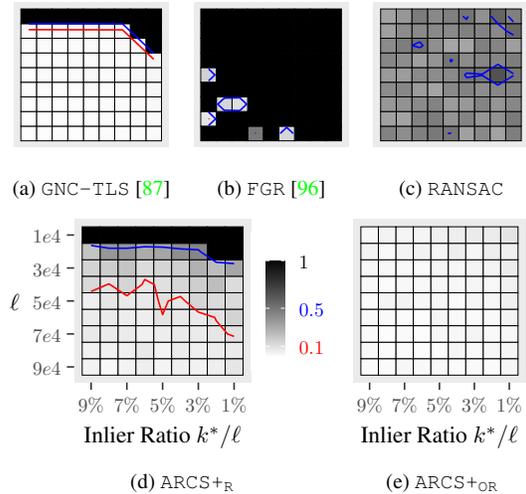

	\centering
	\subfloat[$\GNCTLS$ \cite{Yang-RA-L2020} ]{\input{./Figures/phase_transition_GNC_TLS_10.tex} \label{fig:phase_transition_GNC_TLS} }
	\hspace{-0.5cm}
	\subfloat[$\FGR$ \cite{Zhou-ECCV2016} ]{\input{./Figures/phase_transition_FGR_10.tex} \label{fig:phase_transition_FGR} }
	\hspace{-0.5cm}
	\subfloat[$\RANSAC$ ]{\input{./Figures/phase_transition_RANSAC_10.tex} \label{fig:phase_transition_RANSAC} }
	
	\subfloat[$\ARCSplus_\texttt{R}$]{\input{./Figures/phase_transition_RSGD_10.tex} \label{fig:phase_transition_RSGD} }
	\hspace{-0.7cm}
	\subfloat[$\ARCSplus_{\texttt{OR}}$]{\input{./Figures/phase_transition_Stabber3D_RSGD_10.tex} \label{fig:phase_transition_Stabber3D_RSGD} }
	\caption{Average rotation errors in degrees of different robust rotation search approaches on medium-scale synthetic $3$D point sets of sizes varying from $10^4$ to $9\times 10^4$ with inlier ratios ranging from $1\%$ to $9\%$. Experiments run with $50$ trials, $\sigma=0.01$ fixed.\label{fig:phase_transition}}
\end{figure}

\subsection{Phase Transition}
In Figure \ref{fig:phase_transition}, we showed the performances of algorithms for different inlier ratios $k^*/\ell$ and different number $\ell$ of points; whiter means smaller errors and errors larger than $1$ were truncated to $1$. The major point we would like to clarify here is that, whether or not an algorithm can tolerate say $99\%$ outliers might depend on the total number of points (cf. Figures \ref{fig:phase_transition_GNC_TLS} and \ref{fig:phase_transition_RSGD}), so sentences such as ``our algorithm can tolerate $99\%$ outliers'' might be inaccurate, even though such description has been widely used in recent papers. Indeed, no algorithm can tolerate $99\%$ outliers if $\ell=100$. Also, as mentioned in \S \ref{section:R}, one theorem of \cite{Bohorquez-arXiv2020v3} has  shed light on this phenomenon. The other important observation here is that, $\GNCTLS$ achieved higher accuracy than $\ARCSplus_{\texttt{R}}$, although they exhibited nearly the same breaking down points. One reason is that $\GNCTLS$ takes advantage of the inlier threshold $c$ as extra information. This empirically suggests that combing $\ARCSplus_\texttt{O}$ and $\GNCTLS$ might further boost the performance for robust rotation search.

\subsection{Robustness to Noise}
\begin{figure}
	\centering 
\begin{tikzpicture}[x=1pt,y=1pt]
\definecolor{fillColor}{RGB}{255,255,255}
\path[use as bounding box,fill=fillColor,fill opacity=0.00] (0,0) rectangle (148.15, 93.95);
\begin{scope}
\path[clip] (  0.00,  0.00) rectangle (148.15, 93.95);
\definecolor{drawColor}{RGB}{255,255,255}
\definecolor{fillColor}{RGB}{255,255,255}

\path[draw=drawColor,line width= 0.6pt,line join=round,line cap=round,fill=fillColor] (  0.00,  0.00) rectangle (148.15, 93.95);
\end{scope}
\begin{scope}
\path[clip] ( 40.18, 27.77) rectangle (108.63, 88.45);
\definecolor{fillColor}{gray}{0.92}

\path[fill=fillColor] ( 40.18, 27.77) rectangle (108.63, 88.45);
\definecolor{drawColor}{RGB}{0,0,0}

\path[draw=drawColor,line width= 0.6pt,line join=round] ( 43.29, 30.53) --
	( 50.21, 37.03) --
	( 57.12, 43.50) --
	( 64.04, 49.90) --
	( 70.95, 56.20) --
	( 77.86, 62.42) --
	( 84.78, 68.43) --
	( 91.69, 74.44) --
	( 98.61, 80.15) --
	(105.52, 85.69);

\path[draw=drawColor,line width= 0.4pt,line join=round,line cap=round] ( 41.87, 29.10) rectangle ( 44.72, 31.96);

\path[draw=drawColor,line width= 0.4pt,line join=round,line cap=round] ( 48.78, 35.60) rectangle ( 51.63, 38.45);

\path[draw=drawColor,line width= 0.4pt,line join=round,line cap=round] ( 55.69, 42.07) rectangle ( 58.55, 44.93);

\path[draw=drawColor,line width= 0.4pt,line join=round,line cap=round] ( 62.61, 48.47) rectangle ( 65.46, 51.32);

\path[draw=drawColor,line width= 0.4pt,line join=round,line cap=round] ( 69.52, 54.77) rectangle ( 72.38, 57.62);

\path[draw=drawColor,line width= 0.4pt,line join=round,line cap=round] ( 76.44, 60.99) rectangle ( 79.29, 63.84);

\path[draw=drawColor,line width= 0.4pt,line join=round,line cap=round] ( 83.35, 67.00) rectangle ( 86.21, 69.86);

\path[draw=drawColor,line width= 0.4pt,line join=round,line cap=round] ( 90.27, 73.01) rectangle ( 93.12, 75.87);

\path[draw=drawColor,line width= 0.4pt,line join=round,line cap=round] ( 97.18, 78.73) rectangle (100.03, 81.58);

\path[draw=drawColor,line width= 0.4pt,line join=round,line cap=round] (104.10, 84.27) rectangle (106.95, 87.12);
\definecolor{drawColor}{RGB}{160,32,240}

\path[draw=drawColor,draw opacity=0.50,line width= 0.6pt,line join=round] ( 43.29, 30.53) --
	( 50.21, 37.03) --
	( 57.12, 43.50) --
	( 64.04, 49.90) --
	( 70.95, 56.20) --
	( 77.86, 62.42) --
	( 84.78, 68.43) --
	( 91.69, 74.44) --
	( 98.61, 80.15) --
	(105.52, 85.69);

\path[draw=drawColor,draw opacity=0.50,line width= 0.4pt,line join=round,line cap=round] ( 43.29, 32.75) --
	( 45.21, 29.42) --
	( 41.37, 29.42) --
	( 43.29, 32.75);

\path[draw=drawColor,draw opacity=0.50,line width= 0.4pt,line join=round,line cap=round] ( 50.21, 39.25) --
	( 52.13, 35.92) --
	( 48.29, 35.92) --
	( 50.21, 39.25);

\path[draw=drawColor,draw opacity=0.50,line width= 0.4pt,line join=round,line cap=round] ( 57.12, 45.72) --
	( 59.04, 42.39) --
	( 55.20, 42.39) --
	( 57.12, 45.72);

\path[draw=drawColor,draw opacity=0.50,line width= 0.4pt,line join=round,line cap=round] ( 64.04, 52.11) --
	( 65.96, 48.79) --
	( 62.11, 48.79) --
	( 64.04, 52.11);

\path[draw=drawColor,draw opacity=0.50,line width= 0.4pt,line join=round,line cap=round] ( 70.95, 58.41) --
	( 72.87, 55.09) --
	( 69.03, 55.09) --
	( 70.95, 58.41);

\path[draw=drawColor,draw opacity=0.50,line width= 0.4pt,line join=round,line cap=round] ( 77.86, 64.64) --
	( 79.79, 61.31) --
	( 75.94, 61.31) --
	( 77.86, 64.64);

\path[draw=drawColor,draw opacity=0.50,line width= 0.4pt,line join=round,line cap=round] ( 84.78, 70.65) --
	( 86.70, 67.32) --
	( 82.86, 67.32) --
	( 84.78, 70.65);

\path[draw=drawColor,draw opacity=0.50,line width= 0.4pt,line join=round,line cap=round] ( 91.69, 76.66) --
	( 93.62, 73.33) --
	( 89.77, 73.33) --
	( 91.69, 76.66);

\path[draw=drawColor,draw opacity=0.50,line width= 0.4pt,line join=round,line cap=round] ( 98.61, 82.37) --
	(100.53, 79.05) --
	( 96.69, 79.05) --
	( 98.61, 82.37);

\path[draw=drawColor,draw opacity=0.50,line width= 0.4pt,line join=round,line cap=round] (105.52, 87.91) --
	(107.44, 84.58) --
	(103.60, 84.58) --
	(105.52, 87.91);
\end{scope}
\begin{scope}
\path[clip] (  0.00,  0.00) rectangle (148.15, 93.95);
\definecolor{drawColor}{RGB}{0,0,0}

\node[text=drawColor,anchor=base east,inner sep=0pt, outer sep=0pt, scale=  0.74] at ( 35.23, 38.84) {$10^5$};

\node[text=drawColor,anchor=base east,inner sep=0pt, outer sep=0pt, scale=  0.74] at ( 35.23, 56.27) {$2\times 10^5$};

\node[text=drawColor,anchor=base east,inner sep=0pt, outer sep=0pt, scale=  0.74] at ( 35.23, 73.69) {$3\times 10^5$};
\end{scope}
\begin{scope}
\path[clip] (  0.00,  0.00) rectangle (148.15, 93.95);
\definecolor{drawColor}{gray}{0.20}

\path[draw=drawColor,line width= 0.6pt,line join=round] ( 37.43, 41.40) --
	( 40.18, 41.40);

\path[draw=drawColor,line width= 0.6pt,line join=round] ( 37.43, 58.82) --
	( 40.18, 58.82);

\path[draw=drawColor,line width= 0.6pt,line join=round] ( 37.43, 76.25) --
	( 40.18, 76.25);
\end{scope}
\begin{scope}
\path[clip] (  0.00,  0.00) rectangle (148.15, 93.95);
\definecolor{drawColor}{gray}{0.20}

\path[draw=drawColor,line width= 0.6pt,line join=round] (108.63, 37.92) --
	(111.38, 37.92);

\path[draw=drawColor,line width= 0.6pt,line join=round] (108.63, 51.82) --
	(111.38, 51.82);

\path[draw=drawColor,line width= 0.6pt,line join=round] (108.63, 65.79) --
	(111.38, 65.79);

\path[draw=drawColor,line width= 0.6pt,line join=round] (108.63, 79.77) --
	(111.38, 79.77);
\end{scope}
\begin{scope}
\path[clip] (  0.00,  0.00) rectangle (148.15, 93.95);
\definecolor{drawColor}{RGB}{160,32,240}

\node[text=drawColor,anchor=base west,inner sep=0pt, outer sep=0pt, scale=  0.74] at (113.58, 35.36) {$10\%$};

\node[text=drawColor,anchor=base west,inner sep=0pt, outer sep=0pt, scale=  0.74] at (113.58, 49.27) {$20\%$};

\node[text=drawColor,anchor=base west,inner sep=0pt, outer sep=0pt, scale=  0.74] at (113.58, 63.24) {$30\%$};

\node[text=drawColor,anchor=base west,inner sep=0pt, outer sep=0pt, scale=  0.74] at (113.58, 77.21) {$40\%$};
\end{scope}
\begin{scope}
\path[clip] (  0.00,  0.00) rectangle (148.15, 93.95);
\definecolor{drawColor}{gray}{0.20}

\path[draw=drawColor,line width= 0.6pt,line join=round] ( 43.29, 25.02) --
	( 43.29, 27.77);

\path[draw=drawColor,line width= 0.6pt,line join=round] ( 64.04, 25.02) --
	( 64.04, 27.77);

\path[draw=drawColor,line width= 0.6pt,line join=round] ( 84.78, 25.02) --
	( 84.78, 27.77);

\path[draw=drawColor,line width= 0.6pt,line join=round] (105.52, 25.02) --
	(105.52, 27.77);
\end{scope}
\begin{scope}
\path[clip] (  0.00,  0.00) rectangle (148.15, 93.95);
\definecolor{drawColor}{gray}{0.30}

\node[text=drawColor,anchor=base,inner sep=0pt, outer sep=0pt, scale=  0.82] at ( 43.29, 17.14) {$1\%$};

\node[text=drawColor,anchor=base,inner sep=0pt, outer sep=0pt, scale=  0.82] at ( 64.04, 17.14) {$4\%$};

\node[text=drawColor,anchor=base,inner sep=0pt, outer sep=0pt, scale=  0.82] at ( 84.78, 17.14) {$7\%$};

\node[text=drawColor,anchor=base,inner sep=0pt, outer sep=0pt, scale=  0.82] at (105.52, 17.14) {$10\%$};
\end{scope}
\begin{scope}
\path[clip] (  0.00,  0.00) rectangle (148.15, 93.95);
\definecolor{drawColor}{RGB}{0,0,0}

\node[text=drawColor,anchor=base,inner sep=0pt, outer sep=0pt, scale=  0.82] at ( 74.41,  7.10) {Noise Level ($\sigma$)};
\end{scope}
\begin{scope}
\path[clip] (  0.00,  0.00) rectangle (148.15, 93.95);
\definecolor{drawColor}{RGB}{0,0,0}

\node[text=drawColor,anchor=base east,inner sep=0pt, outer sep=0pt, scale=  0.82] at (  8.94, 55.27) {$\ell$};
\end{scope}
\begin{scope}
\path[clip] (  0.00,  0.00) rectangle (148.15, 93.95);
\definecolor{drawColor}{RGB}{160,32,240}

\node[text=drawColor,anchor=base east,inner sep=0pt, outer sep=0pt, scale=  0.88] at (142.65, 55.08) {$\frac{\ell}{mn}$};
\end{scope}
\end{tikzpicture}
	\caption{Sensitivity of $\ARCSplus_{\texttt{N}}$ to noise: $\ell$ increases linearly as $\sigma$ grows. $100$ trials, $m=1000,n=800,k^*=200$, $c=5.54\sigma$.  \label{fig:experiments_ARCSN_noise}}
\end{figure}
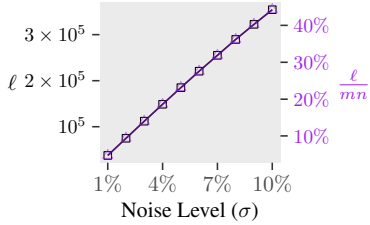

Figure \ref{fig:experiments_ARCSN_noise} showed that $\ARCSplus_{\texttt{N}}$ is sensitive to noise: In particular, for $c=5.54\sigma$ fixed, the number $\ell$ of output point pairs grows proportionally as a linear function of $\sigma$. A similar phenomenon can be found in \cite{Yang-T-R2021} and its follow-up works: Higher noise leads to denser graphs, and thus to intractable maximal clique problems (recall Section \ref{section:intro}).

\begin{figure}
	\centering
	\subfloat[$k^*/\ell = 100/1000$]{
\begin{tikzpicture}[x=1pt,y=1pt]
\definecolor{fillColor}{RGB}{255,255,255}
\path[use as bounding box,fill=fillColor,fill opacity=0.00] (0,0) rectangle (101.18, 95.40);
\begin{scope}
\path[clip] (  0.00,  0.00) rectangle (101.18, 95.40);
\definecolor{drawColor}{RGB}{255,255,255}
\definecolor{fillColor}{RGB}{255,255,255}

\path[draw=drawColor,line width= 0.6pt,line join=round,line cap=round,fill=fillColor] (  0.00,  0.00) rectangle (101.18, 95.40);
\end{scope}
\begin{scope}
\path[clip] ( 29.83, 28.26) rectangle ( 95.68, 89.90);
\definecolor{fillColor}{gray}{0.92}

\path[fill=fillColor] ( 29.83, 28.26) rectangle ( 95.68, 89.90);
\definecolor{drawColor}{RGB}{0,0,0}

\path[draw=drawColor,line width= 0.6pt,line join=round] ( 32.82, 31.08) --
	( 39.47, 31.38) --
	( 46.13, 31.81) --
	( 52.78, 32.31) --
	( 59.43, 32.67) --
	( 66.08, 33.34) --
	( 72.73, 34.52) --
	( 79.38, 35.90) --
	( 86.03, 37.51) --
	( 92.68, 37.39);

\path[draw=drawColor,line width= 0.6pt,line join=round] ( 32.82, 32.54) --
	( 39.47, 34.44) --
	( 46.13, 35.10) --
	( 52.78, 38.28) --
	( 59.43, 41.22) --
	( 66.08, 42.85) --
	( 72.73, 43.12) --
	( 79.38, 45.46) --
	( 86.03, 51.42) --
	( 92.68, 50.73);

\path[draw=drawColor,line width= 0.6pt,line join=round] ( 32.82, 36.53) --
	( 39.47, 44.17) --
	( 46.13, 54.47) --
	( 52.78, 56.22) --
	( 59.43, 61.55) --
	( 66.08, 66.32) --
	( 72.73, 73.53) --
	( 79.38, 77.81) --
	( 86.03, 87.09) --
	( 92.68, 85.27);

\path[draw=drawColor,line width= 0.6pt,line join=round] ( 32.82, 31.06) --
	( 39.47, 31.23) --
	( 46.13, 31.67) --
	( 52.78, 31.74) --
	( 59.43, 31.93) --
	( 66.08, 32.36) --
	( 72.73, 33.03) --
	( 79.38, 33.56) --
	( 86.03, 34.31) --
	( 92.68, 34.42);

\path[draw=drawColor,line width= 0.4pt,line join=round,line cap=round] ( 32.82, 31.08) circle (  1.96);

\path[draw=drawColor,line width= 0.4pt,line join=round,line cap=round] ( 39.47, 31.38) circle (  1.96);

\path[draw=drawColor,line width= 0.4pt,line join=round,line cap=round] ( 46.13, 31.81) circle (  1.96);

\path[draw=drawColor,line width= 0.4pt,line join=round,line cap=round] ( 52.78, 32.31) circle (  1.96);

\path[draw=drawColor,line width= 0.4pt,line join=round,line cap=round] ( 59.43, 32.67) circle (  1.96);

\path[draw=drawColor,line width= 0.4pt,line join=round,line cap=round] ( 66.08, 33.34) circle (  1.96);

\path[draw=drawColor,line width= 0.4pt,line join=round,line cap=round] ( 72.73, 34.52) circle (  1.96);

\path[draw=drawColor,line width= 0.4pt,line join=round,line cap=round] ( 79.38, 35.90) circle (  1.96);

\path[draw=drawColor,line width= 0.4pt,line join=round,line cap=round] ( 86.03, 37.51) circle (  1.96);

\path[draw=drawColor,line width= 0.4pt,line join=round,line cap=round] ( 92.68, 37.39) circle (  1.96);

\path[draw=drawColor,line width= 0.4pt,line join=round,line cap=round] ( 32.82, 35.59) --
	( 35.46, 31.02) --
	( 30.18, 31.02) --
	( 32.82, 35.59);

\path[draw=drawColor,line width= 0.4pt,line join=round,line cap=round] ( 39.47, 37.49) --
	( 42.12, 32.92) --
	( 36.83, 32.92) --
	( 39.47, 37.49);

\path[draw=drawColor,line width= 0.4pt,line join=round,line cap=round] ( 46.13, 38.15) --
	( 48.77, 33.57) --
	( 43.48, 33.57) --
	( 46.13, 38.15);

\path[draw=drawColor,line width= 0.4pt,line join=round,line cap=round] ( 52.78, 41.33) --
	( 55.42, 36.76) --
	( 50.13, 36.76) --
	( 52.78, 41.33);

\path[draw=drawColor,line width= 0.4pt,line join=round,line cap=round] ( 59.43, 44.27) --
	( 62.07, 39.69) --
	( 56.79, 39.69) --
	( 59.43, 44.27);

\path[draw=drawColor,line width= 0.4pt,line join=round,line cap=round] ( 66.08, 45.90) --
	( 68.72, 41.33) --
	( 63.44, 41.33) --
	( 66.08, 45.90);

\path[draw=drawColor,line width= 0.4pt,line join=round,line cap=round] ( 72.73, 46.18) --
	( 75.37, 41.60) --
	( 70.09, 41.60) --
	( 72.73, 46.18);

\path[draw=drawColor,line width= 0.4pt,line join=round,line cap=round] ( 79.38, 48.51) --
	( 82.02, 43.94) --
	( 76.74, 43.94) --
	( 79.38, 48.51);

\path[draw=drawColor,line width= 0.4pt,line join=round,line cap=round] ( 86.03, 54.47) --
	( 88.68, 49.90) --
	( 83.39, 49.90) --
	( 86.03, 54.47);

\path[draw=drawColor,line width= 0.4pt,line join=round,line cap=round] ( 92.68, 53.79) --
	( 95.33, 49.21) --
	( 90.04, 49.21) --
	( 92.68, 53.79);

\path[draw=drawColor,line width= 0.4pt,line join=round,line cap=round] ( 30.05, 36.53) --
	( 32.82, 39.30) --
	( 35.60, 36.53) --
	( 32.82, 33.75) --
	( 30.05, 36.53);

\path[draw=drawColor,line width= 0.4pt,line join=round,line cap=round] ( 36.70, 44.17) --
	( 39.47, 46.95) --
	( 42.25, 44.17) --
	( 39.47, 41.40) --
	( 36.70, 44.17);

\path[draw=drawColor,line width= 0.4pt,line join=round,line cap=round] ( 43.35, 54.47) --
	( 46.13, 57.25) --
	( 48.90, 54.47) --
	( 46.13, 51.70) --
	( 43.35, 54.47);

\path[draw=drawColor,line width= 0.4pt,line join=round,line cap=round] ( 50.00, 56.22) --
	( 52.78, 59.00) --
	( 55.55, 56.22) --
	( 52.78, 53.45) --
	( 50.00, 56.22);

\path[draw=drawColor,line width= 0.4pt,line join=round,line cap=round] ( 56.65, 61.55) --
	( 59.43, 64.32) --
	( 62.20, 61.55) --
	( 59.43, 58.77) --
	( 56.65, 61.55);

\path[draw=drawColor,line width= 0.4pt,line join=round,line cap=round] ( 63.30, 66.32) --
	( 66.08, 69.09) --
	( 68.85, 66.32) --
	( 66.08, 63.54) --
	( 63.30, 66.32);

\path[draw=drawColor,line width= 0.4pt,line join=round,line cap=round] ( 69.96, 73.53) --
	( 72.73, 76.30) --
	( 75.51, 73.53) --
	( 72.73, 70.76) --
	( 69.96, 73.53);

\path[draw=drawColor,line width= 0.4pt,line join=round,line cap=round] ( 76.61, 77.81) --
	( 79.38, 80.59) --
	( 82.16, 77.81) --
	( 79.38, 75.04) --
	( 76.61, 77.81);

\path[draw=drawColor,line width= 0.4pt,line join=round,line cap=round] ( 83.26, 87.09) --
	( 86.03, 89.87) --
	( 88.81, 87.09) --
	( 86.03, 84.32) --
	( 83.26, 87.09);

\path[draw=drawColor,line width= 0.4pt,line join=round,line cap=round] ( 89.91, 85.27) --
	( 92.68, 88.05) --
	( 95.46, 85.27) --
	( 92.68, 82.50) --
	( 89.91, 85.27);
\definecolor{fillColor}{RGB}{255,0,0}

\path[draw=drawColor,line width= 0.4pt,line join=round,line cap=round,fill=fillColor] ( 31.08, 29.32) rectangle ( 34.56, 32.80);

\path[draw=drawColor,line width= 0.4pt,line join=round,line cap=round,fill=fillColor] ( 37.73, 29.49) rectangle ( 41.21, 32.97);

\path[draw=drawColor,line width= 0.4pt,line join=round,line cap=round,fill=fillColor] ( 44.39, 29.93) rectangle ( 47.86, 33.41);

\path[draw=drawColor,line width= 0.4pt,line join=round,line cap=round,fill=fillColor] ( 51.04, 30.00) rectangle ( 54.52, 33.48);

\path[draw=drawColor,line width= 0.4pt,line join=round,line cap=round,fill=fillColor] ( 57.69, 30.19) rectangle ( 61.17, 33.67);

\path[draw=drawColor,line width= 0.4pt,line join=round,line cap=round,fill=fillColor] ( 64.34, 30.62) rectangle ( 67.82, 34.10);

\path[draw=drawColor,line width= 0.4pt,line join=round,line cap=round,fill=fillColor] ( 70.99, 31.29) rectangle ( 74.47, 34.77);

\path[draw=drawColor,line width= 0.4pt,line join=round,line cap=round,fill=fillColor] ( 77.64, 31.82) rectangle ( 81.12, 35.29);

\path[draw=drawColor,line width= 0.4pt,line join=round,line cap=round,fill=fillColor] ( 84.29, 32.57) rectangle ( 87.77, 36.05);

\path[draw=drawColor,line width= 0.4pt,line join=round,line cap=round,fill=fillColor] ( 90.95, 32.69) rectangle ( 94.42, 36.16);
\end{scope}
\begin{scope}
\path[clip] (  0.00,  0.00) rectangle (101.18, 95.40);
\definecolor{drawColor}{gray}{0.10}

\node[text=drawColor,anchor=base east,inner sep=0pt, outer sep=0pt, scale=  0.73] at ( 24.88, 27.70) {0};

\node[text=drawColor,anchor=base east,inner sep=0pt, outer sep=0pt, scale=  0.73] at ( 24.88, 45.69) {5};

\node[text=drawColor,anchor=base east,inner sep=0pt, outer sep=0pt, scale=  0.73] at ( 24.88, 63.69) {10};

\node[text=drawColor,anchor=base east,inner sep=0pt, outer sep=0pt, scale=  0.73] at ( 24.88, 81.68) {15};
\end{scope}
\begin{scope}
\path[clip] (  0.00,  0.00) rectangle (101.18, 95.40);
\definecolor{drawColor}{gray}{0.20}

\path[draw=drawColor,line width= 0.6pt,line join=round] ( 27.08, 30.73) --
	( 29.83, 30.73);

\path[draw=drawColor,line width= 0.6pt,line join=round] ( 27.08, 48.72) --
	( 29.83, 48.72);

\path[draw=drawColor,line width= 0.6pt,line join=round] ( 27.08, 66.72) --
	( 29.83, 66.72);

\path[draw=drawColor,line width= 0.6pt,line join=round] ( 27.08, 84.71) --
	( 29.83, 84.71);
\end{scope}
\begin{scope}
\path[clip] (  0.00,  0.00) rectangle (101.18, 95.40);
\definecolor{drawColor}{gray}{0.20}

\path[draw=drawColor,line width= 0.6pt,line join=round] ( 32.82, 25.51) --
	( 32.82, 28.26);

\path[draw=drawColor,line width= 0.6pt,line join=round] ( 52.78, 25.51) --
	( 52.78, 28.26);

\path[draw=drawColor,line width= 0.6pt,line join=round] ( 72.73, 25.51) --
	( 72.73, 28.26);

\path[draw=drawColor,line width= 0.6pt,line join=round] ( 92.68, 25.51) --
	( 92.68, 28.26);
\end{scope}
\begin{scope}
\path[clip] (  0.00,  0.00) rectangle (101.18, 95.40);
\definecolor{drawColor}{gray}{0.10}

\node[text=drawColor,anchor=base,inner sep=0pt, outer sep=0pt, scale=  0.73] at ( 32.82, 17.25) {$1\%$};

\node[text=drawColor,anchor=base,inner sep=0pt, outer sep=0pt, scale=  0.73] at ( 52.78, 17.25) {$4\%$};

\node[text=drawColor,anchor=base,inner sep=0pt, outer sep=0pt, scale=  0.73] at ( 72.73, 17.25) {$7\%$};

\node[text=drawColor,anchor=base,inner sep=0pt, outer sep=0pt, scale=  0.73] at ( 92.68, 17.25) {$10\%$};
\end{scope}
\begin{scope}
\path[clip] (  0.00,  0.00) rectangle (101.18, 95.40);
\definecolor{drawColor}{gray}{0.10}

\node[text=drawColor,anchor=base,inner sep=0pt, outer sep=0pt, scale=  0.69] at ( 62.75,  7.10) {Noise Level ($\sigma$)};
\end{scope}
\begin{scope}
\path[clip] (  0.00,  0.00) rectangle (101.18, 95.40);
\definecolor{drawColor}{gray}{0.10}

\node[text=drawColor,rotate= 90.00,anchor=base,inner sep=0pt, outer sep=0pt, scale=  0.76] at ( 11.75, 59.08) {Rotation Error};
\end{scope}
\end{tikzpicture}  \label{fig:experiments_ARCSO_noise_error}
	}
	\subfloat[$k^*/\ell = 100/1000$]{
		\input{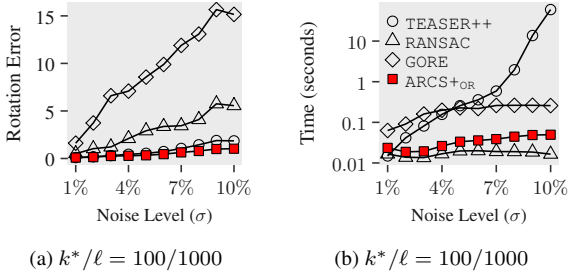}  \label{fig:experiments_ARCSO_noise_time}
	}
	\caption{Robustness of various methods to noise. $20$ trials. \label{fig:experiments_ARCSO_noise} }
\end{figure}

However, $\ARCSplus_{\texttt{O}}$ and $\ARCSplus_{\texttt{OR}}$ behave reasonably well as noise varies. This was shown in Figure \ref{fig:experiments_ARCSO_noise}, where we observed that, for $k^*/\ell=100/1000$, $\ARCSplus_{\texttt{OR}}$ are competitive to $\TEASER$ in terms of accuracy (Figure \ref{fig:experiments_ARCSO_noise_error}) and to $\RANSAC$ in terms of speed (Figure \ref{fig:experiments_ARCSO_noise_time}); $10\%$ inliers are enough for $\RANSAC$ to be fast. Also note that the running time of $\TEASER$ increases exponentially as noise grows, and that $\GORE$ would achieve higher accuracy if some local refinement methods were applied. 

\subsection{Procrustes's Experiments on Stanford Bunny}
\begin{figure}
	\centering 
	\subfloat[Bunny] {\includegraphics[width=0.12\textwidth]{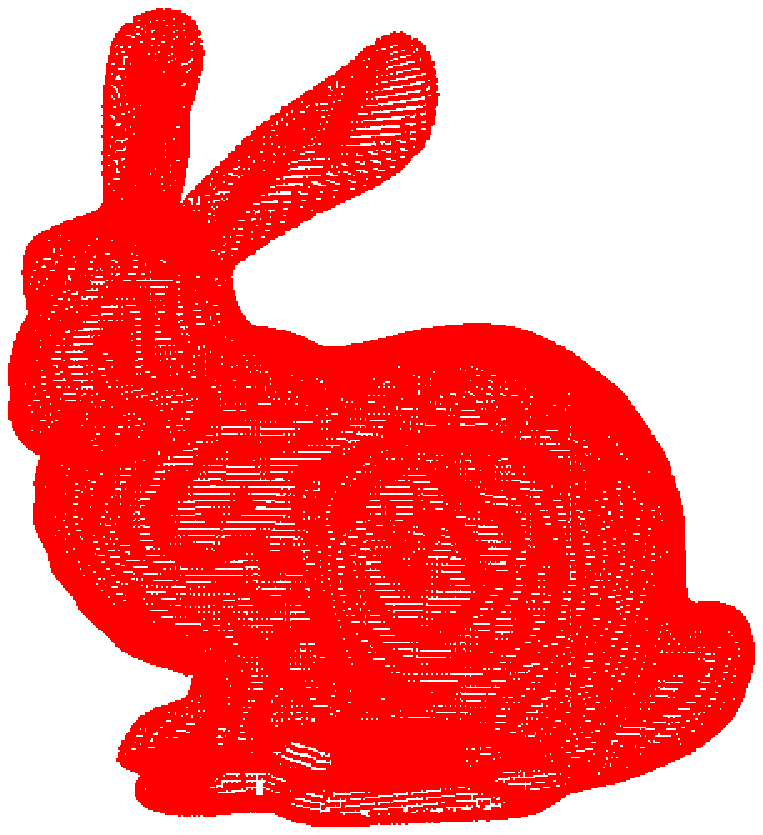} \label{fig:Bunny}  }
	\subfloat[Input ($\cQ$)] {\includegraphics[width=0.12\textwidth]{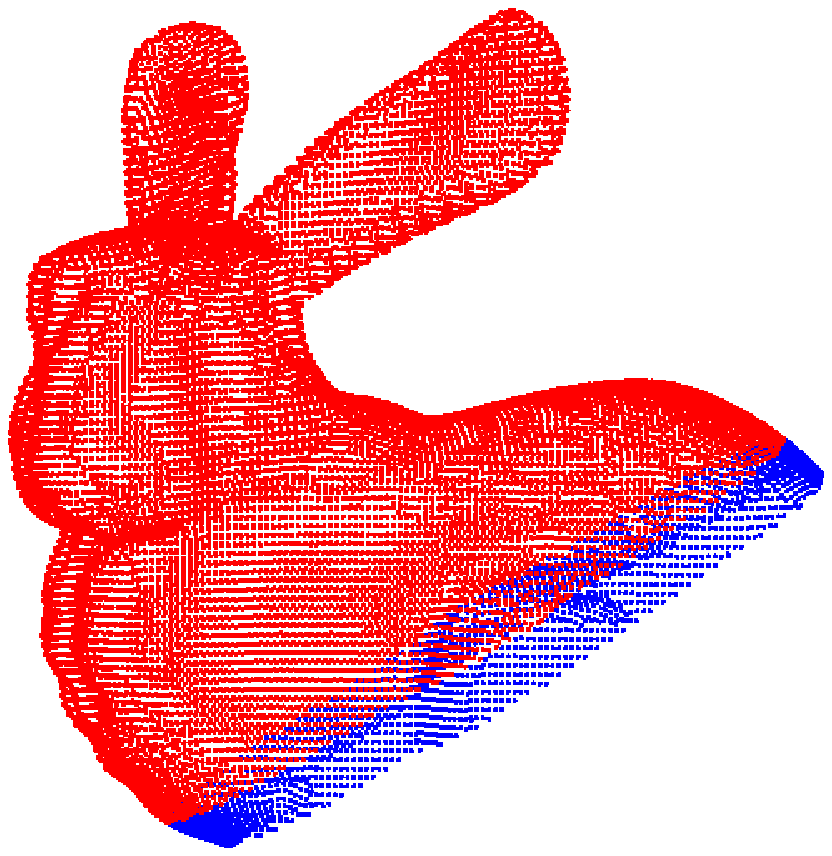} \label{fig:Bunny_Q1} }
	\subfloat[Input ($\cP$)] {\includegraphics[width=0.12\textwidth]{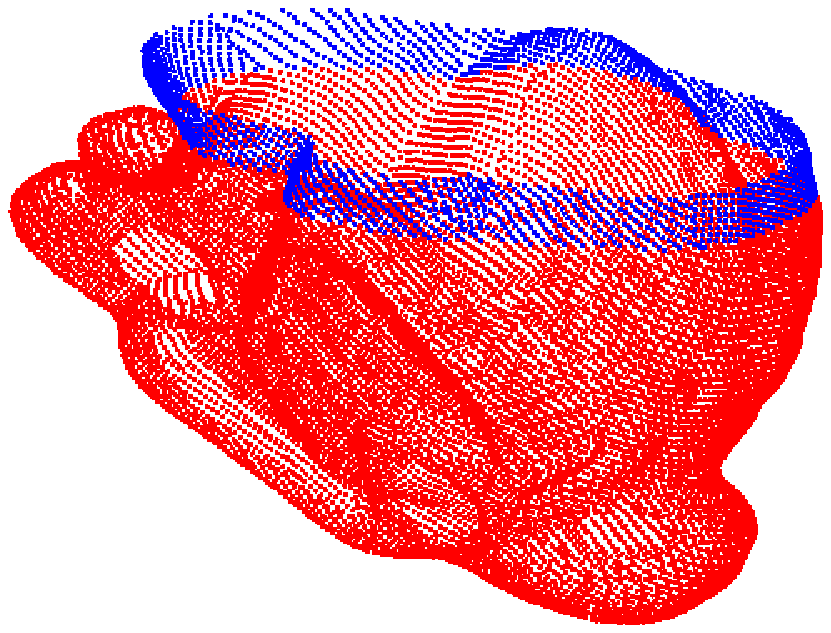} \label{fig:Bunny_P1} }
	\subfloat[$\ARCSplus$ ] {\includegraphics[width=0.12\textwidth]{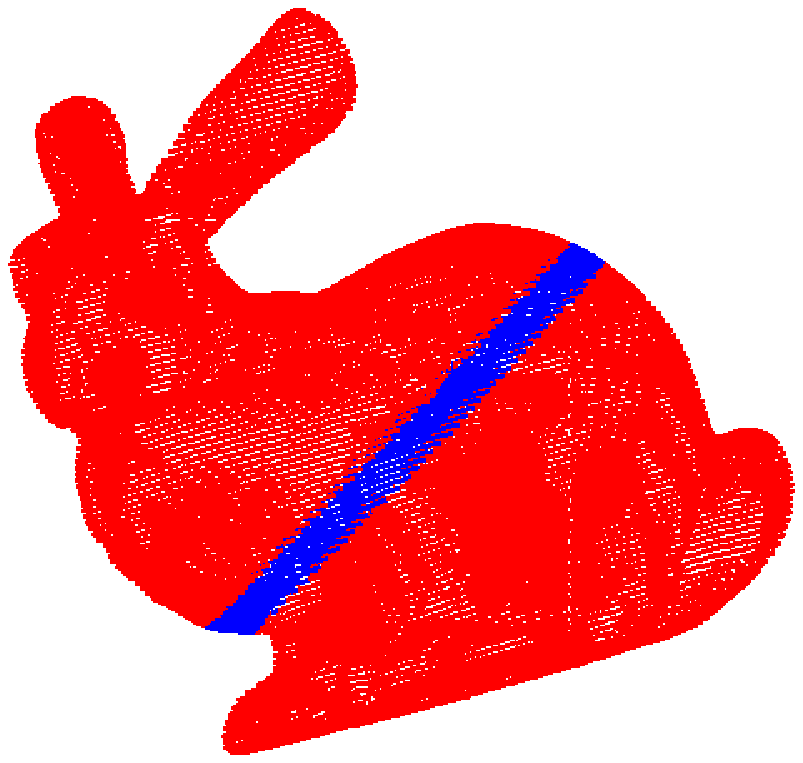} \label{fig:Bunny_I2plusplus1} }
	
	\subfloat[Input ($\cQ$)] {\includegraphics[width=0.12\textwidth]{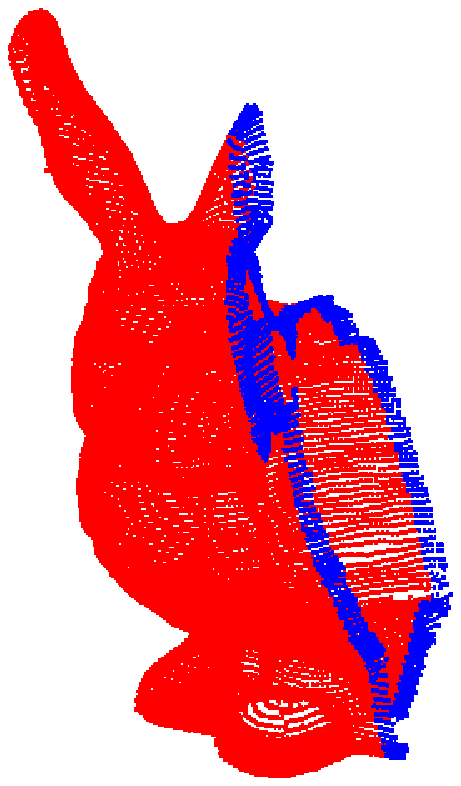} \label{fig:Bunny_Q2} }
	\subfloat[Input ($\cP$)] {\includegraphics[width=0.12\textwidth]{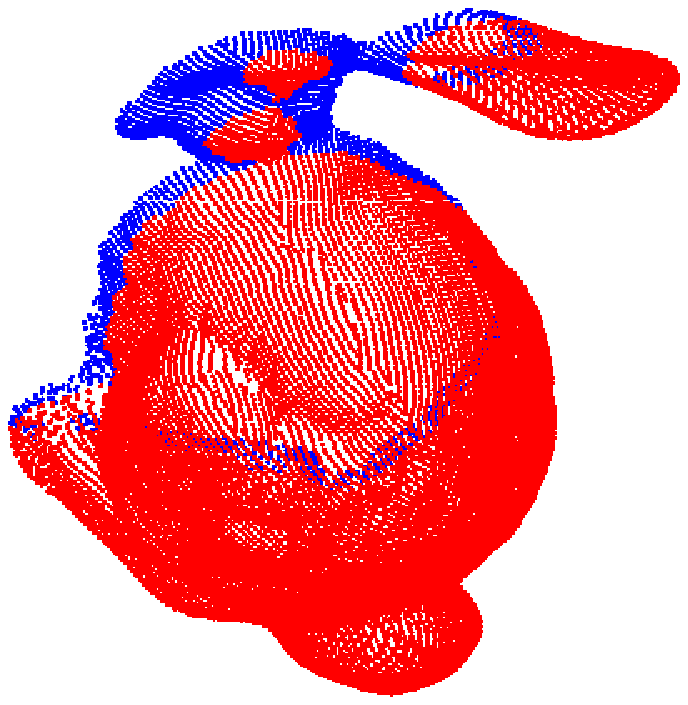} \label{fig:Bunny_P2} }
	\subfloat[$\ARCSplus$ ] {\includegraphics[width=0.12\textwidth]{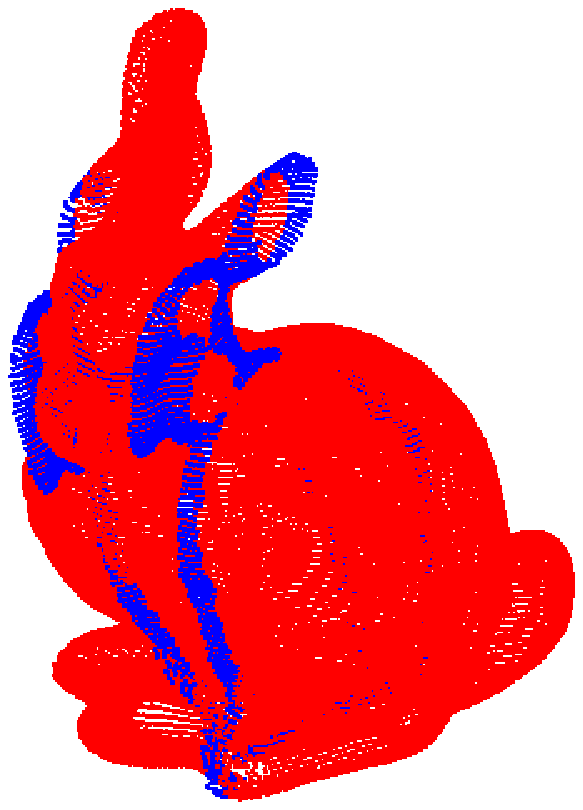} \label{fig:Bunny_I2plusplus2} }
	\caption{Bunny (\ref{fig:Bunny}) was cut through its body into two parts, $\cQ$ (\ref{fig:Bunny_Q1})  and $\cP$ (\ref{fig:Bunny_P1}), with $k^*/m=1997/19970=10\%$ overlapping points in blue. $\cP$ was randomly rotated and corrupted by $1\%$ random noise. $\ARCSplus$ successfully aligned $\cQ$ and $\cP$ (\ref{fig:Bunny_I2plusplus1}). For a different cut through the ear of Bunny (\ref{fig:Bunny_Q2}-\ref{fig:Bunny_P2}), $\ARCSplus$ failed (\ref{fig:Bunny_I2plusplus2}). }
\end{figure}

\begin{figure}
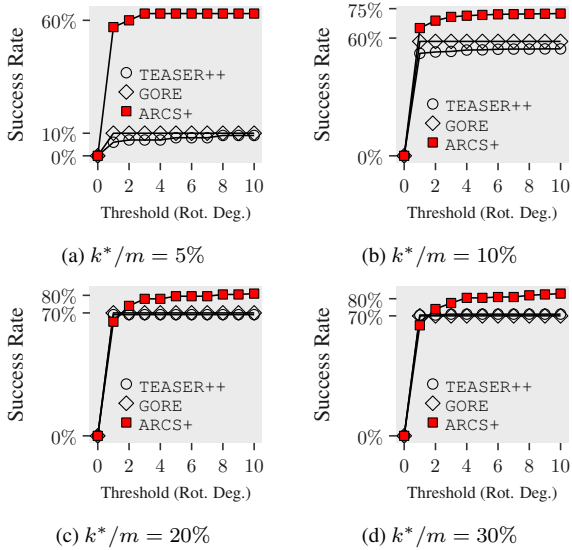

	\centering
	\subfloat[$k^*/m=5\%$] { \input{./Figures/experiments_Bunny_Overlap5.tex} }
	\subfloat[$k^*/m=10\%$] { \input{./Figures/experiments_Bunny_Overlap10.tex} }
	
	\subfloat[$k^*/m=20\%$] { \input{./Figures/experiments_Bunny_Overlap20.tex} }
	\subfloat[$k^*/m=30\%$] { \input{./Figures/experiments_Bunny_Overlap30.tex} }
	\caption{Success Rates of the methods on Stanford Bunny with different overlapping ratios $k^*/m$. $1000$ trials. In each trial, Bunny was randomly cut into two parts $\cQ$ and $\cP$, and $\cP$ was then rotated randomly and corrupted by $1\%$ random Gaussian noise. \label{fig:experiments_Bunny_Overlap} } 
\end{figure}

Here we use $\ARCSplus$ for simultaneous search of rotation
\& correspondences on a popular benchmark, the Stanford
Bunny dataset \cite{Curless-1996}.\footnote{In view of our opening quote, Bunny here is a victim of Procrustes.} Bunny has $35947$ points with every coordinate of the points located in $[-1, 1]$ (Figure \ref{fig:Bunny}). We randomly cut it into two parts, $\cQ$ and $\cP$, of sizes $m$ and $n$ respectively and of different overlapping ratios $k^*/\max\{m,n\}=k^*/m$ (Figures \ref{fig:Bunny_Q1}-\ref{fig:Bunny_P1} or \ref{fig:Bunny_Q2}-\ref{fig:Bunny_P2}). For simplicity we set $n= \ceil{35947/2}=17974$ and, $m=\floor{35947/2}+k^*$, so the exact values of $m$ and $k^*$ can be calculated as per a given overlapping ratio $k^*/m$. We then randomly rotated $\cP$ and added $1\%$ random Gaussian noise to it. The goal is to align $\cP$ and $\cQ$. $\ARCSplus$ can be applied directly to this task (Figures \ref{fig:Bunny_I2plusplus1} or \ref{fig:Bunny_I2plusplus2}). For comparison, we gave $\GORE$ and $\TEASER$ the correspondences established by $\FPFH$. For all methods we set $c=5\times 10^{-5}$. Figure \ref{fig:experiments_Bunny_Overlap} showed the results for different overlapping ratios, from which we made a few observations: $\ARCSplus$ achieved higher success rates in all experiments, while the performance of $\FPFH$, and thus of $\TEASER$ and $\GORE$, improved as the overlapping ratios increased. We did not put $\RANSAC$ into comparison here, because $\FPFH$ often gave few to none inlier pairs for small $k^*/m$ and so $\RANSAC$ used much longer time to reach a confidence of $0.99$.

\section{Handling The Translation Case}
Utilizing ideas that have been known in prior works, it is easy to extend our algorithms to the situation where there is an extra unknown translation. As we did for Problems \ref{problem:SRSC} and \ref{problem:RRS}, we first define the two problems that we will discuss:
\begin{problem}[\textit{simultaneous pose and correspondences}]\label{problem:SPC} Let the two point sets $\cQ$ and $\cP$ of Problem \ref{problem:SRSC} instead satisfy
	\begin{align}
		\bq_{i} = \bR^*\bp_j + \bt^* + \bo_{i,j} + \beps_{i,j},
	\end{align} 
	where $\bt^*\in\bbR^3$ is an extra unknown translation vector. 
	The task is to simultaneously estimate the rotation $\bR^*$, translation $\bt^*$, and correspondences $\cC^*$ from $\cQ$ and $\cP$.
\end{problem}
\begin{problem}[\textit{robust registration}]\label{problem:RR}
	Let the $\ell$ pairs of $3$D points $\{(\by_i,\bx_i)\}_{i=1}^\ell$ of Problem \ref{problem:RRS} instead satisfy
	\begin{align}\label{eq:RR-inlier}
		\by_i = \bR^* \bx_i + \bt^* + \bo_i + \beps_i.
	\end{align}
	The task is to find $\bR^*$, $\bt^*$, and correspondences $\cI^*$.
\end{problem}
We will discuss more about Problem \ref{problem:SPC} in our future work; here we focus on the its special case, Problem \ref{problem:RR}. Specifically, we next extend our $\ARCSplus_{\texttt{OR}}$ algorithm to handle Problem \ref{problem:RR} (Appendix \ref{subsection:RR}), and present its performance on the $3$DMatch dataset \cite{Zeng-CVPR2017} (Appendix \ref{subsection:3DMatch}).
\subsection{Extension for Robust Registration}\label{subsection:RR}
Here we present an extension of our $\ARCSplus_{\texttt{OR}}$ algorithm for solving Problem \ref{problem:RR}. In this extension, we essentially combine $\ARCSplus_{\texttt{OR}}$ with \textit{known} techniques. Thus, the presentation here serves more as an useful demonstration of concepts, and less as an entirely novel insight into, or the most efficient method for, solving Problem \ref{problem:RR}. Nevertheless, we will show in Appendix \ref{subsection:3DMatch} that our extension does enjoy state-of-the-art performance on the $3$DMatch dataset \cite{Zeng-CVPR2017}.

We first review three crucial ingredients that are useful for solving Problem \ref{problem:RR}: \textit{translation elimination (\texttt{TE}}), rotation elimination (\texttt{RE}), and \textit{outlier removal}.

\myparagraph{Translation Elimination (\texttt{TE})} For each $i,j\in[\ell]$, $i>j$, define $\mathrm{\mathbf{y}}_{ij}=\by_i-\by_j$ and $\mathrm{\mathbf{x}}_{ij}=\bx_i-\bx_j$, then
\begin{align}\label{eq:TIM-RRS}
	\mathrm{\mathbf{y}}_{ij} = \bR^* \mathrm{\mathbf{x}}_{ij} + (\bo_i-\bo_j) + (\beps_i - \beps_j).
\end{align}
Here $(\mathrm{\mathbf{y}}_{ij}, \mathrm{\mathbf{x}}_{ij})$ is referred to in the literature as \textit{translation invariant measurements}, as \eqref{eq:TIM-RRS} no longer involves translation. As a consequence, robust rotation search might be performed over $\{(\mathrm{\mathbf{y}}_{ij}, \mathrm{\mathbf{x}}_{ij})\}_{i>j}$, yielding an estimate of rotation and correspondences. After this, the translation can be easily computed. A disadvantage here is that computing all $(\mathrm{\mathbf{y}}_{ij}, \mathrm{\mathbf{x}}_{ij})$'s needs $O(\ell^2)$ time; also note though that this computation can be implemented in parallel and thus can be efficient for medium-size datasets (\eg, $\ell\leq 3\times 10^4$).

\myparagraph{Rotation Elimination (\texttt{RE})} Every inlier $(\by_i,\bx_i)$ satisfying \eqref{eq:RR-inlier} with $\bo_i=0$ also necessarily satisfies
\begin{align}\label{eq:RIM-SL}
	\norm{\bx_i+\beps_{i}} = \norm{\by_i-\bt^*} \Leftrightarrow \norm{\bx_i} \approx \norm{\by_i-\bt^*}.
\end{align}
If there were no outliers, estimating $t^*$ from relation \eqref{eq:RIM-SL} is the problem of \textit{source localization} that appears in signal processing applications \cite{Beck-TSP2008}. Estimating translation from \eqref{eq:RIM-SL} in the presence of outliers is more challenging. A possible algorithm is combining the least-squares solvers of \cite{Beck-TSP2008} with an iterative reweighting strategy, but this does not have global optimality guarantee. The other approach, which we employ, is to estimate $t^*$ via branch \& bound, solving the following optimization problem:
\begin{align}\label{eq:max-consensus-t}
	&\max_{\cI\subset[\ell], \bt\in\bbR^3} \ \ \ \ \ \ \  |\cI| \\
	\textnormal{s.t.}\ \ \ \  &\ \   \big| \| \by_i-\bt\|_2 - \|\bx_i\|_2\big| \leq c,\ \ \forall i\in \cI \nonumber
\end{align} 
If directly applying branch \& bound to \eqref{eq:max-consensus-t}, one would branch over $\bbR^3$ (cf. \cite{Liu-ECCV18}). On the other hand, our development in \S \ref{section:STABBER3D} implies that branching over $\bbR^2$, where the first two coordinates of $\bt$ lie, suffices, as the third coordinate can be determined by interval stabbing. 
In short, we solve \eqref{eq:max-consensus-t} via branching over the two-dimensional space $\bbR^2$ if needed. As a matter of fact, branch \& bound runs much faster even if the parameter space has smaller dimension.
\begin{remark}[\texttt{TE} versus \texttt{RE}] Translation elimination (\texttt{TE}) yields $O(\ell^2)$ measurements, leads to the problem of robust rotation search, and is also used in the $2$D-$3$D \textit{perspective-three-point} problem (see, \eg, \cite{Persson-ECCV2018}); many recent papers on  $3$D-$3$D registration used \texttt{TE} (see, \eg, \cite{Yang-T-R2021} and its follow-up works). \texttt{RE} yields $O(\ell)$ measurements, leads to a less familiar problem, and receives fewer attention; \cite{Liu-ECCV18} is the only paper, which we know, that uses \texttt{RE} (for Problem \ref{problem:SPC}).
\end{remark}

\myparagraph{Outlier Removal} Even though rotation or translation can be estimated independently of each other (using \texttt{TE} or \texttt{RE} respectively), they might not be able to handle the case of extreme outlier rates. In particular, if using \texttt{TE} then the inlier ratio decreases from $k^*/\ell$ to $O\big((k^*/\ell)^2\big)$. This is why an outlier removal procedure is needed prior to estimation. For this, create (in mind) a graph $\cG$ with $\ell$ vertices representing the $\ell$ point pairs $\{(\by_i,\bx_i)\}_{i=1}^\ell$. Moreover, create an edge between two vertices $i$ and $j$, if $|\mathrm{\mathbf{y}}_{ij} - \mathrm{\mathbf{x}}_{ij}| \leq 2c$, where $\mathrm{\mathbf{y}}_{ij}$ and  $\mathrm{\mathbf{x}}_{ij}$ are defined in \eqref{eq:TIM-RRS}. Then, find a maximum clique of $\cG$, and remove all point pairs whose corresponding vertices are not contained in the maximum clique. See \cite{Parra-arXiv2020,Yang-T-R2021,Shi-arXiv2020v2} for more transparent discussion on this idea.

For implementation, we use the code of \cite{Parra-arXiv2020} to create $\cG$ and compute a maximum clique of it.

\myparagraph{Algorithms} Having reviewed the three ingredients, we are ready to extend $\ARCSplus_{\texttt{OR}}$ for Problem \ref{problem:RR}. We have two extensions, ($\ARCSplusplus_{\texttt{OR}}$)$^{\texttt{TE}}$ and ($\ARCSplusplus_{\texttt{OR}}$)$^{\texttt{RE}}$, summarized in Table \ref{Table:arcs++}. Both of them have the same first step, outlier removal via finding a maximum clique from the constructed graph. Their next steps proceed by working with point pairs that survive from outlier removal. Step 2 of ($\ARCSplusplus_{\texttt{OR}}$)$^{\texttt{TE}}$ is to eliminate the translation (\texttt{TE}), and step 3 is to estimate the rotation via $\ARCSplus_{\texttt{OR}}$ from the point pairs $\{(\mathrm{\mathbf{y}}_{ij},\mathrm{\mathbf{x}}_{ij})\}_{i>j}$ \eqref{eq:TIM-RRS}. Step 4 of ($\ARCSplusplus_{\texttt{OR}}$)$^{\texttt{TE}}$ would estimate the translation from the remaining point pairs, with an estimated rotation given by $\ARCSplus_{\texttt{OR}}$. But we leave step 4 unspecified, as translation estimation in this situation is straightforward. On the other hand, step 2 of ($\ARCSplusplus_{\texttt{OR}}$)$^{\texttt{RE}}$ is to eliminate the rotation (\texttt{RE}), step 3 is to compute a translation $\hat{\bt}$ by solving \eqref{eq:max-consensus-t}, and step 4 is to estimate the rotation via $\ARCSplus_{\texttt{OR}}$, operating on point pairs $(\by_i-\hat{\bt},\bx_i)$'s. Finally, one might use an extra step 5, to refine the solution, \eg, by singular value decomposition.

\begin{table}
	\centering
	\caption{Two Extensions of $\ARCSplusplus_{\texttt{OR}}$ for Problem \ref{problem:RR}.\label{Table:arcs++}}
	\begin{tabular}{ccc}
		\toprule
		&  $(\ARCSplusplus_{\texttt{OR}})^{\texttt{TE}}$ & $(\ARCSplusplus_{\texttt{OR}})^{\texttt{RE}}$ \\
		\midrule
		Step 1 & \multicolumn{2}{c}{Outlier Removal} \\ 
		Step 2 & \texttt{TE} & \texttt{RE} \\
		Step 3 & $\ARCSplus_{\texttt{OR}}$ \eqref{eq:TIM-RRS} & Branch \& Bound \eqref{eq:max-consensus-t} \\
		Step 4 & --- & $\ARCSplus_{\texttt{OR}}$ \\
		Step 5 & \multicolumn{2}{c}{Local Refinement (optional)} \\
		\bottomrule 
	\end{tabular}
\end{table}

\begin{table*}
	\centering
	\caption{Success rates of methods run on the scene pairs of the 3DMatch dataset \cite{Zeng-CVPR2017} for which the ground-truth transformations are provided (rotation error smaller than $10$ degree means a success \cite{Yang-T-R2021}; see also the first paragraph of Appendix \ref{appendix:experiments}). \label{Table:3DMatch}} 
	\begin{tabular}{cccccccccc}  
		\toprule
		Scene Type & Kitchen &Home 1 & Home 2& Hotel 1& Hotel 2& Hotel 3& Study Room & MIT Lab & Overall  \\
		\# Scene Pairs & $506$ & $156$ & $208$ & $226$ & $104$ & $54$ & $292$ & $77$ & $1623$ \\
		\midrule
		($\TEASER$)$^{\texttt{TE}}$ & $98.4\%$ &$\textbf{92.9\%}$ & $89.9\%$ & $98.2\%$ & $92.3\%$ & $94.4\%$& $\textbf{93.2}\%$ & $88.3\%$ & $94.82\%$ \\
		($\TEASER$)$^{\texttt{RE}}$ & $\textbf{99.0\%}$ &$\textbf{92.3\%}$ & $89.4\%$ & $98.7\%$ & $91.3\%$ & $94.4\%$& $92.5\%$ & $88.3\%$  &   $94.76\%$ \\
		($\TEASER$)$^*$ & $\textbf{99.0\%}$ &$\textbf{98.1\%}$ & $94.7\%$ & $98.7\%$ & $\textbf{99.0\%}$ & $98.1\%$& $97.0\%$ & $94.8\%$  & $97.72\%$ \\
		\midrule
		($\ARCSplusplus_{\texttt{OR}}$)$^{\texttt{TE}}$  & $\textbf{98.6\%}$ & $92.3\%$ & $\textbf{90.4\%}$ & $\textbf{98.7\%}$ & $\textbf{93.3\%}$ & $94.4\%$ & $92.5\%$ & $88.3\%$ & $\textbf{94.89\%}$ \\
		($\ARCSplusplus_{\texttt{OR}}$)$^{\text{\texttt{RE}}}$  & $98.4\%$ & $91.7\%$ & $\textbf{89.9\%}$ & $\textbf{99.1\%}$ & $\textbf{94.2\%}$ & $94.4\%$ & $92.5\%$ & $88.3\%$ & $\textbf{94.82\%}$ \\
		($\ARCSplusplus_{\texttt{OR}}$)$^*$  & $98.4\%$ & $97.4\%$&$\textbf{95.7\%}$&$98.7\%$ & $98.1\%$ & $\textbf{100\%}$ & $97.3\%$ & $\textbf{96.1\%}$  & $97.72\%$ \\
		\bottomrule
	\end{tabular}
\end{table*}
\subsection{Experiments on 3DMatch}\label{subsection:3DMatch}
\myparagraph{Data} The 3DMatch dataset \cite{Zeng-CVPR2017} contains more than $1000$ point clouds for testing, representing $8$ different scenes (such as kitchen, hotel, etc.), while the number of  point clouds for each scene ranges from $77$ to $506$. Each point cloud has more than $10^5$ points, yet in \cite{Zeng-CVPR2017} there are $5000$ keypoints for each cloud. We used the pretrained model\footnote{\url{https://github.com/zgojcic/3DSmoothNet}} of the \texttt{3DSmoothNet} \cite{Gojcic-CVPR2019} to extract descriptors from these key points, and matched them using the Matlab function \texttt{pcmatchfeatures}, with its parameter \texttt{MatchThreshold} set to the maximum $1$. It remains to solve Problem \ref{problem:RR} using these hypothetical correspondences.
    
\myparagraph{Metrics} We report success rates of the methods. Success rates were defined in the beginning of Appendix \ref{appendix:experiments}. The default threshold $10$ on rotation degrees is the one that was used in $\TEASER$ \cite{Yang-T-R2021}. We do not report errors in terms of translation for two reasons: i) rotation search is the main theme of the paper, ii) if the rotation is estimated accurately, then so will be the translation (see, \eg, algorithms of \cite{Yang-T-R2021}).
  
\myparagraph{Methods} We apply ($\ARCSplusplus_{\texttt{OR}}$)$^{\texttt{TE}}$ and ($\ARCSplusplus_{\texttt{OR}}$)$^{\texttt{RE}}$ to restore the rotation and translation from these correspondences. We use singular value decomposition as an extra step 5 for ($\ARCSplusplus_{\texttt{OR}}$)$^{\texttt{RE}}$ to refine the solution and account for inaccuracy of translation estimation via branch \& bound \eqref{eq:max-consensus-t}. For reference, we also apply ($\ARCSplusplus_{\texttt{OR}}$)$^*$, which uses the ground-truth translation $\bt^*$ and point pairs $(\by_i-\bt^*,\bx_i)$'s to estimate a rotation via $\ARCSplusplus_{\texttt{OR}}$.
\begin{figure}
	\centering
	\subfloat[\texttt{RE}]{
\begin{tikzpicture}[x=1pt,y=1pt]
\definecolor{fillColor}{RGB}{255,255,255}
\path[use as bounding box,fill=fillColor,fill opacity=0.00] (0,0) rectangle (104.79, 95.40);
\begin{scope}
\path[clip] (  0.00,  0.00) rectangle (104.79, 95.40);
\definecolor{drawColor}{RGB}{255,255,255}
\definecolor{fillColor}{RGB}{255,255,255}

\path[draw=drawColor,line width= 0.6pt,line join=round,line cap=round,fill=fillColor] (  0.00,  0.00) rectangle (104.79, 95.40);
\end{scope}
\begin{scope}
\path[clip] ( 34.80, 27.77) rectangle ( 99.29, 89.90);
\definecolor{fillColor}{gray}{0.92}

\path[fill=fillColor] ( 34.80, 27.77) rectangle ( 99.29, 89.90);
\definecolor{drawColor}{RGB}{0,0,0}

\path[draw=drawColor,line width= 0.6pt,line join=round] ( 39.07, 30.60) --
	( 44.40, 42.71) --
	( 49.72, 62.26) --
	( 55.05, 74.16) --
	( 60.38, 79.10) --
	( 65.71, 81.37) --
	( 71.04, 82.65) --
	( 76.37, 83.38) --
	( 81.70, 83.77) --
	( 87.03, 83.94) --
	( 92.36, 84.15);

\path[draw=drawColor,line width= 0.6pt,line join=round] ( 41.73, 30.60) --
	( 47.06, 43.89) --
	( 52.39, 65.18) --
	( 57.72, 76.04) --
	( 63.05, 80.29) --
	( 68.38, 82.20) --
	( 73.71, 83.18) --
	( 79.04, 83.52) --
	( 84.37, 83.87) --
	( 89.70, 84.01) --
	( 95.03, 84.11);
\definecolor{fillColor}{RGB}{255,0,0}

\path[draw=drawColor,line width= 0.4pt,line join=round,line cap=round,fill=fillColor] ( 37.33, 28.86) rectangle ( 40.80, 32.33);

\path[draw=drawColor,line width= 0.4pt,line join=round,line cap=round,fill=fillColor] ( 42.66, 40.97) rectangle ( 46.13, 44.44);

\path[draw=drawColor,line width= 0.4pt,line join=round,line cap=round,fill=fillColor] ( 47.99, 60.52) rectangle ( 51.46, 64.00);

\path[draw=drawColor,line width= 0.4pt,line join=round,line cap=round,fill=fillColor] ( 53.32, 72.42) rectangle ( 56.79, 75.90);

\path[draw=drawColor,line width= 0.4pt,line join=round,line cap=round,fill=fillColor] ( 58.65, 77.36) rectangle ( 62.12, 80.84);

\path[draw=drawColor,line width= 0.4pt,line join=round,line cap=round,fill=fillColor] ( 63.98, 79.63) rectangle ( 67.45, 83.10);

\path[draw=drawColor,line width= 0.4pt,line join=round,line cap=round,fill=fillColor] ( 69.30, 80.91) rectangle ( 72.78, 84.39);

\path[draw=drawColor,line width= 0.4pt,line join=round,line cap=round,fill=fillColor] ( 74.63, 81.65) rectangle ( 78.11, 85.12);

\path[draw=drawColor,line width= 0.4pt,line join=round,line cap=round,fill=fillColor] ( 79.96, 82.03) rectangle ( 83.44, 85.51);

\path[draw=drawColor,line width= 0.4pt,line join=round,line cap=round,fill=fillColor] ( 85.29, 82.20) rectangle ( 88.77, 85.68);

\path[draw=drawColor,line width= 0.4pt,line join=round,line cap=round,fill=fillColor] ( 90.62, 82.41) rectangle ( 94.10, 85.89);
\definecolor{fillColor}{RGB}{190,190,190}

\path[draw=drawColor,line width= 0.4pt,line join=round,line cap=round,fill=fillColor] ( 41.73, 30.60) circle (  1.96);

\path[draw=drawColor,line width= 0.4pt,line join=round,line cap=round,fill=fillColor] ( 47.06, 43.89) circle (  1.96);

\path[draw=drawColor,line width= 0.4pt,line join=round,line cap=round,fill=fillColor] ( 52.39, 65.18) circle (  1.96);

\path[draw=drawColor,line width= 0.4pt,line join=round,line cap=round,fill=fillColor] ( 57.72, 76.04) circle (  1.96);

\path[draw=drawColor,line width= 0.4pt,line join=round,line cap=round,fill=fillColor] ( 63.05, 80.29) circle (  1.96);

\path[draw=drawColor,line width= 0.4pt,line join=round,line cap=round,fill=fillColor] ( 68.38, 82.20) circle (  1.96);

\path[draw=drawColor,line width= 0.4pt,line join=round,line cap=round,fill=fillColor] ( 73.71, 83.18) circle (  1.96);

\path[draw=drawColor,line width= 0.4pt,line join=round,line cap=round,fill=fillColor] ( 79.04, 83.52) circle (  1.96);

\path[draw=drawColor,line width= 0.4pt,line join=round,line cap=round,fill=fillColor] ( 84.37, 83.87) circle (  1.96);

\path[draw=drawColor,line width= 0.4pt,line join=round,line cap=round,fill=fillColor] ( 89.70, 84.01) circle (  1.96);

\path[draw=drawColor,line width= 0.4pt,line join=round,line cap=round,fill=fillColor] ( 95.03, 84.11) circle (  1.96);
\end{scope}
\begin{scope}
\path[clip] (  0.00,  0.00) rectangle (104.79, 95.40);
\definecolor{drawColor}{gray}{0.10}

\node[text=drawColor,anchor=base east,inner sep=0pt, outer sep=0pt, scale=  0.73] at ( 29.85, 27.57) {$0\%$};

\node[text=drawColor,anchor=base east,inner sep=0pt, outer sep=0pt, scale=  0.73] at ( 29.85, 41.68) {$25\%$};

\node[text=drawColor,anchor=base east,inner sep=0pt, outer sep=0pt, scale=  0.73] at ( 29.85, 55.80) {$50\%$};

\node[text=drawColor,anchor=base east,inner sep=0pt, outer sep=0pt, scale=  0.73] at ( 29.85, 69.92) {$75\%$};

\node[text=drawColor,anchor=base east,inner sep=0pt, outer sep=0pt, scale=  0.73] at ( 29.85, 84.04) {$100\%$};
\end{scope}
\begin{scope}
\path[clip] (  0.00,  0.00) rectangle (104.79, 95.40);
\definecolor{drawColor}{gray}{0.20}

\path[draw=drawColor,line width= 0.6pt,line join=round] ( 32.05, 30.60) --
	( 34.80, 30.60);

\path[draw=drawColor,line width= 0.6pt,line join=round] ( 32.05, 44.71) --
	( 34.80, 44.71);

\path[draw=drawColor,line width= 0.6pt,line join=round] ( 32.05, 58.83) --
	( 34.80, 58.83);

\path[draw=drawColor,line width= 0.6pt,line join=round] ( 32.05, 72.95) --
	( 34.80, 72.95);

\path[draw=drawColor,line width= 0.6pt,line join=round] ( 32.05, 87.07) --
	( 34.80, 87.07);
\end{scope}
\begin{scope}
\path[clip] (  0.00,  0.00) rectangle (104.79, 95.40);
\definecolor{drawColor}{gray}{0.20}

\path[draw=drawColor,line width= 0.6pt,line join=round] ( 40.40, 25.02) --
	( 40.40, 27.77);

\path[draw=drawColor,line width= 0.6pt,line join=round] ( 51.06, 25.02) --
	( 51.06, 27.77);

\path[draw=drawColor,line width= 0.6pt,line join=round] ( 61.72, 25.02) --
	( 61.72, 27.77);

\path[draw=drawColor,line width= 0.6pt,line join=round] ( 72.38, 25.02) --
	( 72.38, 27.77);

\path[draw=drawColor,line width= 0.6pt,line join=round] ( 83.04, 25.02) --
	( 83.04, 27.77);

\path[draw=drawColor,line width= 0.6pt,line join=round] ( 93.70, 25.02) --
	( 93.70, 27.77);
\end{scope}
\begin{scope}
\path[clip] (  0.00,  0.00) rectangle (104.79, 95.40);
\definecolor{drawColor}{gray}{0.10}

\node[text=drawColor,anchor=base,inner sep=0pt, outer sep=0pt, scale=  0.73] at ( 40.40, 16.76) {$0$};

\node[text=drawColor,anchor=base,inner sep=0pt, outer sep=0pt, scale=  0.73] at ( 51.06, 16.76) {$2$};

\node[text=drawColor,anchor=base,inner sep=0pt, outer sep=0pt, scale=  0.73] at ( 61.72, 16.76) {$4$};

\node[text=drawColor,anchor=base,inner sep=0pt, outer sep=0pt, scale=  0.73] at ( 72.38, 16.76) {$6$};

\node[text=drawColor,anchor=base,inner sep=0pt, outer sep=0pt, scale=  0.73] at ( 83.04, 16.76) {$8$};

\node[text=drawColor,anchor=base,inner sep=0pt, outer sep=0pt, scale=  0.73] at ( 93.70, 16.76) {$10$};
\end{scope}
\begin{scope}
\path[clip] (  0.00,  0.00) rectangle (104.79, 95.40);
\definecolor{drawColor}{gray}{0.10}

\node[text=drawColor,anchor=base,inner sep=0pt, outer sep=0pt, scale=  0.64] at ( 67.05,  7.00) {Rotation Err. (Deg.)};
\end{scope}
\begin{scope}
\path[clip] (  0.00,  0.00) rectangle (104.79, 95.40);
\definecolor{drawColor}{RGB}{0,0,0}
\definecolor{fillColor}{RGB}{255,0,0}

\path[draw=drawColor,line width= 0.4pt,line join=round,line cap=round,fill=fillColor] ( 54.89, 39.59) rectangle ( 58.37, 43.07);
\end{scope}
\begin{scope}
\path[clip] (  0.00,  0.00) rectangle (104.79, 95.40);
\definecolor{drawColor}{RGB}{0,0,0}
\definecolor{fillColor}{RGB}{190,190,190}

\path[draw=drawColor,line width= 0.4pt,line join=round,line cap=round,fill=fillColor] ( 56.63, 33.56) circle (  1.96);
\end{scope}
\begin{scope}
\path[clip] (  0.00,  0.00) rectangle (104.79, 95.40);
\definecolor{drawColor}{RGB}{0,0,0}

\node[text=drawColor,anchor=base west,inner sep=0pt, outer sep=0pt, scale=  0.73] at ( 61.36, 38.30) {$\texttt{ARCS++}_{\texttt{OR}}$};
\end{scope}
\begin{scope}
\path[clip] (  0.00,  0.00) rectangle (104.79, 95.40);
\definecolor{drawColor}{RGB}{0,0,0}

\node[text=drawColor,anchor=base west,inner sep=0pt, outer sep=0pt, scale=  0.73] at ( 61.36, 30.53) {\texttt{TEASER++}};
\end{scope}
\end{tikzpicture} \label{fig:3DMatch_Recall_EST}}
	\subfloat[\texttt{TE}]{
\begin{tikzpicture}[x=1pt,y=1pt]
\definecolor{fillColor}{RGB}{255,255,255}
\path[use as bounding box,fill=fillColor,fill opacity=0.00] (0,0) rectangle (104.79, 95.40);
\begin{scope}
\path[clip] (  0.00,  0.00) rectangle (104.79, 95.40);
\definecolor{drawColor}{RGB}{255,255,255}
\definecolor{fillColor}{RGB}{255,255,255}

\path[draw=drawColor,line width= 0.6pt,line join=round,line cap=round,fill=fillColor] (  0.00,  0.00) rectangle (104.79, 95.40);
\end{scope}
\begin{scope}
\path[clip] ( 34.80, 27.77) rectangle ( 99.29, 89.90);
\definecolor{fillColor}{gray}{0.92}

\path[fill=fillColor] ( 34.80, 27.77) rectangle ( 99.29, 89.90);
\definecolor{drawColor}{RGB}{0,0,0}

\path[draw=drawColor,line width= 0.6pt,line join=round] ( 39.07, 30.60) --
	( 44.40, 43.75) --
	( 49.72, 65.15) --
	( 55.05, 75.90) --
	( 60.38, 80.15) --
	( 65.71, 82.10) --
	( 71.04, 83.24) --
	( 76.37, 83.42) --
	( 81.70, 83.94) --
	( 87.03, 84.11) --
	( 92.36, 84.18);

\path[draw=drawColor,line width= 0.6pt,line join=round] ( 41.73, 30.60) --
	( 47.06, 43.26) --
	( 52.39, 64.66) --
	( 57.72, 75.38) --
	( 63.05, 80.04) --
	( 68.38, 81.78) --
	( 73.71, 82.76) --
	( 79.04, 83.35) --
	( 84.37, 83.80) --
	( 89.70, 84.11) --
	( 95.03, 84.15);
\definecolor{fillColor}{RGB}{255,0,0}

\path[draw=drawColor,line width= 0.4pt,line join=round,line cap=round,fill=fillColor] ( 37.33, 28.86) rectangle ( 40.80, 32.33);

\path[draw=drawColor,line width= 0.4pt,line join=round,line cap=round,fill=fillColor] ( 42.66, 42.01) rectangle ( 46.13, 45.49);

\path[draw=drawColor,line width= 0.4pt,line join=round,line cap=round,fill=fillColor] ( 47.99, 63.41) rectangle ( 51.46, 66.89);

\path[draw=drawColor,line width= 0.4pt,line join=round,line cap=round,fill=fillColor] ( 53.32, 74.16) rectangle ( 56.79, 77.64);

\path[draw=drawColor,line width= 0.4pt,line join=round,line cap=round,fill=fillColor] ( 58.65, 78.41) rectangle ( 62.12, 81.89);

\path[draw=drawColor,line width= 0.4pt,line join=round,line cap=round,fill=fillColor] ( 63.98, 80.36) rectangle ( 67.45, 83.84);

\path[draw=drawColor,line width= 0.4pt,line join=round,line cap=round,fill=fillColor] ( 69.30, 81.51) rectangle ( 72.78, 84.98);

\path[draw=drawColor,line width= 0.4pt,line join=round,line cap=round,fill=fillColor] ( 74.63, 81.68) rectangle ( 78.11, 85.16);

\path[draw=drawColor,line width= 0.4pt,line join=round,line cap=round,fill=fillColor] ( 79.96, 82.20) rectangle ( 83.44, 85.68);

\path[draw=drawColor,line width= 0.4pt,line join=round,line cap=round,fill=fillColor] ( 85.29, 82.38) rectangle ( 88.77, 85.85);

\path[draw=drawColor,line width= 0.4pt,line join=round,line cap=round,fill=fillColor] ( 90.62, 82.45) rectangle ( 94.10, 85.92);
\definecolor{fillColor}{RGB}{190,190,190}

\path[draw=drawColor,line width= 0.4pt,line join=round,line cap=round,fill=fillColor] ( 41.73, 30.60) circle (  1.96);

\path[draw=drawColor,line width= 0.4pt,line join=round,line cap=round,fill=fillColor] ( 47.06, 43.26) circle (  1.96);

\path[draw=drawColor,line width= 0.4pt,line join=round,line cap=round,fill=fillColor] ( 52.39, 64.66) circle (  1.96);

\path[draw=drawColor,line width= 0.4pt,line join=round,line cap=round,fill=fillColor] ( 57.72, 75.38) circle (  1.96);

\path[draw=drawColor,line width= 0.4pt,line join=round,line cap=round,fill=fillColor] ( 63.05, 80.04) circle (  1.96);

\path[draw=drawColor,line width= 0.4pt,line join=round,line cap=round,fill=fillColor] ( 68.38, 81.78) circle (  1.96);

\path[draw=drawColor,line width= 0.4pt,line join=round,line cap=round,fill=fillColor] ( 73.71, 82.76) circle (  1.96);

\path[draw=drawColor,line width= 0.4pt,line join=round,line cap=round,fill=fillColor] ( 79.04, 83.35) circle (  1.96);

\path[draw=drawColor,line width= 0.4pt,line join=round,line cap=round,fill=fillColor] ( 84.37, 83.80) circle (  1.96);

\path[draw=drawColor,line width= 0.4pt,line join=round,line cap=round,fill=fillColor] ( 89.70, 84.11) circle (  1.96);

\path[draw=drawColor,line width= 0.4pt,line join=round,line cap=round,fill=fillColor] ( 95.03, 84.15) circle (  1.96);
\end{scope}
\begin{scope}
\path[clip] (  0.00,  0.00) rectangle (104.79, 95.40);
\definecolor{drawColor}{gray}{0.10}

\node[text=drawColor,anchor=base east,inner sep=0pt, outer sep=0pt, scale=  0.73] at ( 29.85, 27.57) {$0\%$};

\node[text=drawColor,anchor=base east,inner sep=0pt, outer sep=0pt, scale=  0.73] at ( 29.85, 41.68) {$25\%$};

\node[text=drawColor,anchor=base east,inner sep=0pt, outer sep=0pt, scale=  0.73] at ( 29.85, 55.80) {$50\%$};

\node[text=drawColor,anchor=base east,inner sep=0pt, outer sep=0pt, scale=  0.73] at ( 29.85, 69.92) {$75\%$};

\node[text=drawColor,anchor=base east,inner sep=0pt, outer sep=0pt, scale=  0.73] at ( 29.85, 84.04) {$100\%$};
\end{scope}
\begin{scope}
\path[clip] (  0.00,  0.00) rectangle (104.79, 95.40);
\definecolor{drawColor}{gray}{0.20}

\path[draw=drawColor,line width= 0.6pt,line join=round] ( 32.05, 30.60) --
	( 34.80, 30.60);

\path[draw=drawColor,line width= 0.6pt,line join=round] ( 32.05, 44.71) --
	( 34.80, 44.71);

\path[draw=drawColor,line width= 0.6pt,line join=round] ( 32.05, 58.83) --
	( 34.80, 58.83);

\path[draw=drawColor,line width= 0.6pt,line join=round] ( 32.05, 72.95) --
	( 34.80, 72.95);

\path[draw=drawColor,line width= 0.6pt,line join=round] ( 32.05, 87.07) --
	( 34.80, 87.07);
\end{scope}
\begin{scope}
\path[clip] (  0.00,  0.00) rectangle (104.79, 95.40);
\definecolor{drawColor}{gray}{0.20}

\path[draw=drawColor,line width= 0.6pt,line join=round] ( 40.40, 25.02) --
	( 40.40, 27.77);

\path[draw=drawColor,line width= 0.6pt,line join=round] ( 51.06, 25.02) --
	( 51.06, 27.77);

\path[draw=drawColor,line width= 0.6pt,line join=round] ( 61.72, 25.02) --
	( 61.72, 27.77);

\path[draw=drawColor,line width= 0.6pt,line join=round] ( 72.38, 25.02) --
	( 72.38, 27.77);

\path[draw=drawColor,line width= 0.6pt,line join=round] ( 83.04, 25.02) --
	( 83.04, 27.77);

\path[draw=drawColor,line width= 0.6pt,line join=round] ( 93.70, 25.02) --
	( 93.70, 27.77);
\end{scope}
\begin{scope}
\path[clip] (  0.00,  0.00) rectangle (104.79, 95.40);
\definecolor{drawColor}{gray}{0.10}

\node[text=drawColor,anchor=base,inner sep=0pt, outer sep=0pt, scale=  0.73] at ( 40.40, 16.76) {$0$};

\node[text=drawColor,anchor=base,inner sep=0pt, outer sep=0pt, scale=  0.73] at ( 51.06, 16.76) {$2$};

\node[text=drawColor,anchor=base,inner sep=0pt, outer sep=0pt, scale=  0.73] at ( 61.72, 16.76) {$4$};

\node[text=drawColor,anchor=base,inner sep=0pt, outer sep=0pt, scale=  0.73] at ( 72.38, 16.76) {$6$};

\node[text=drawColor,anchor=base,inner sep=0pt, outer sep=0pt, scale=  0.73] at ( 83.04, 16.76) {$8$};

\node[text=drawColor,anchor=base,inner sep=0pt, outer sep=0pt, scale=  0.73] at ( 93.70, 16.76) {$10$};
\end{scope}
\begin{scope}
\path[clip] (  0.00,  0.00) rectangle (104.79, 95.40);
\definecolor{drawColor}{gray}{0.10}

\node[text=drawColor,anchor=base,inner sep=0pt, outer sep=0pt, scale=  0.64] at ( 67.05,  7.00) {Rotation Err. (Deg.)};
\end{scope}
\begin{scope}
\path[clip] (  0.00,  0.00) rectangle (104.79, 95.40);
\definecolor{drawColor}{RGB}{0,0,0}
\definecolor{fillColor}{RGB}{255,0,0}

\path[draw=drawColor,line width= 0.4pt,line join=round,line cap=round,fill=fillColor] ( 54.89, 39.59) rectangle ( 58.37, 43.07);
\end{scope}
\begin{scope}
\path[clip] (  0.00,  0.00) rectangle (104.79, 95.40);
\definecolor{drawColor}{RGB}{0,0,0}
\definecolor{fillColor}{RGB}{190,190,190}

\path[draw=drawColor,line width= 0.4pt,line join=round,line cap=round,fill=fillColor] ( 56.63, 33.56) circle (  1.96);
\end{scope}
\begin{scope}
\path[clip] (  0.00,  0.00) rectangle (104.79, 95.40);
\definecolor{drawColor}{RGB}{0,0,0}

\node[text=drawColor,anchor=base west,inner sep=0pt, outer sep=0pt, scale=  0.73] at ( 61.36, 38.30) {$\texttt{ARCS++}_{\texttt{OR}}$};
\end{scope}
\begin{scope}
\path[clip] (  0.00,  0.00) rectangle (104.79, 95.40);
\definecolor{drawColor}{RGB}{0,0,0}

\node[text=drawColor,anchor=base west,inner sep=0pt, outer sep=0pt, scale=  0.73] at ( 61.36, 30.53) {\texttt{TEASER++}};
\end{scope}
\end{tikzpicture}\label{fig:3DMatch_Recall_TIM}}
	
	\subfloat[Ground-Truth Translation]{
\begin{tikzpicture}[x=1pt,y=1pt]
\definecolor{fillColor}{RGB}{255,255,255}
\path[use as bounding box,fill=fillColor,fill opacity=0.00] (0,0) rectangle (110.57, 95.40);
\begin{scope}
\path[clip] (  0.00,  0.00) rectangle (110.57, 95.40);
\definecolor{drawColor}{RGB}{255,255,255}
\definecolor{fillColor}{RGB}{255,255,255}

\path[draw=drawColor,line width= 0.6pt,line join=round,line cap=round,fill=fillColor] (  0.00,  0.00) rectangle (110.57, 95.40);
\end{scope}
\begin{scope}
\path[clip] ( 43.50, 27.77) rectangle (105.07, 89.90);
\definecolor{fillColor}{gray}{0.92}

\path[fill=fillColor] ( 43.50, 27.77) rectangle (105.07, 89.90);
\definecolor{drawColor}{RGB}{0,0,0}

\path[draw=drawColor,line width= 0.6pt,line join=round] ( 47.58, 30.60) --
	( 52.66, 57.49) --
	( 57.75, 76.98) --
	( 62.84, 82.58) --
	( 67.93, 84.39) --
	( 73.02, 85.02) --
	( 78.11, 85.33) --
	( 83.19, 85.47) --
	( 88.28, 85.61) --
	( 93.37, 85.68) --
	( 98.46, 85.79);

\path[draw=drawColor,line width= 0.6pt,line join=round] ( 50.12, 30.60) --
	( 55.21, 43.47) --
	( 60.30, 65.88) --
	( 65.38, 76.84) --
	( 70.47, 81.92) --
	( 75.56, 83.70) --
	( 80.65, 85.02) --
	( 85.74, 85.37) --
	( 90.83, 85.47) --
	( 95.91, 85.72) --
	(101.00, 85.79);
\definecolor{fillColor}{RGB}{255,0,0}

\path[draw=drawColor,line width= 0.4pt,line join=round,line cap=round,fill=fillColor] ( 45.84, 28.86) rectangle ( 49.31, 32.33);

\path[draw=drawColor,line width= 0.4pt,line join=round,line cap=round,fill=fillColor] ( 50.92, 55.76) rectangle ( 54.40, 59.23);

\path[draw=drawColor,line width= 0.4pt,line join=round,line cap=round,fill=fillColor] ( 56.01, 75.24) rectangle ( 59.49, 78.72);

\path[draw=drawColor,line width= 0.4pt,line join=round,line cap=round,fill=fillColor] ( 61.10, 80.84) rectangle ( 64.58, 84.32);

\path[draw=drawColor,line width= 0.4pt,line join=round,line cap=round,fill=fillColor] ( 66.19, 82.65) rectangle ( 69.67, 86.13);

\path[draw=drawColor,line width= 0.4pt,line join=round,line cap=round,fill=fillColor] ( 71.28, 83.28) rectangle ( 74.76, 86.76);

\path[draw=drawColor,line width= 0.4pt,line join=round,line cap=round,fill=fillColor] ( 76.37, 83.59) rectangle ( 79.84, 87.07);

\path[draw=drawColor,line width= 0.4pt,line join=round,line cap=round,fill=fillColor] ( 81.45, 83.73) rectangle ( 84.93, 87.21);

\path[draw=drawColor,line width= 0.4pt,line join=round,line cap=round,fill=fillColor] ( 86.54, 83.87) rectangle ( 90.02, 87.35);

\path[draw=drawColor,line width= 0.4pt,line join=round,line cap=round,fill=fillColor] ( 91.63, 83.94) rectangle ( 95.11, 87.42);

\path[draw=drawColor,line width= 0.4pt,line join=round,line cap=round,fill=fillColor] ( 96.72, 84.05) rectangle (100.20, 87.52);
\definecolor{fillColor}{RGB}{190,190,190}

\path[draw=drawColor,line width= 0.4pt,line join=round,line cap=round,fill=fillColor] ( 50.12, 30.60) circle (  1.96);

\path[draw=drawColor,line width= 0.4pt,line join=round,line cap=round,fill=fillColor] ( 55.21, 43.47) circle (  1.96);

\path[draw=drawColor,line width= 0.4pt,line join=round,line cap=round,fill=fillColor] ( 60.30, 65.88) circle (  1.96);

\path[draw=drawColor,line width= 0.4pt,line join=round,line cap=round,fill=fillColor] ( 65.38, 76.84) circle (  1.96);

\path[draw=drawColor,line width= 0.4pt,line join=round,line cap=round,fill=fillColor] ( 70.47, 81.92) circle (  1.96);

\path[draw=drawColor,line width= 0.4pt,line join=round,line cap=round,fill=fillColor] ( 75.56, 83.70) circle (  1.96);

\path[draw=drawColor,line width= 0.4pt,line join=round,line cap=round,fill=fillColor] ( 80.65, 85.02) circle (  1.96);

\path[draw=drawColor,line width= 0.4pt,line join=round,line cap=round,fill=fillColor] ( 85.74, 85.37) circle (  1.96);

\path[draw=drawColor,line width= 0.4pt,line join=round,line cap=round,fill=fillColor] ( 90.83, 85.47) circle (  1.96);

\path[draw=drawColor,line width= 0.4pt,line join=round,line cap=round,fill=fillColor] ( 95.91, 85.72) circle (  1.96);

\path[draw=drawColor,line width= 0.4pt,line join=round,line cap=round,fill=fillColor] (101.00, 85.79) circle (  1.96);
\end{scope}
\begin{scope}
\path[clip] (  0.00,  0.00) rectangle (110.57, 95.40);
\definecolor{drawColor}{gray}{0.10}

\node[text=drawColor,anchor=base east,inner sep=0pt, outer sep=0pt, scale=  0.73] at ( 38.55, 27.57) {$0\%$};

\node[text=drawColor,anchor=base east,inner sep=0pt, outer sep=0pt, scale=  0.73] at ( 38.55, 41.68) {$25\%$};

\node[text=drawColor,anchor=base east,inner sep=0pt, outer sep=0pt, scale=  0.73] at ( 38.55, 55.80) {$50\%$};

\node[text=drawColor,anchor=base east,inner sep=0pt, outer sep=0pt, scale=  0.73] at ( 38.55, 69.92) {$75\%$};

\node[text=drawColor,anchor=base east,inner sep=0pt, outer sep=0pt, scale=  0.73] at ( 38.55, 84.04) {$100\%$};
\end{scope}
\begin{scope}
\path[clip] (  0.00,  0.00) rectangle (110.57, 95.40);
\definecolor{drawColor}{gray}{0.20}

\path[draw=drawColor,line width= 0.6pt,line join=round] ( 40.75, 30.60) --
	( 43.50, 30.60);

\path[draw=drawColor,line width= 0.6pt,line join=round] ( 40.75, 44.71) --
	( 43.50, 44.71);

\path[draw=drawColor,line width= 0.6pt,line join=round] ( 40.75, 58.83) --
	( 43.50, 58.83);

\path[draw=drawColor,line width= 0.6pt,line join=round] ( 40.75, 72.95) --
	( 43.50, 72.95);

\path[draw=drawColor,line width= 0.6pt,line join=round] ( 40.75, 87.07) --
	( 43.50, 87.07);
\end{scope}
\begin{scope}
\path[clip] (  0.00,  0.00) rectangle (110.57, 95.40);
\definecolor{drawColor}{gray}{0.20}

\path[draw=drawColor,line width= 0.6pt,line join=round] ( 48.85, 25.02) --
	( 48.85, 27.77);

\path[draw=drawColor,line width= 0.6pt,line join=round] ( 59.02, 25.02) --
	( 59.02, 27.77);

\path[draw=drawColor,line width= 0.6pt,line join=round] ( 69.20, 25.02) --
	( 69.20, 27.77);

\path[draw=drawColor,line width= 0.6pt,line join=round] ( 79.38, 25.02) --
	( 79.38, 27.77);

\path[draw=drawColor,line width= 0.6pt,line join=round] ( 89.55, 25.02) --
	( 89.55, 27.77);

\path[draw=drawColor,line width= 0.6pt,line join=round] ( 99.73, 25.02) --
	( 99.73, 27.77);
\end{scope}
\begin{scope}
\path[clip] (  0.00,  0.00) rectangle (110.57, 95.40);
\definecolor{drawColor}{gray}{0.10}

\node[text=drawColor,anchor=base,inner sep=0pt, outer sep=0pt, scale=  0.73] at ( 48.85, 16.76) {$0$};

\node[text=drawColor,anchor=base,inner sep=0pt, outer sep=0pt, scale=  0.73] at ( 59.02, 16.76) {$2$};

\node[text=drawColor,anchor=base,inner sep=0pt, outer sep=0pt, scale=  0.73] at ( 69.20, 16.76) {$4$};

\node[text=drawColor,anchor=base,inner sep=0pt, outer sep=0pt, scale=  0.73] at ( 79.38, 16.76) {$6$};

\node[text=drawColor,anchor=base,inner sep=0pt, outer sep=0pt, scale=  0.73] at ( 89.55, 16.76) {$8$};

\node[text=drawColor,anchor=base,inner sep=0pt, outer sep=0pt, scale=  0.73] at ( 99.73, 16.76) {$10$};
\end{scope}
\begin{scope}
\path[clip] (  0.00,  0.00) rectangle (110.57, 95.40);
\definecolor{drawColor}{gray}{0.10}

\node[text=drawColor,anchor=base,inner sep=0pt, outer sep=0pt, scale=  0.64] at ( 74.29,  7.00) {Rotation Err. (Deg.)};
\end{scope}
\begin{scope}
\path[clip] (  0.00,  0.00) rectangle (110.57, 95.40);
\definecolor{drawColor}{gray}{0.10}

\node[text=drawColor,rotate= 90.00,anchor=base,inner sep=0pt, outer sep=0pt, scale=  0.96] at ( 13.45, 58.83) {Success Rate};
\end{scope}
\begin{scope}
\path[clip] (  0.00,  0.00) rectangle (110.57, 95.40);
\definecolor{drawColor}{RGB}{0,0,0}
\definecolor{fillColor}{RGB}{255,0,0}

\path[draw=drawColor,line width= 0.4pt,line join=round,line cap=round,fill=fillColor] ( 61.69, 39.59) rectangle ( 65.17, 43.07);
\end{scope}
\begin{scope}
\path[clip] (  0.00,  0.00) rectangle (110.57, 95.40);
\definecolor{drawColor}{RGB}{0,0,0}
\definecolor{fillColor}{RGB}{190,190,190}

\path[draw=drawColor,line width= 0.4pt,line join=round,line cap=round,fill=fillColor] ( 63.43, 33.56) circle (  1.96);
\end{scope}
\begin{scope}
\path[clip] (  0.00,  0.00) rectangle (110.57, 95.40);
\definecolor{drawColor}{RGB}{0,0,0}

\node[text=drawColor,anchor=base west,inner sep=0pt, outer sep=0pt, scale=  0.73] at ( 68.17, 38.30) {$\texttt{ARCS++}_{\texttt{OR}}$};
\end{scope}
\begin{scope}
\path[clip] (  0.00,  0.00) rectangle (110.57, 95.40);
\definecolor{drawColor}{RGB}{0,0,0}

\node[text=drawColor,anchor=base west,inner sep=0pt, outer sep=0pt, scale=  0.73] at ( 68.17, 30.53) {\texttt{TEASER++}};
\end{scope}
\end{tikzpicture}\label{fig:3DMatch_Recall_GT}}
	\caption{Success rates of $\TEASER$ and $\ARCSplusplus_{\texttt{OR}}$ on the 3D Match dataset, using either estimated translation (Fig. \ref{fig:3DMatch_Recall_EST}) or $\TIM$s (Fig. \ref{fig:3DMatch_Recall_TIM}) or ground-truth translation (Fig. \ref{fig:3DMatch_Recall_GT}).\label{fig:3DMatch} }
\end{figure}
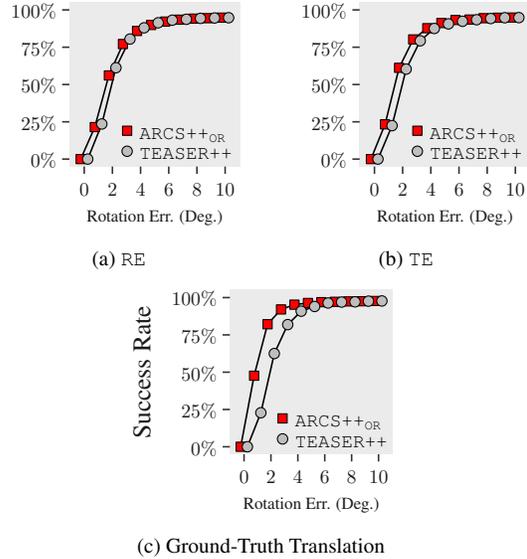

We compare our algorithms with $\TEASER$ \cite{Yang-T-R2021}. Similarly, we use three versions of $\TEASER$. The first version is ($\TEASER$)$^{\texttt{TE}}$. This is the standard $\TEASER$, and the difference between ($\TEASER$)$^{\texttt{TE}}$ and ($\ARCSplusplus_{\texttt{OR}}$)$^{\texttt{TE}}$ is that, ($\TEASER$)$^{\texttt{TE}}$ estimates the rotation by $\GNCTLS$, not $\ARCSplusplus_{\texttt{OR}}$. The second version is ($\TEASER$)$^{\texttt{TE}}$, where we treat $\TEASER$ as a robust rotation search method and let it play the role of $\ARCSplusplus_{\texttt{OR}}$ in ($\ARCSplusplus_{\texttt{OR}}$)$^{\texttt{RE}}$. The third version is ($\TEASER$)$^*$, where we assume the ground-truth translation $\bt^*$ is given and run $\TEASER$ on $(\by_i-\bt^*,\bx_i)$'s. Finally, we did not compare other methods here, as $\TEASER$ currently has the best performance (to the best of our knowledge) on the 3DMatch dataset, see \cite{Yang-T-R2021} for comparison with optimization-based methods, and also read from \cite{Choy-CVPR2020} the success rates (recall) of other deep learning methods.

\myparagraph{Results} Following \cite{Yang-T-R2021}, we set $c=0.05$. We presented results in Table \ref{Table:3DMatch} and Figure \ref{fig:3DMatch}. In Table \ref{Table:3DMatch}  we observed that $\ARCSplusplus_{\texttt{OR}}$ and $\TEASER$ have very close performance, although $\ARCSplusplus_{\texttt{OR}}$ has slight advantage (\eg, in $12$ cases in bold $\ARCSplusplus_{\texttt{OR}}$ has higher success rates). In terms of running times, $\ARCSplusplus_{\texttt{OR}}$ is slower than $\TEASER$. One reason is that we used an industrial-strength implementation\footnote{\url{https://github.com/MIT-SPARK/TEASER-plusplus}} of $\TEASER$, while $\ARCSplusplus_{\texttt{OR}}$ was implemented in plain Matlab. This suggests our current idea of extending $\ARCSplus_{\texttt{OR}}$ into the translation case might be sub-optimal, and will motivate us to design even faster algorithms for that purpose, which though will require serious innovations. Finally, in Figure \ref{fig:3DMatch}, we reported the success rates averaged over all testing scenes of $3$DMatch and with the threshold (rotation degree) varying from $0$ to $100$. This delivers the same message that our direct extension of $\ARCSplus_{\texttt{OR}}$ maintains a state-of-the-art performance for solving Problem \ref{problem:RR}.
%

\end{document}